\documentclass{article}

\usepackage[numbers]{natbib}

\usepackage[final]{neurips_2024}

\usepackage[utf8]{inputenc} 
\usepackage[T1]{fontenc}    
\usepackage{hyperref}       
\usepackage{url}            
\usepackage{booktabs}       
\usepackage{amsfonts}       
\usepackage{nicefrac}       
\usepackage{microtype}      

\usepackage{amsmath}
\usepackage{amssymb}
\usepackage{mathtools}
\usepackage{amsthm}
\usepackage{upgreek}
\usepackage{algorithmic}
\usepackage{algorithm}

\usepackage[dvipsnames]{xcolor}

\usepackage[capitalize,noabbrev]{cleveref}
\usepackage{autonum}

\newcommand{\yw}[1]{{}}

\newcommand{\normal}{\mathcal{N}}

\title{Non-Stationary Learning of Neural Networks\\ with Automatic Soft Parameter Reset}

\author{%
  Alexandre Galashov\thanks{Corresponding author} \\
  UCL Gatsby \\
  Google DeepMind \\
  \texttt{agalashov@google.com} \\
  \And
  Michalis K. Titsias \\
  Google DeepMind \\
  \texttt{mtitsias@google.com} \\
  \And
  Andr\'as Gy\"orgy \\
  Google DeepMind \\
  \texttt{agyorgy@google.com} \\
  \And
  Clare Lyle \\
  Google DeepMind \\
  \texttt{clarelyle@google.com} \\
  \And
  Razvan Pascanu \\
  Google DeepMind \\
  \texttt{razp@google.com} \\
  \And
  Yee Whye Teh \\
  Google DeepMind \\
  University of Oxford \\
  \texttt{ywteh@google.com} \\
  \And
  Maneesh Sahani \\
  UCL Gatsby \\
  \texttt{maneesh@gatsby.ucl.ac.uk} \\
}

\begin{document}

\maketitle

\begin{abstract}
    Neural networks are traditionally trained under the assumption that data come from a stationary distribution. However, settings which violate this assumption are becoming more popular;  examples include supervised learning under distributional shifts, reinforcement learning, continual learning and non-stationary contextual bandits. In this work we introduce a novel learning approach that automatically models and adapts to non-stationarity, via an Ornstein-Uhlenbeck process with an adaptive drift parameter. The adaptive drift tends to draw the parameters towards the initialisation distribution, so the approach can be understood as a form of \emph{soft} parameter reset. We show empirically that our approach performs well in non-stationary supervised and off-policy reinforcement learning settings.
\end{abstract}

\section{Introduction}

Neural networks (NNs) are typically trained using algorithms like stochastic gradient descent (SGD), assuming data comes from a stationary distribution. This assumption fails in scenarios such as continual learning, reinforcement learning, non-stationary contextual bandits, and supervised learning with distribution shifts~\citep{HADSELL20201028,verwimp2024continual}. A phenomenon occurring in non-stationary settings is the \textit{loss of plasticity}~\citep{Dohare2024LossOP,ash2020warmstarting,dohare2022continual}, manifesting either as a failure to generalize to new data despite reduced training loss~\citep{berariu2021study,ash2020warmstarting}, or as an inability to reduce training error as the data distribution changes~\citep{dohare2022continual,lyle2022understanding,abbas2023loss,nikishin2023deep,kumar2023maintaining}.

In~\citep{lyle2024disentanglingcausesplasticityloss}, the authors argue for two factors that lead to the loss of plasticity: preactivation distribution shift, leading to dead or dormant neurons~\cite{sokar23a}, and parameter norm growth causing training instabilities.
To address these issues, strategies often involve \emph{hard resets} based on heuristics like detecting dormant units \citep{sokar23a}, assessing neuron utility \citep{dohare2022continual,Dohare2024LossOP}, or simply after a fixed number of steps \citep{nikishin2022primacy}. Though effective at increasing plasticity, hard resets can be inefficient as they can discard valuable knowledge captured by the parameters.

We propose an algorithm that implements a mechanism of \emph{soft} parameter resets, in contrast to the \emph{hard} resets discussed earlier. A \emph{soft} reset partially moves NN parameters towards the initialization while keeping them close to their previous values. It also increases learning rate of the learning algorithm, allowing new NN parameters to adapt faster to the changing data. The amount by which the parameters move towards the initialization and the amount of learning rate increase are controlled by the \emph{drift} parameters which are learned online. The exact implementation of \emph{soft} reset mechanism is based on the use of a \emph{drift} model in NN parameters update \emph{before} observing new data. Similar ideas which modify the starting point of SGD as well as increase the learning rate of SGD depending on non-stationarity were explored (see~\citep{hall2013dynamical,khodak2019adaptive}) in an online convex optimization setting. Specifically, in \citep{hall2013dynamical}, the authors assume that the optimal parameter of SGD changes according to some dynamical model out of a finite family of models. They propose an algorithm to identify this model and a way to leverage this model in a modified SGD algorithm. Compared to these works, we operate in a general non-convex setting. Proposed drift model can be thought as a dynamical Bayesian prior over Neural Network parameters, which is adapted online to new data. We make a specific choice of \emph{drift} model which implements \emph{soft} resets mechanism.

Our contributions can be summarized as follows. First, we propose an explicit model for the drift in NN parameters and describe the procedure to estimate the parameters of this model online from the stream of data. Second, we describe how the estimated drift model is incorporated in the learning algorithm. Third, we empirically demonstrate the effectiveness of this approach in preventing the loss of plasticity as well as in an off-policy reinforcement learning setting.

\section{Non-stationary learning with Online SGD}
\label{sec:ogd}

In a non-stationary learning setting with changing data distributions $p_t(x, y)$, where $x \in \mathbb{R}^L,y \in \mathbb{R}^K$, we define the loss function for parameters $\theta \in \mathbb{R}^{D}$ as
\begin{equation}
    \textstyle
    \label{eq:loss_def}
    \mathcal{L}_{t}(\theta) = \mathbb{E}_{(x_t, y_t) \sim p_t} \mathcal{L}_{t}(\theta, x_t, y_t)
\end{equation}
Our goal is to find a parameter sequence $\Theta=(\theta_1,\ldots,\theta_T)$ that minimizes the dynamic regret:
\begin{equation}
    \textstyle
    R_T(\Theta,\Theta^\star) = \frac{1}{T} \sum_{t=1}^{T} \left(\mathcal{L}_{t}(\theta_t) - \mathcal{L}_{t}(\theta^\star_t)\right),
    \label{eq:dynamical_regret}
\end{equation}
with a reference sequence $\Theta^\star=(\theta^\star_{1},\ldots,\theta^\star_{T})$, satisfying $\theta^\star_{t} = \arg\min_{\theta} \mathcal{L}_{t}(\theta)$. A common approach to the online learning problem is online stochastic gradient descent (SGD)~\citep{hazan2023introduction}. Starting from initial parameters $\theta_0$, the method updates these parameters sequentially for each batch of data $\{(x^{i}_t, y^{i}_{t})\}_{i=1}^{B}$ s.t. $(x^{i}_t, y^{i}_{t}) \sim p_t(x_t, y_t)$. The update rule is:
\begin{equation}
    \textstyle
    \label{eq:sgd_main}
    \theta_{t+1} = \theta_{t} - \alpha_{t} \nabla_{\theta} \mathcal{L}_{t+1}(\theta_t),
\end{equation}
where $\nabla_{\theta} \mathcal{L}_{t+1}(\theta_t) = \frac{1}{B} \sum_{i=1}^{B} \nabla_{\theta} \mathcal{L}_{t+1}(\theta_t, x^i_t, y^i_t)$ and $\alpha_t$ is learning rate. See also Appendix~\ref{app:proximal_sgd} for the connection of SGD to proximal optimization.

\textbf{Convex Setting.} In the convex setting, online SGD with a fixed learning rate \( \alpha \) can handle non-stationarity~\citep{zinkevich2003}. By selecting \( \alpha \) appropriately -- potentially using additional knowledge about the reference sequence—we can optimize the dynamic regret in \eqref{eq:dynamical_regret}. In general, algorithms that adapt to the observed level of non-stationarity can outperform standard online SGD. For example, in~\citep{khodak2019adaptive}, the authors propose to adjust the learning rate \( \alpha_t \), while in~\citep{hall2013dynamical} and in~\citep{khodak2019adaptive}, the authors suggest modifying the starting point of SGD from \( \theta_t \) to an adjusted \( \theta_t' \) proportional to the level of non-stationarity.

\textbf{Non-Convex Setting.} Non-stationary learning with NNs is more complex, since now there is a changing set of local minima as the data distribution changes. Such changes can lead to a loss of plasticity and other pathologies. Alternative optimization methods like Adam~\citep{kingma2017adam},
do not fully resolve this issue~\citep[]{dohare2022continual, lyle2022understanding, abbas2023loss, nikishin2023deep, kumar2023maintaining}. Parameter resets~\citep{dohare2022continual, sokar2023dormantneuronphenomenondeep, Dohare2024LossOP} partially mitigate the problem, but could be too aggressive if the data distributions are similar.

\section{Online non-stationary learning with learned parameter resets}
\label{sec:our_method}

\paragraph{Notation.} We denote by $\mathcal{N}(\theta; \mu, \sigma^2)$ a Gaussian distribution on $\theta$ with mean $\mu$ and variance $\sigma^2$. We denote $\theta^i$ the $i$-the component of the vector $\theta=(\theta^1,\ldots,\theta^D)$. Unless explicitly mentioned, we assume distributions are defined per NN parameter and we omit the index $i$. We denote as $\mathcal{L}_{t+1}(\theta) = -\log p(y_{t+1} | x_{t+1}, \theta)$ the negative log likelihood on $(y_{t+1},x_{t+1})$ for parameters $\theta$.

We introduce \emph{Soft Resets}, an approach that enhances learning algorithms on non-stationary data distributions and prevents plasticity loss. The main idea is to assume that the data is generated in non-i.i.d. fashion such that a change in the data distribution is modeled by the \emph{drift} in the parameters $p(\theta_{t+1} | \theta_t, \gamma_t)$ at every time $t+1$ before new data is observed. We assume a class of \emph{drift} models $p(\theta_{t+1} | \theta_t, \gamma_t)$ which encourages the parameters to move closer to the initialization. The amount of drift (and level of non-stationarity) is controlled by $\gamma_t$ which are estimated online from the data.

As can be seen below, in the context of SGD, this approach adjusts the starting point $\theta_t$ of the update to a point $\tilde{\theta}_t(\gamma_t)$, which is closer to the initialization and increases the learning rate proportionally to the drift. In the context of Bayesian inference, this approach shrinks the mean of the estimated posterior towards the prior and increases the variance proportional to $\gamma_t$.
This approach is inspired by prior work in online convex optimization for non-stationary environments~\citep[e.g.,][]{HeWa98, hall2013dynamical, CeLu06, GySz16, khodak2019adaptive}.

\subsection{Toy illustration of the advantage of drift models}
\label{sec:bayesian_intuition}

\begin{figure}[t]
    \centering
    \includegraphics[width=0.9\textwidth]{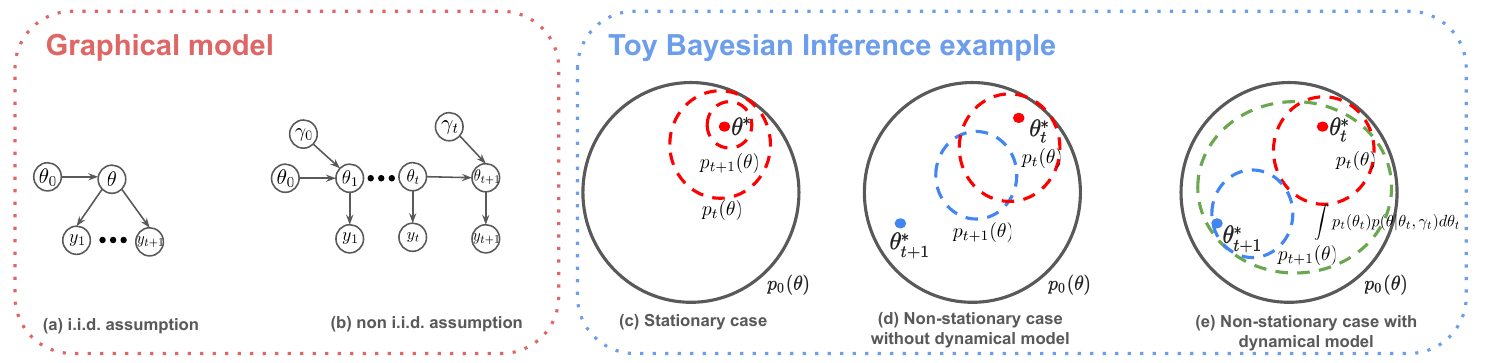}
    \caption{\textbf{Left}: graphical model for data generating process in the (a) stationary case and (b) non-stationary case with drift model $p(\theta_{t+1} | \theta_t, \gamma_t)$. 
    \textbf{Right}: (c) In a stationary online learning regime, the Bayesian  posterior (\textcolor{red}{red} dashed circles) in the long run will concentrate around $\theta^*$ (\textcolor{red}{red} dot). (d) In a non-stationary regime where the optimal parameters suddenly change from current value $\theta_t^*$ to  new value $\theta^*_{t+1}$ (\textcolor{blue}{blue} dot) online Bayesian estimation can be less data efficient and take time to recover when the change-point occurs. (e)
    The use of
    $p(\theta|\theta_t, \gamma_t)$
    and the estimation of $\gamma_t$
    allows to increase the uncertainty, by soft resetting the posterior to make it closer to the prior (\textcolor{Green}{green} dashed circle), so that the updated Bayesian posterior $p_{t+1}(\theta)$ (\textcolor{blue}{blue} dashed circle) can faster track $\theta_{t+1}^*$.}
    \label{fig:idea}
\end{figure}

Consider online Bayesian inference with 2-D observations ${y_t = \theta^\star + \epsilon_t}$
, where $\theta^\star \in \mathbb{R}^2$ are unknown \emph{true} parameters and $\epsilon_t\sim\mathcal{N}(0;\sigma^2 I)$ is Gaussian noise with variance $\sigma^2$. Starting from a Gaussian prior $p_0(\theta)=\mathcal{N}(\theta; \mu_0; \Sigma_0)$, the posterior distribution $p_{t+1}(\theta) = p(\theta | y_1,\ldots,y_t)=\mathcal{N}(\theta; \mu_{t+1}, \Sigma_{t+1})$ is updated using Bayes' rule
\begin{equation}
    \textstyle
    p_{t+1}(\theta) \sim p(y_{t+1}|\theta) p_{t}(\theta).
    \label{eq:online_bayesian_update}
\end{equation}
The posterior update~\eqref{eq:online_bayesian_update} comes from the i.i.d. assumption on the data generation process (Figure~\ref{fig:idea}a), since $p_{t+1}(\theta) \sim p_0(\theta) \prod_{s=1}^{t+1} p(y_{s}|\theta)$. By the Central Limit Theorem (CLT), the posterior mean $\mu_{t}$ converges to $\theta^\star$ and the covariance matrix $\Sigma_t$ shrinks to zero (the radius of red circle in Figure~\ref{fig:idea}c).

Suppose now that the \emph{true} parameters $\theta^\star_t$ (kept fixed before $t$) change to new parameters $\theta^\star_{t+1}$ at time $t+1$. The i.i.d. assumption (Figure~\ref{fig:idea}a) is violated and the update~\eqref{eq:online_bayesian_update} becomes problematic because the low uncertainty (small radius of red dashed circle in Figure~\ref{fig:idea}d) in $p_t(\theta)$ causes the posterior $p_{t+1}(\theta)$ (see blue circle) to adjust slowly towards $\theta^\star_{t+1}$ (blue dot) as illustrated in Figure~\ref{fig:idea}d.

To address this issue, we assume that before observing new data, the parameters \emph{drift} according to $p(\theta_{t+1} | \theta_t, \gamma_t)$ where the amount of drift is controlled by $\gamma_t$. The corresponding conditional independence structure is shown in Figure~\ref{fig:idea}b. The posterior update then becomes:
\begin{equation}
    \textstyle
    p_{t+1}(\theta) \sim p(y_{t+1}|\theta) \int p(\theta | \theta'_t, \gamma_t) p_{t}(\theta'_t) d\theta'_t.
    \label{eq:online_bayesian_update_drifted}
\end{equation}
For a suitable choice of \emph{drift} model $p(\theta_{t+1} | \theta_t, \gamma_t)$, this modification allows $p_{t+1}(\theta)$ (blue circle) to adjust more rapidly towards the new $\theta^\star_{t+1}$ (blue dot), see Figure~\ref{fig:idea}e. This is because the new prior $\int p(\theta | \theta'_t, \gamma_t) p_{t}(\theta'_t) d\theta'_t$ has larger variance (green circle) than $p_t(\theta)$ and its mean is closer to the center of the circle. Ideally, the parameter $\gamma_t$ should capture the underlying non-stationarity in the data distribution in order to control the impact of the prior $\int p(\theta | \theta'_t, \gamma_t) p_{t}(\theta'_t) d\theta'_t$. For example, if at some point the non-stationarity disappears, we want the drift model to exhibit no-drift to recover the posterior update~\eqref{eq:online_bayesian_update}. This highlights the importance of the adaptive nature of the drift model.


\subsection{Ornstein-Uhlenbeck parameter drift model}
\label{sec:ou_sec}

We motivate the specific choice of the drift model which is useful for maintaining plasticity. We assume that our Neural Network has enough capacity to learn any \emph{stationary} dataset in a fixed number of iterations starting from a good initialization $\theta_0 \sim p_0(\theta)$ \citep[see, e.g., ][]{he2015delving,pmlr-v9-glorot10a}. Informally, we call the initialization $\theta_0$ \emph{plastic} and the region around $\theta_0$ a \emph{plastic region}.

Consider now a piecewise stationary datastream
that switches between a distribution $p_a$, with a set of local minima $\mathcal{M}_a$ of the negative likelihood $\mathcal{L}(\theta)$, to a distribution $p_b$ at time $t+1$, with a set of local minima $M_b$.  If $M_b$ is far from $M_a$, then \emph{hard reset} might be beneficial, but if $M_b$ is close to $M_a$, resetting parameters is suboptimal. Furthermore,
since $\theta$ is high-dimensional, different dimensions
might need to be treated differently. We want a drift
model that can capture all of these scenarios.

\textbf{Drift model.} The drift model $p(\theta_{t+1} | \theta_t, \gamma_t)$ which exhibits the above properties is given by
\begin{align}
    \textstyle
    \label{eq:ou_model}
    p(\theta | \theta_{t}, \gamma_{t}) = \normal(\theta; \gamma_{t} \theta_{t} + (1-\gamma_{t}) \mu_{0}; (1 - \gamma_{t}^2) \sigma_{0}^2),
\end{align}
which is separately defined for every parameter dimension $\theta^i$ where $p_0(\theta^i_{0}) \sim \normal(\theta^i_{0}; \mu^i_{0}; \left[\sigma^i_{0}\right]^2)$ is the per-parameter prior distribution and $\gamma_t=(\gamma^1_t,\ldots,\gamma^D_t)$. The model is a discretized Ornstein-Uhlenbeck (OU) process~\citep{ou_process} (see Appendix~\ref{app:ou_process} for the derivation).

The parameter $\gamma_t \in [0, 1]$ is a \emph{drift} parameter and controls the amount of non-stationarity in each parameter. For $\gamma_t=1$, there is no drift
and
for $\gamma_t = 0$, the drift model reverts the parameters back to the prior. A value of $\gamma_t \in (0, 1)$ interpolates between these two extremities. A remarkable property of~\eqref{eq:ou_model} is that starting from the current parameter $\theta_t$, if we simulate a long trajectory, as $T \rightarrow \infty$, the distribution of $p(\theta_{T} | \theta_t)$ will converge to the prior $p(\theta_0)$. This is only satisfied (for $\gamma_t \in (0, 1)$) due to the variance $\sigma^2_0(1 - \gamma_t^2)$. Replacing it by an arbitrary variance $\sigma^2$ would result in the variance of $p(\theta_{T} | \theta_t)$ either going to $0$ or growing to $\infty$, harming learning. Thus, the model \eqref{eq:ou_model} encourages parameters to move towards \emph{plastic} region (initialization). In Appendix~\ref{sec:other_models}, we discuss this further and other potential choices for the drift model.

\subsection{Online estimation of drift model}
\label{sec:estimate_drift}

The drift model $p(\theta_{t+1} | \theta_{t}, \gamma_{t})$ quantifies \emph{prior belief} about the change in parameters before seeing new data. A suitable choice of an objective to select $\gamma_t$ is \emph{predictive likelihood} which quantifies the probability of new data under our current parameters and drift model. From Bayesian perspective, it means selecting the prior distribution which explains the future data the best.

We derive the drift estimation procedure in the context of \emph{approximate} online variational inference~\citep{Broderick2013} with Bayesian Neural Networks (BNN). Let $\Gamma_t=(\gamma_1, \ldots, \gamma_t)$ be the history of observed parameters of the drift model and $\mathcal{S}_{t}=\{(x_{1},y_{1}),\ldots,(x_t,y_t)\}$ be the history of observed data. The objective of \emph{approximate} online variational inference is to propagate an \emph{approximate} posterior $q_t(\theta | \mathcal{S}_{t}, \Gamma_{t-1})$ over parameters, such that it is constrained to some family $\mathcal{Q}$ of probability distributions. In the context of BNNs, it is typical~\citep{Blundell15} to assume a family ${\mathcal{Q} = \{ q(\theta) : q(\theta) \sim \prod_{i=1}^{D} \mathcal{N}(\theta^i; \mu^i, \left[\sigma^{i}\right]^2); \theta = (\theta^1,\ldots,\theta^D)\}}$ of Gaussian mean-field distributions over parameters $\theta \in \mathbb{R}^{D}$ (separate Gaussian per parameter). For simplicity of notation, we omit the index $i$. Let $q_t(\theta) \triangleq q_t(\theta | \mathcal{S}_{t}, \Gamma_{t-1}) \in \mathcal{Q}$ be the Gaussian \emph{approximate} posterior at time $t$ with mean $\mu_t$ and variance $\sigma^2_t$ for every parameter. The new approximate posterior $q_{t+1}(\theta) \in \mathcal{Q}$ is found by
\begin{equation}
    \textstyle
    q_{t+1}(\theta) = \arg\min_{q} \mathbb{KL} \left[q(\theta) || p(y_{t+1} | x_{t+1}, \theta) q_t(\theta |\gamma_t) \right],
    \label{eq:variation_bayes_update}
\end{equation}
where the prior term is the approximate predictive look-ahead prior given by
\begin{equation}
    \textstyle
    q_t(\theta |\gamma_t) = \int q_t(\theta_{t}) p(\theta | \theta_{t}, \gamma_{t}) d \theta_{t} = \mathcal{N}(\theta; \tilde{\mu}_t(\gamma_t), \tilde{\sigma}^2_t(\gamma_t))
    \label{eq:approx_predictive_prior}
\end{equation}
that has parameters $\tilde{\mu}_t(\gamma_t) = \gamma_t \mu_t + (1-\gamma_t) \mu_0, \tilde{\sigma}^2_t(\gamma_t) = \gamma_t^2 \sigma^2_t + (1-\gamma_t^2) \sigma^2_0$, see Appendix~\ref{sec:bnns} for derivation. The form of this prior $q_t(\theta|\gamma_t)$ comes from the non i.i.d. assumption (see Figure~\ref{fig:idea}b) and the form of the drift model~\eqref{eq:ou_model}. For new batch of data $(x_{t+1}, y_{t+1})$ at time $t+1$, the \emph{approximate predictive log-likelihood} equals to
\begin{equation}
    \textstyle
    \log q_t(y_{t+1} | x_{t+1}, \gamma_t) = \log \int p(y_{t+1} | x_{t+1}, \theta) q_t(\theta | \gamma_{t}) d \theta.
    \label{eq:approximate_predictive_likelihood}
\end{equation}
The log-likelihood ~\eqref{eq:approximate_predictive_likelihood} allows us to quantify predictions on new data $(x_{t+1}, y_{t+1})$ given our current distribution $q_t(\theta)$ and the drift model from \eqref{eq:ou_model}. We want to find such $\gamma^\star_{t}$ that
\begin{equation}
    \textstyle
    \gamma^\star_{t} \approx \arg\max_{\gamma_t} \log q_t(y_{t+1} | x_{t+1},  \gamma_{t})
    \label{eq:goal_for_gamma}
\end{equation}
Using $\gamma^\star_{t}$ in~\eqref{eq:ou_model} modifies the prior distribution~\eqref{eq:approx_predictive_prior} to fit the most recent observations the best by putting more mass on the region where the new parameter could be found (see Figure~\ref{fig:idea},right).

\textbf{Gradient-based optimization for $\gamma_t$.} The approximate predictive prior in \eqref{eq:approx_predictive_prior} is Gaussian which allows us to use the so-called reparameterisation trick to optimize~\eqref{eq:approximate_predictive_likelihood} via gradient descent. Starting from an initial value of drift parameter $\gamma^0_{t}$ at time $t$, we perform $K$ updates with learning rate $\eta_{\gamma}$
\begin{equation}
    \textstyle
    \label{eq:drift_via_predictive_ll}
    \gamma_{t,k+1} = \gamma_{t,k} + \eta_{\gamma} \nabla_{\gamma} \log \int p(y_{t+1} | x_{t+1}, \tilde{\mu}_t(\gamma_{t,k}) + \epsilon \tilde{\sigma}_{t}(\gamma_{t,k})) \mathcal{N}(\epsilon; 0, I) d\epsilon,
\end{equation}
The integral is evaluated by Monte-Carlo (MC) using $M$ samples $\epsilon_{i} \sim \mathcal{N}(\epsilon; 0, I)$, $i=1,\ldots,M$
\begin{equation}
    \textstyle
    \label{eq:monte_carlo}
    \int p(y_{t+1} | x_{t+1}, \tilde{\mu}_t(\gamma_{t,k}) + \epsilon \tilde{\sigma}_{t}(\gamma_{t,k})) \mathcal{N}(\epsilon; 0, I) d\epsilon \approx \frac{1}{M} \sum_{i=1}^{M} p(y_{t+1} | x_{t+1}, \tilde{\mu}_t(\gamma_{t,k}) + \epsilon_{i} \tilde{\sigma}_{t}(\gamma_{t,k}))
\end{equation}
Inductive bias in the drift model is captured by $\gamma_t^0$, where $\gamma_{t,0} = 1$ encourages stationarity, while $\gamma_{t,0} = \gamma_{t-1,K}$ promotes temporal smoothness. In practice, we found $\gamma_{t,0}=1$ was the most effective.

\textbf{Structure in the drift model.} The drift model can be defined to be shared across different \emph{subsets} of parameters which reduces the expressivity of the drift model but also provides regularization to~\eqref{eq:drift_via_predictive_ll}. We consider $\gamma_t$ to be either defined for each \emph{parameter} or for each \emph{layer}. See Section~\ref{sec:experiments} for details as well as corresponding results in Appendix~\ref{app:bayesian_is_better}.

\textbf{Interpretation of $\gamma_t$.} By linearising $\log p(y_{t+1}| x_{t+1}, \theta)$ around $\mu_t$, we can compute~\eqref{eq:approximate_predictive_likelihood} in a closed form and get the following loss for $\gamma_t$ (see Appendix~\ref{app:linearisation} for the proof) optimizing~\eqref{eq:goal_for_gamma}
\begin{equation}
    \textstyle
    \label{eq:closed_form_gradient_steps}
    \mathcal{F}(\gamma_t) = 0.5 (\sigma^2_{t}(\gamma_t) \odot g_{t+1})^\top g_{t+1} - (\gamma_t \odot \mu_t + (1-\gamma_t) \odot \mu_0 )^\top g_{t+1},
\end{equation}
where $g_t = \nabla \mathcal{L}_{t+1}(\mu_t)$ and $\mathcal{L}_{t+1}(\theta) = -\log p(y_{t+1}| x_{t+1}, \mu_t)$, and where $\odot$ denotes element-wise product performed only over parameters for which $\gamma_t$ is shared (see paragraph about structure in drift model). The transpose operation is also defined on a subset of parameters for which $\gamma_t$ is shared. Adding the $\ell_2$ penalty $\frac{1}{2} \lambda ||\gamma_t - \gamma^0_t||^2$ encoding the starting point $\gamma^0_t$, gives us the closed form for $\gamma_t$
\begin{equation}
    \textstyle
    \label{eq:closed_form_gamma}
    \gamma_t = \frac{(\mu_t - \mu_0)^T g_{t+1} + \lambda \gamma^0_t}{((\sigma^2_0 - \sigma^2_t) \cdot g_{t+1})^T g_{t+1}  + \lambda},
\end{equation}
where we also clip parameters $\gamma_t$ to $[0, 1]$. The expression~\eqref{eq:closed_form_gamma} gives us the geometric interpretation for $\gamma_t$. The value of $\gamma_t$ depends on the angle between $(\mu_t - \mu_0)$ and $g_{t+1}$ When these vectors are aligned, $\gamma_t$ is high and is low otherwise. When these vectors are orthogonal or the gradient $g_{t+1} \approx 0$, the value of $\gamma_t$ is heavily influenced by $\gamma^0_t$. Moreover, when $g_{t+1} \approx 0$, we can interpret it as being close to a local minimum, i.e., stationary, which means that we want $\gamma_t \approx 1$, therefore adding the $\ell_2$ penalty is important. Also, when the norm of the gradients $g_{t+1}$ is high, the value of $\gamma_t$ is encouraged to decrease, introducing the drift. This means that using $\gamma_t$ in the parameter update (see Section~\ref{sec:map_inference}) encourages the norm of the gradient to stay small. In practice, we found that update~\eqref{eq:closed_form_gamma} was unstable suggesting that linearization of the log-likelihood might not be a good approximation for learning $\gamma_t$.

\subsection{Approximate Bayesian update of posterior $q_t(\theta)$ with BNNs}
\label{sec:bayesian_inference}

The optimization problem~\eqref{eq:variation_bayes_update} for the per-parameter Gaussian $q(\theta)=\mathcal{N}(\theta; \mu, \sigma^2)$ with Gaussian prior $q_t(\theta) = \mathcal{N}(\theta; \mu_t, \sigma^2_t)$, both defined for every parameter of NN, can be written (see Appendix~\ref{sec:bnns}) to minimize the following loss
\begin{equation}
    \textstyle
    \tilde{\mathcal{F}}_{t}(\mu, \sigma, \gamma_t) = \mathbb{E}_{\epsilon \sim \mathcal{N}(0; I)}\left[\mathcal{L}_{t+1}(\mu + \epsilon \sigma) \right] + \sum_{i=1}^{D} \textcolor{red}{\lambda^i_{t}} \left[ \frac{(\mu^i - \tilde{\mu}^i_{t}(\gamma_t))^2 + \left[\sigma^i\right]^2}{2 \left[\tilde{\sigma}^i_{t}(\gamma_t)\right]^2} - \frac{1}{2}\log \left[\sigma^i\right]^2 \right],
    \label{eq:elbo_gaussian_expanded_temp}
\end{equation}
where $\lambda^i_{t} > 0$ are per-parameter temperature coefficients. The use of small temperature $\lambda > 0$ parameter (shared for all NN parameters) was shown to improve empirical performance of Bayesian Neural Networks \citep{wenzel2020good}. Given that in~\eqref{eq:elbo_gaussian_expanded_temp}, the variance $\tilde{\sigma}^2_{t}(\gamma_t)$ can be small, in order to control the strength of the regularization, we propose to use per parameter temperature $\textcolor{red}{\lambda^i_{t}} = \textcolor{blue}{\lambda} \times \left[\sigma^i_{t}\right]^2$, where $\lambda > 0$ is a global constant. This leads to the following objective
\begin{equation}
    \textstyle
    \hat{\mathcal{F}}_{t}(\mu, \sigma, \gamma_t) = \mathbb{E}_{\epsilon \sim \mathcal{N}(0; I)}\left[\mathcal{L}_{t+1}(\mu + \epsilon \sigma) \right] + \frac{\textcolor{blue}{\lambda}}{2} \sum_i \textcolor{red}{r^i_{t}} \left[ (\mu_i - \tilde{\mu}^i_{t}(\gamma_t))^2 + [\sigma^i]^2 - \left[\tilde{\sigma}^i_{t}(\gamma_t)\right]^2 \log [\sigma^i]^2 \right],
    \label{eq:elbo_gaussian_expanded_efficient_temperature}
\end{equation}
where the quantity $\textcolor{red}{r^i_{t}} = [\sigma^i_{t}]^2 / [\sigma^i_{t}(\gamma_t)]^2$ is a relative change in the posterior variance due to the drift. The ratio $r^i_{t}=1$ when $\gamma_t=1$. For $\gamma_{t} < 1$ since typically $\sigma^2_{t} < \sigma^2_{0}$, the ratio is $r^i_{t} < 1$. Thus, as long as there is non-stationarity ($\gamma_{t} < 1$), the objective~\eqref{eq:elbo_gaussian_expanded_efficient_temperature} favors the data term $\mathbb{E}_{\epsilon \sim \mathcal{N}(0; I)}\left[\mathcal{L}_{t+1}(\mu + \epsilon \sigma) \right]$ allowing the optimization to respond faster to changes in the data distribution. To find new parameters, let $\mu_{t+1,0}=\tilde{\mu}_t(\gamma_t)$ and $\sigma_{t+1,0} = \tilde{\sigma}_t(\gamma_t)$, and perform updates $K$ on~\eqref{eq:elbo_gaussian_expanded_efficient_temperature}
\begin{equation}
    \textstyle
    \label{eq:bayesian_updates}
    \mu_{t+1,k+1} = \mu_{t+1,k} - \alpha_{\mu} \hat{\mathcal{F}}_t(\mu_{t+1,k}, \sigma_{t+1,k}, \gamma_t), \ \  \sigma_{t+1,k+1} = \sigma_{t+1,k} - \alpha_{\sigma} \hat{\mathcal{F}}_t(\mu_{t+1,k}, \sigma_{t+1,k}, \gamma_t),
\end{equation}
where $\alpha_{\mu}$ and $\alpha_{\sigma}$ are learning rates for the mean and for the standard deviation correspondingly. All derivations are provided in Appendix~\ref{sec:bnns}. The full procedure is described in Algorithm~\ref{alg:alg_bayesian}.

\subsection{Fast MAP update of posterior $q_t(\theta)$}
\label{sec:map_inference}

As a faster alternative to propagating the posterior~\eqref{eq:variation_bayes_update}, we do MAP updates with the prior $p_0(\theta) = \mathcal{N}(\theta; \mu_0; \sigma^2_0)$ and the approximate posterior $q_t(\theta) = \mathcal{N}(\theta; \theta_t; \sigma^2_t=s^2 \sigma^2_0)$, where $s \leq 1$ is a hyperparameter controlling the variance $\sigma^2_t$ of
$q_t(\theta)$. Since a fixed $s$ may not capture the true parameters variance, using a Bayesian method (see Section~\ref{sec:bayesian_inference}) is preferred but comes at a high computational cost (see Appendix~\ref{app:computational_complexity} for discussion). The MAP update is given by (see Appendix~\ref{app:map_inference} for derivations) finding a minimum of the following proximal objective
\begin{equation}
    \textstyle
     G(\theta) = \mathcal{L}_{t+1}(\theta) + \frac{1}{2} \sum_{i=1}^{D} \frac{|\theta^i - \tilde{\theta}^i_{t}(\gamma_t)|^2}{\tilde{\alpha}^i_t(\gamma_t)}
    \label{eq:proximal_objective}
\end{equation}
where the regularization target for the parameter dimension $i$ is given by
\begin{equation}
    \textstyle
    \tilde{\theta}^i_{t}(\gamma_t) = \gamma^i_t \theta^i_t + (1-\gamma^i_t)\mu^i_0
    \label{eq:adapted_reg}
\end{equation}
and the per-parameter learning rate is given as (assuming that $\alpha_t$ the base SGD learning rate)
\begin{equation}
    \textstyle
    \tilde{\alpha}^i_t(\gamma_t) = \alpha_t \left( (\gamma^i_t)^2 + \frac{1 - (\gamma^i_t)^2}{s^2} \right).
    \label{eq:adapted_lr}
\end{equation}
Linearising $\mathcal{L}_{t+1}(\theta)$ around $\tilde{\theta}_t(\gamma_t)$ and optimizing~\eqref{eq:proximal_objective} for $\theta$ leads to (see Appendix~\ref{app:map_inference})
\begin{equation}
    \textstyle
    \label{eq:linearised_map}
    \theta_{t+1} = \tilde{\theta}_t(\gamma_t) - \tilde{\alpha}_t(\gamma_t) \circ \nabla_{\theta} \mathcal{L}_{t+1}(\tilde{\theta}_t(\gamma_t)),
\end{equation}
where $\circ$ is element-wise multiplication. For $\gamma_t=1$, we recover the ordinary SGD update, while the values $\gamma_t < 1$ move the starting point of the modified SGD closer to the initialization as well as increase the learning rate. Algorithm~\ref{alg:soft_reset} describes the full procedure. In Appendix~\ref{app:practical} we describe additional practical choices made for the \emph{Soft Resets} algorithm. Similarly to the Bayesian approach~\eqref{eq:bayesian_updates}, we can do multiple updates on~\eqref{eq:proximal_objective}. We describe this \emph{Soft Resets Proximal} algorithm in Appendix~\ref{app:map_inference} and full procedure is given in Algorithm~\ref{alg:alg_proximal}.

\begin{algorithm}[tb]
   \caption{\emph{Soft-Reset} algoritm}
   \label{alg:soft_reset}
\begin{algorithmic}
   \STATE {\bfseries Input:} Data-stream $\mathcal{S}_{T}=\{(x_{t}, y_{t})\})_{t=1}^{T}$
   \STATE Neural Network (NN) initializing distribution $p_{init}(\theta)$ and specific initialization $\theta_0 \sim p_{init}(\theta)$
   \STATE Learning rate $\alpha_t$ for parameters and $\eta_{\gamma}$ for drift parameters
   \STATE Number of gradient updates $K_{\gamma}$ on drift parameter $\gamma_t$
   \STATE NN initial standard deviation (STD) scaling $p \leq 1$ (see~\eqref{eq_app:modified_prior}) and ratio $s = \frac{\sigma_t}{p\sigma_0}$.
   \FOR{step $t=0,1,2,\ldots,T$}
   \STATE For $(x_{t+1}, y_{t+1})$, predict $\hat{y}_{t+1} = f(x_{t+1}| \theta_{t})$
   \STATE Compute performance metric based on $(y_{t+1}, \hat{y}_{t+1})$
   \STATE Initialize drift parameter $\gamma_{t,0} = 1$
   \FOR{step $k=0,1,2,\ldots,K_{\gamma}$}
    \STATE Sample $\theta'_0 \sim p_{init}(\theta)$
    \STATE Stochastic update~\eqref{eq_app:stochastic_approx} on drift parameter using specific initialization~\eqref{eq_app:drift_from_specific_init}
    \STATE $\gamma_{t,k+1} = \gamma_{t,k} + \eta_{\gamma} \nabla_{\gamma} \left[\log p(y_{t+1} | x_{t+1}, \gamma_t \theta_t + (1-\gamma_t)\theta_0 + \theta'_0 p\sqrt{1-\gamma_t^2 + \gamma_t^2 s^2}) \right]_{\gamma_t = \gamma_{t,k}}$
   \ENDFOR
   \STATE Get $\tilde{\theta}_t(\gamma_{t,K})$ with \eqref{eq:adapted_reg} and $\tilde{\alpha}_t(\gamma_{t,K})$ with \eqref{eq:adapted_lr}
   \STATE Update parameters $\theta_{t+1} = \tilde{\theta}_t(\gamma_{t,K}) - \tilde{\alpha}_t(\gamma_{t,K}) \circ \nabla_{\theta} \mathcal{L}_{t+1}(\tilde{\theta}_t(\gamma_{t,K}))$
   \ENDFOR
\end{algorithmic}
\end{algorithm}

\section{Related Work}
\label{sec:related_work}

\textbf{Plasticity loss in Neural Networks.}
Our model shares similarities with 
reset-based approaches such as Shrink \& Perturb (S\&P)~\citep{ash2020warmstarting} and L2-Init~\citep{kumar2021implicit}; however, whereas we learn drift parameters from data, these
methods do not, leaving them vulnerable to mismatch between assumed non-stationarity and the actual realized non-stationarity in the data.
Continual Backprop~\citep{dohare2022continual} or ReDO~\citep{sokar23a}
apply resets in a data-dependent fashion, e.g. either based on utility or
whether units are dead. But they use hard resets, and cannot amortize the
cost of removing entire features.
%
Interpretation~\eqref{eq:closed_form_gamma} of $\gamma_t$ connects to the notion of parameters utility from~\cite{elsayed2024addressing},
but this quantity is used to prevent catastrophic forgetting by decreasing learning rate for high $\gamma_t$. Our method increases the learning rate for low $\gamma_t$ to maximize adaptability, and is not designed to prevent catastrophic forgetting.

\textbf{Non-stationarity.} Non-stationarity arises naturally in a variety of contexts, the most obvious being continual and reinforcement learning. 
The structure of non-stationarity may vary from problem to problem.
At one extreme, we have a \emph{piece-wise stationary} setting,
for example a change in the location of a camera generating a stream of images, or a hard update to the learner's target network in value-based deep RL algorithms.
This setting has been studied extensively due to its propensity to induce \emph{catastrophic forgetting}~\citep[e.g.][]{Kirkpatrick_2017,PARISI201954,van2019three,Chen2018Lifelong} and \emph{plasticity loss} ~\citep{dohare2022continual,lyle2023understanding,lyle2024disentanglingcausesplasticityloss,kumar2023maintaining}. At the other extreme, we can consider more gradual changes, for example due to improvements in the policy of an RL agent~\citep{mnih2013playing,schulman2017proximal,nikishin2023deep,dohare2022continual} or shifts in the data generating process ~\citep{lin2021clear,zhai2023online, HADSELL20201028, verwimp2024continual}. Further, these scenarios might be combined, for example in \emph{continual reinforcement learning}~\citep{Kirkpatrick_2017,abbas2023loss,dohare2022continual} where the reward function or transition dynamics could change over time.

\textbf{Non-stationary online convex optimization.} Non-stationary prediction has a long history in online convex optimization, where several algorithms have been developed to adapt to changing data \citep[see, e.g., ][]{HeWa98,CeLu06,HaSe09,GyLiLu12,hall2013dynamical,GySz16,khodak2019adaptive}.
Our approach takes an inspiration from these works by employing a drift model as, e.g., \citep{HeWa98,hall2013dynamical} and by changing learning rate as \citep{khodak2019adaptive,metagrad}.
%
Further, our OU drift model bears many similarities to the implicit drift model introduced in the update rule of \citep{HeWa98} (see also \citep{CeLu06,GyLiLu12}), 
where the predictive distribution is mixed with a uniform distribution to ensure the prediction could change quickly enough if the data changes significantly, where in our case $p_0$ plays the same role as the uniform distribution.

\textbf{Bayesian approaches to non-stationary learning.}
A standard approach is Variational Continual Learning~\citep{variationalCL2018},
which focuses on preventing catastrophic forgetting
and is an online version of
“Bayes By Backprop”~\citep{Blundell15}.
This method does not incorporate dynamical parameter drift components. In~\citep{Kurle2020Continual}, the authors applied variational inference (VI) on non-stationary data, using the OU-process and Bayesian forgetting, but unlike in our approach, their drift parameter is not learned.
Further, in~\citep{titsias2023kalman}, the authors considered an OU parameter drift model similar to ours, with an adaptable drift scalar $\gamma$ and analytic Kalman filter updates, but is applied over the final layer weights only, while the remaining  weights of the network
were estimated by online SGD. In~\citep{Jones2024HumanlikeLI}, the authors propose to deal with non-stationarity by assuming that each parameter is a finite sum of random variables following different OU process. They derive VI updates on the posterior of these variables. Compared to this work, we learn drift parameters for every NN parameter rather than assuming a finite set of drift parameters. A different line of research assumes that the drift model is known and use different techniques to estimate the hidden state (the parameters) from the data: in~\citep{chang2023lowrank}, the authors use Extended Kalman Filter to estimate state and in~\citep{bencomo2023implicit}, they propagate the MAP estimate of the hidden state distribution with $K$ gradient updates on a proximal objective similar to~\eqref{eq_app:many_grad_updates}, whereas in Bayesian Online Natural Gradient (BONG)~\citep{jones2024bayesianonlinenaturalgradient}, the authors use natural gradient for the variational parameters.

\section{Experiments}
\label{sec:experiments}

\textbf{Soft reset methods.} There are multiple variations of our method. We call the method implemented by Algorithm~\ref{alg:soft_reset} with $1$ gradient update on the drift parameter \emph{Soft Reset}, while other versions show different parameter choices: \emph{Soft Reset} ($K_{\gamma}=10$) is a version with $10$ updates on the drift parameter, while \emph{Soft Reset} ($K_{\gamma}=10$, $K_{\theta}=10$) is the method of Algorithm~\ref{alg:alg_proximal} in Appendix~\ref{app:map_inference} with $10$ updates on drift parameter, followed by $10$ updates on NN parameters. \emph{Bayesian Soft Reset} ($K_{\gamma}=10$, $K_{\theta}=10$) is a method implemented by Algorithm~\ref{alg:alg_bayesian} with $10$ updates on drift parameter followed by $10$ updates on the mean $\mu_{t}$ and the variance $\sigma^2_t$ (uncertainty) for each NN parameter. Bayesian method performed the best overall but required higher computational complexity (see Appendix~\ref{app:computational_complexity}).  Unless specified, $\gamma_t$ is shared for all the parameters in each layer (separately for weight and biases).

\begin{figure}[tb]
    \centering
    \includegraphics[scale=0.3]{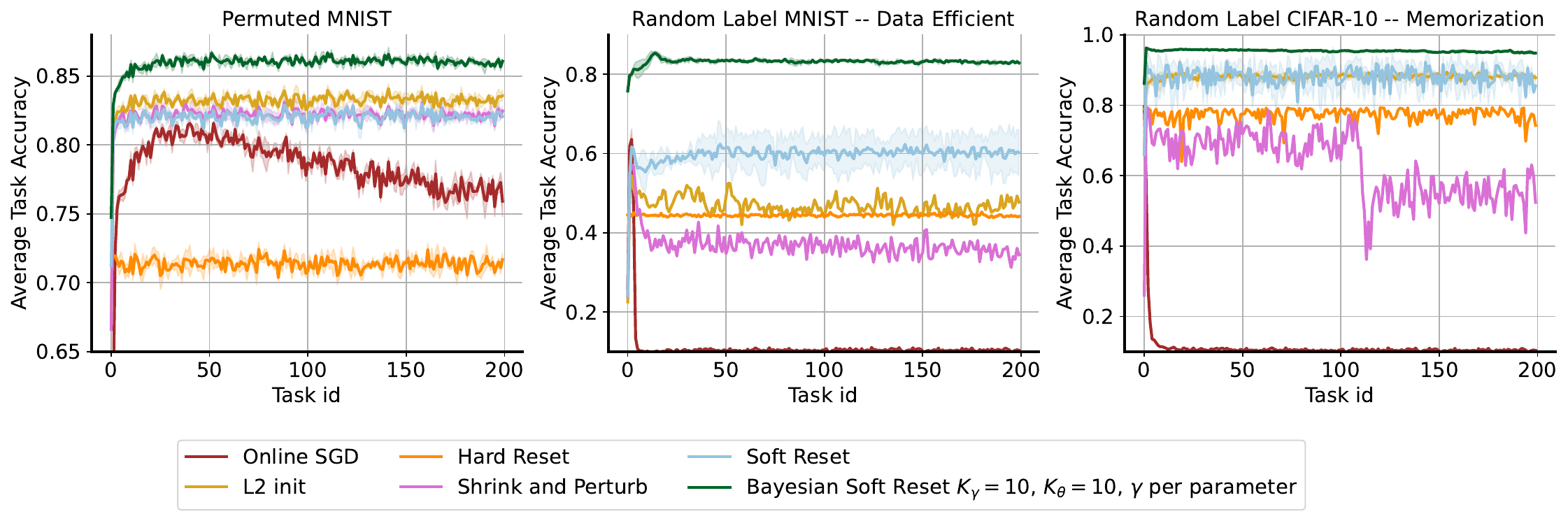}
    \caption{Plasticity benchmarks. \textbf{Left:} performance on \emph{permuted MNIST}. \textbf{Center:} performance on \emph{random-label MNIST} (data efficient). \textbf{Right:} performance on \emph{random-label CIFAR-10} (memorization). The x-axis is the task id and the y-axis is the per-task training accuracy~\eqref{eq:per_task_avg_accuracy}.}
    \label{fig:plasticity_benchmarks}
\end{figure}

\begin{figure}[tb]
    \centering
    \includegraphics[scale=0.3]{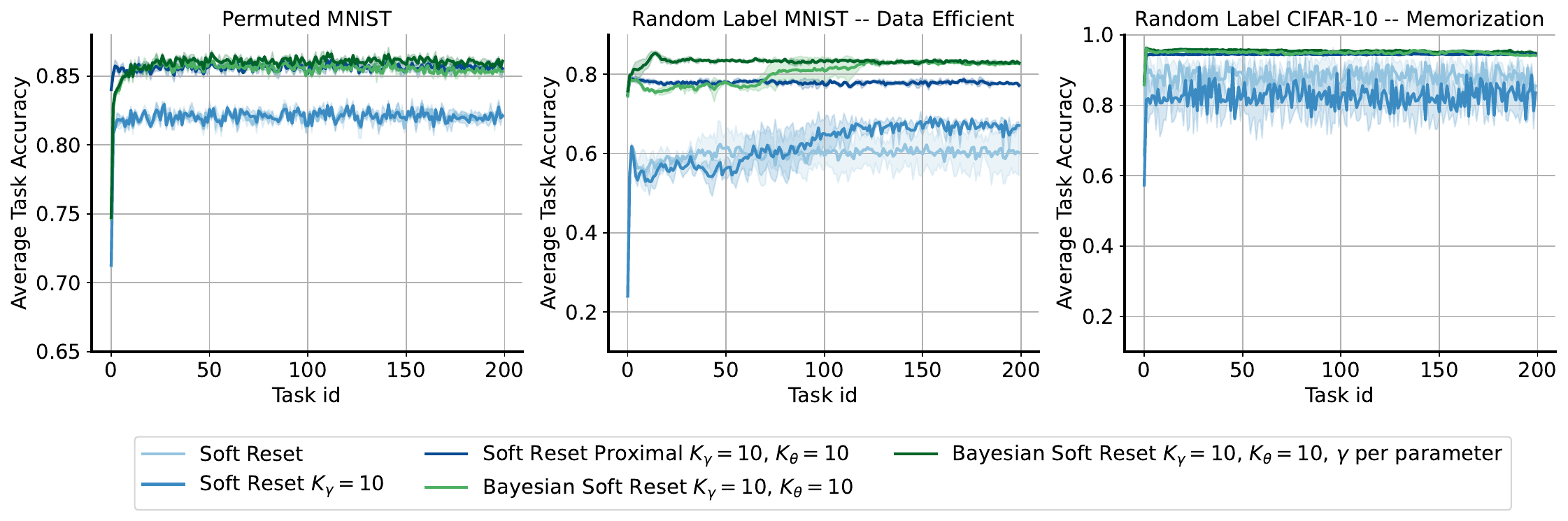}
    \caption{Different variants of \emph{Soft Resets}. \textbf{Left:} performance on \emph{permuted MNIST}. \textbf{Center:} performance on \emph{random-label MNIST} (data efficient). \textbf{Right:} performance on \emph{random-label CIFAR-10} (memorization). The x-axis is the task id and the y-axis is the per-task training accuracy~\eqref{eq:per_task_avg_accuracy}.}
    \label{fig:plasticity_benchmarks_soft_reset_variants}
\end{figure}

\begin{figure}[t]
    \centering
    \includegraphics[scale=0.25]{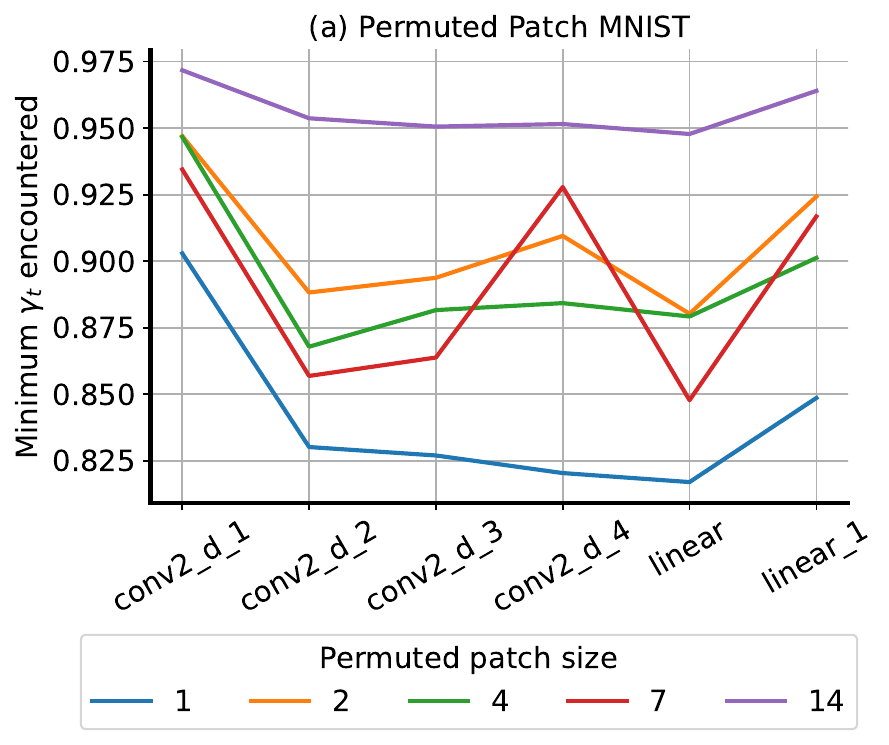}
    \includegraphics[scale=0.2]{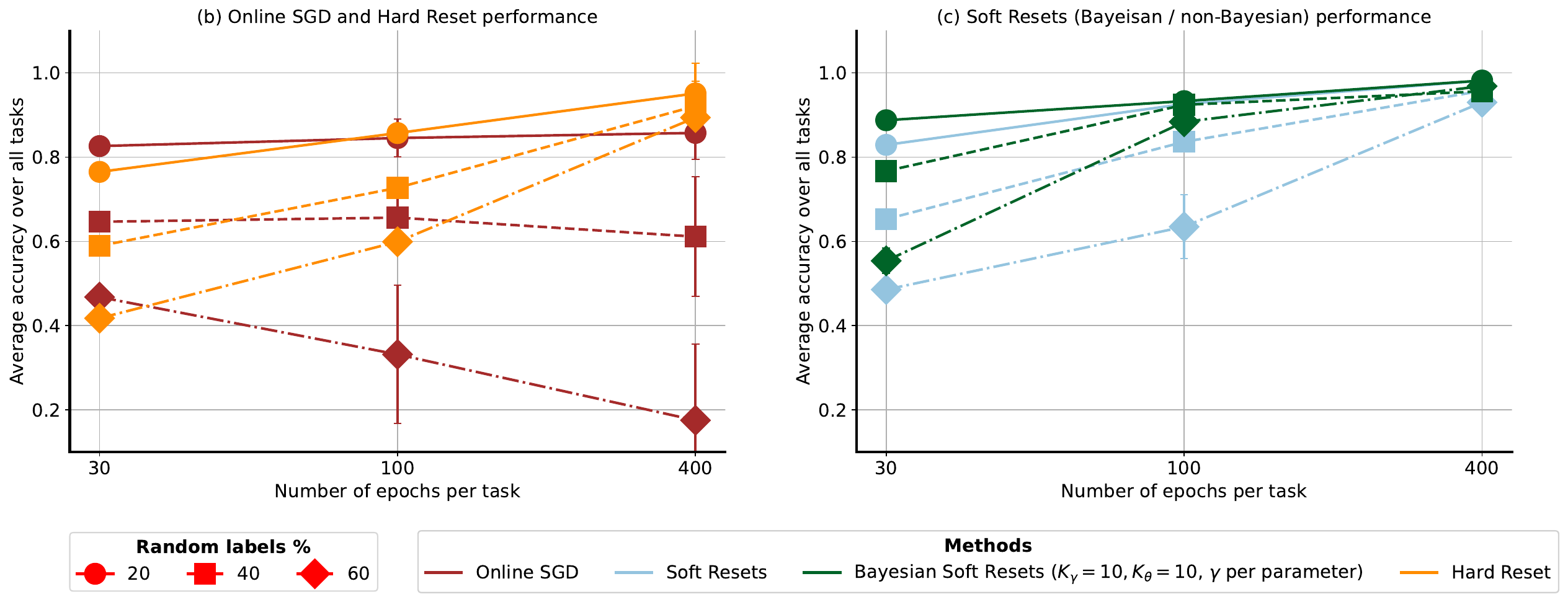}
    \vspace{-6pt}
    \caption{\textbf{(a)} the x-axis denotes the layer, the y-axis denotes the minimum encountered $\gamma_t$ for each convolutional and fully-connected layer when trained on permuted Patches MNIST, color is the patch size. The impact of non-stationarity on performance on random-label MNIST of Online SGD and Hard Reset is shown in \textbf{(b)} while the one of \emph{Soft Resets} is shown in \textbf{(c)}. The x-axis denotes the number of epochs each task lasts, while the marker and line styles denote the percentage of random labels within each task, circle (solid) represents $20\%$, rectangle(dashed) $40\%$, while rhombus (dashed and dot) $60\%$. The y-axis denotes the average performance (over $3$ seeds) on the stream of $200$ tasks.}
    \label{fig:impact_of_randomness}
\end{figure}

\textbf{Loss of plasticity.} We analyze the performance of our method on \emph{plasticity benchmarks}~\citep{kumar2023maintaining,lyle2023understanding,lyle2024disentanglingcausesplasticityloss}. Here, we have a sequence of tasks, where each task consists of a fixed (for all tasks) subset of $10000$ images images from either CIFAR-10~\citep{Krizhevsky2009LearningML} or MNIST, where either pixels are permuted or the label for each image is randomly chosen. Several papers \citep{kumar2023maintaining,lyle2023understanding,lyle2024disentanglingcausesplasticityloss} study a \emph{memorization} random-label setting where \emph{SGD} can perfectly learn each task from scratch. To highlight the data-efficiency of our approach, we  study the \emph{data-efficient} setting where \emph{SGD} achieves only $50\%$ accuracy on each task when trained from scratch. Here, we expect that algorithms taking into account similarity in the data, to perform better. To study the impact of the non-stationarity of the input data, we consider \emph{permuted MNIST} where pixels are randomly permuted within each task (the same task as considered by \citealp{kumar2023maintaining}). As baselines, we use \emph{Online SGD} and \emph{Hard Reset} at task boundaries. We also consider \emph{L2 init}~\citep{kumar2023maintaining}, which adds $L2$ penalty $||\theta-\theta_0||^2$ to the fixed initialization $\theta_0$ as well as  \emph{Shrink\&Perturb}~\citep{ash2020warmstarting}, which multiplies each parameter by a scalar $\lambda \leq 1$ and adds random Gaussian noise with fixed variance $\sigma$. See Appendix~\ref{app:plasticity_experiment} for all details. As metrics, we use \emph{average per-task online accuracy}~\eqref{eq:per_task_avg_accuracy}, which is
\begin{equation}
    \textstyle
    \mathcal{A}_{t} = \frac{1}{N} \sum_{i=1}^{N} a_i^t,
    \label{eq_new:per_task_avg_accuracy}
\end{equation}
where $a_i^t$ are the online accuracies collected on the task $t$ via $N$ timesteps, corresponding to the duration of the task. In Figure~\ref{fig:impact_of_randomness}, we also use average accuracy over all $T$ tasks, i.e.
\begin{equation}
\textstyle
    \mathcal{A}_{T} = \frac{1}{T} \sum_{t=1}^{T} \mathcal{A}_{t}
    \label{eq_new:all_tasks_accuracy}
\end{equation}
The results are provided in Figure~\ref{fig:plasticity_benchmarks}. We observe that \emph{Soft Reset} is always better than \emph{Hard Reset} and most baselines despite the lack of knowledge of task boundaries. The gap is larger in the \emph{data efficient} regime. Moreover, we see that \emph{L2 Init} only performs well in the \emph{memorization} regime, and achieves comparable performance to \emph{Hard Reset} in the \emph{data efficient} one. The method \emph{L2 Init} could be viewed as an instantiation of our \emph{Soft Reset Proximal} method optimizing~\eqref{eq:proximal_objective} with $\gamma_t=0$ at every step, which is sub-optimal when there is similarity in the data. \emph{Bayesian Soft Reset} demonstrates significantly better performance overall, see also discussion below.

In Figure~\ref{fig:plasticity_benchmarks_soft_reset_variants}, we compare different variants of \emph{Soft Reset}. We observe that adding more compute for estimating $\gamma_{t}$ (thus, estimating non-stationarity, $K_\gamma=10$) as well as doing more updates on NN parameters (thus, more accurately adapting to non-staionarity, $K_\theta=10$) leads to better performance. All variants of \emph{Soft Reset} $\gamma_t$ parameters are shared for each NN layer, except for the Bayesian method. This variant is able to take advantage of a more complex \emph{per-parameter} drift model, while other variants performed considerably worse, see Appendix~\ref{app:bayesian_method_is_better_than_non_bayesian}. We hypothesize this is due to the NN parameters uncertainty estimates $\sigma_t$ which Bayesian method provide, while others do not, which leads to a more accurate drift model estimation, since uncertainty is used in this update~\eqref{eq:drift_via_predictive_ll}. But, this approach comes at a higher computational cost, see Appendix~\ref{app:computational_complexity}. In Appendix~\ref{app:bayesian_is_better}, we provide ablations of the structure of the drift model, as well as of the impact of learning the drift parameter.

\begin{figure}[t]
    \centering
    \includegraphics[scale=0.25]{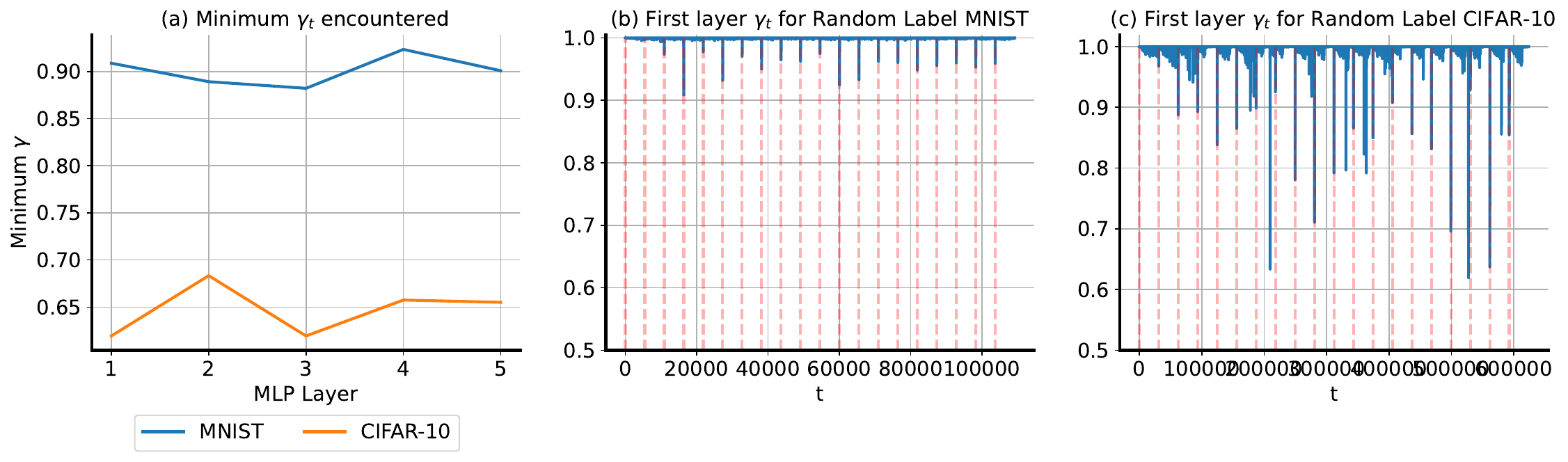}
    \vspace{-5pt}
    \caption{\textbf{Left:} the minimum encountered $\gamma_t$ for each layer on random-label MNIST and CIFAR-10. \textbf{Center:} the dynamics of $\gamma_t$ on the first 20 tasks on MNIST. \textbf{Right:} the same on CIFAR-10.}
    \label{fig:qualitative_behavior_main}
    \vspace{-10pt}

\end{figure}

\textbf{Qualitative behavior of \emph{Soft Resets}.}
For \emph{Soft Reset}, we track the values of $\gamma_t$ for the first MLP layer when trained on random-label tasks studied above (only $20$ tasks), as well as the minimum encountered value of $\gamma_t$ for each layer, which highlights the maximum amount of resets. Figure~\ref{fig:qualitative_behavior_main}b,c shows $\gamma_t$ as a function of $t$, and suggests that $\gamma_t$ aggressively decreases at task boundaries (red dashed lines). The range of values of $\gamma_t$ depends on the task and on the layer, see Figure~\ref{fig:qualitative_behavior_main}a. Overall, $\gamma_t$ changes more aggressively for long duration (memorization) random-label CIFAR-10 and less for shorter (data-efficient) random-label MNIST. See Appendix~\ref{app:qualitative_behavior_of_soft_resets_on_random_label_tasks} for more detailed results.

To study the behavior of \emph{Soft Reset} under input distribution non-stationarity, we consider a variant of Permuted MNIST where each image is partitioned into patches of a given size. The non-stationarity is controlled by permuting the patches (not pixels). Figure~\ref{fig:impact_of_randomness}a shows the minimum encountered $\gamma_t$ for each layer for different patch sizes. As the patch size increases and the problem becomes more stationary, the range of values for $\gamma_t$ is less aggressive. See Appendix~\ref{app:qualitative_behavior_of_soft_resets_on_permutted_patches_mnist} for more detailed results.

\textbf{Impact of non-stationarity.} We consider a variant of random-label MNIST where for each task, an image has either a random or a true label. The label assignment is kept fixed throughout the task and is changed at task boundaries. We consider cases of $20\%$, $40\%$ and $60\%$ of random labels and we control the duration of each task (number of epochs). In total, the stream contains $200$ tasks. In Figure~\ref{fig:impact_of_randomness}b, we show performance of \emph{Online SGD}, \emph{Hard Reset} and in Figure~\ref{fig:impact_of_randomness}c, the one of \emph{Soft Reset} and of \emph{Bayesian Soft Reset}. See Appendix~\ref{app:impact_of_non_stationarity} for more details. The results suggest that for the shortest duration of the tasks, the performance of all the methods is similar. As we increase the duration of each of the task (moving along the x-axis), we see that both \emph{Soft Resets} variants perform better than SGD and the gap widens as the duration increases. This implies that \emph{Soft Resets} is more effective with infrequent data distribution changes. We also observe that Bayesian method performs better in all the cases, highlighting the importance of estimating uncertainty for NN parameters.

\subsection{Reinforcement learning}

\textbf{Reinforcement learning experiments.} We conduct Reinforcement Learning (RL) experiments in the highly off-policy regime, similarly to \citep{nikishin2022primacy}, since in this setting  \emph{loss of plasticity} was observed. We ran \emph{SAC}~\citep{haarnoja2018soft} agent with default parameters from Brax~\citep{brax2021github} on the \emph{Hopper}-v5 and \emph{Humanoid}-v4 GYM~\citep{open_ai_gym} environments (from Brax~\citep{brax2021github}). To reproduce the setting from ~\citep{nikishin2022primacy}, we control the off-policyness of the agent by setting the \emph{off-policy ratio} $M$ such that for every $128$ environment steps, we do $128 M$ gradient steps with batch size of $256$ on the replay buffer. As baselines we consider ordinary \emph{SAC}, hard-coded \emph{Hard Reset} where we reset all the parameters $K=5$ times throughout training (every $200000$ steps), while keeping the replay buffer fixed (similarly to~\citep{nikishin2022primacy}).
We employ our \emph{Soft Reset} method as follows. After we have collected fresh data from the environment, we do one gradient update on $\gamma_t$ (shared for all the parameters within each layer) with batch size of $128$ on this new chunk of data and the previously collected one, i.e., two chunks of data in total. Then we initialize $\tilde{\theta}_t(\gamma_t)$ and we employ the update rule~\eqref{eq_app:many_grad_updates} where the regularization $\tilde{\theta}_t(\gamma_t)$ is kept constant for all the off-policy gradient updates on the replay buffer. See Appendix~\ref{app:rl_experiment} for more details.

\begin{figure}[tb]
    \centering
    \includegraphics[scale=0.3]{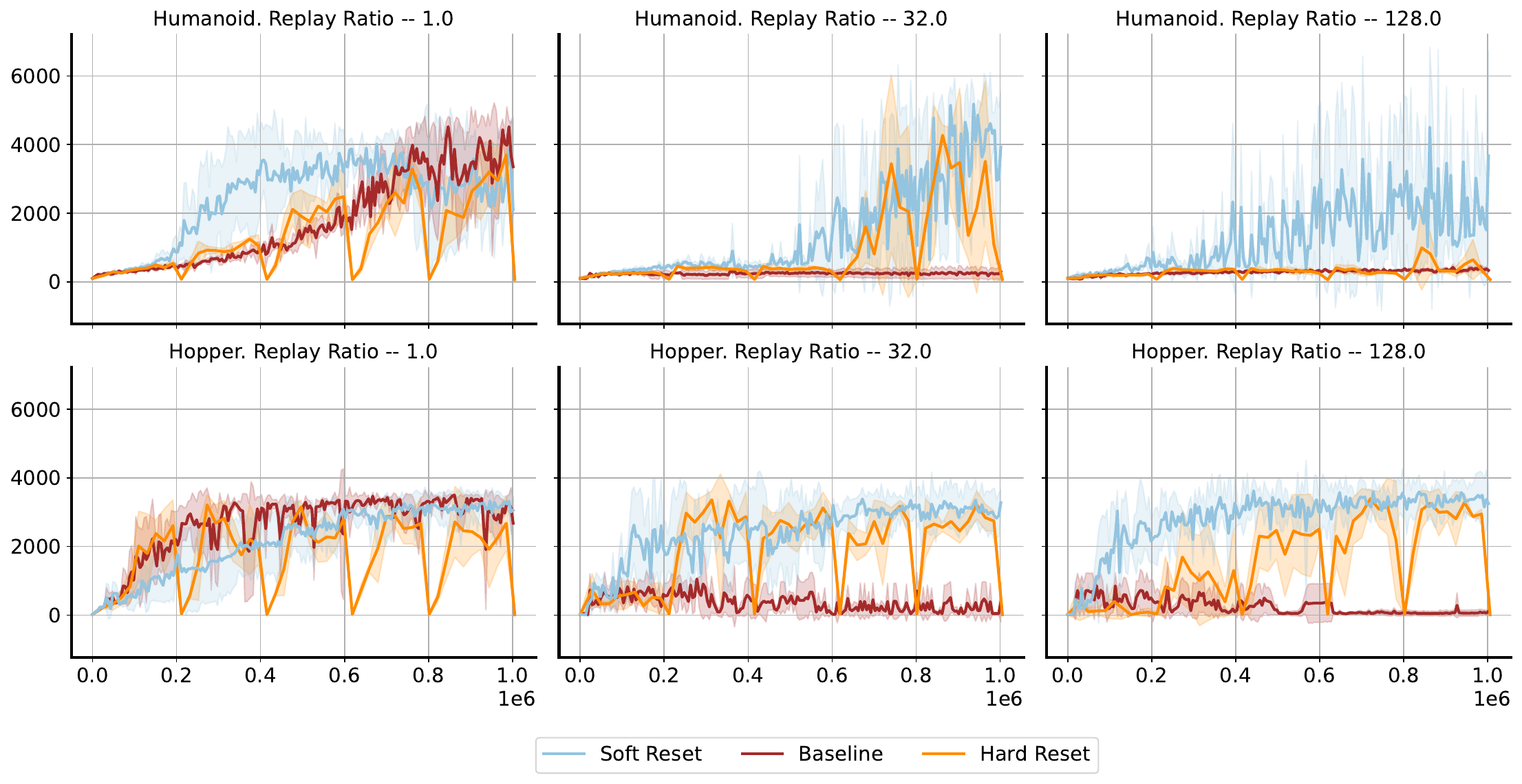}
\vspace{-0.2cm}
    \caption{RL results. First row is humanoid, second is hopper. Each column corresponds to different replay ratio. The x-axis is the number of total timesteps,  the y-axis the average reward. The shaded area denotes the standard deviation across $3$ random seeds and the solid line indicates the mean.}
    \label{fig:rl_results}
\end{figure}

The results are given in Figure~\ref{fig:rl_results}. As the off-policy ratio increases, \emph{Soft Reset} becomes more efficient than the baselines. This is consistent with our finding in Figure~\ref{fig:impact_of_randomness}b,c, where we showed that the performance of \emph{Soft Reset} is better when the data distribution is not changing fast. Figure~\ref{fig:rl_qualitative} in Appendix~\ref{app:rl_experiment} shows the value of learned $\gamma_t$. It shows $\gamma_t$ mostly change for the value function and not for the policy indicating that the main source of non-stationarity comes from the value function.

\vspace{-2pt}
\section{Conclusion}
\vspace{-2pt}

Learning efficiently on non-stationary distributions is critical to a number of applications of deep neural networks, most prominently in reinforcement learning. In this paper, we have proposed a new method, \emph{Soft Resets}, which improves the robustness of stochastic gradient descent to nonstationarities in the data-generating distribution by modeling the drift in Neural Network (NN) parameters. The proposed drift model implements \emph{soft reset} mechanism where the amount of reset is controlled by the drift parameter $\gamma_t$. We showed that we could learn this drift parameter from the data and therefore we could learn \emph{when} and \emph{how far} to reset each Neural Network parameter. We incorporate the drift model in the learning algorithm which improves learning in scenarios with plasticity loss. The variant of our method which models uncertainty in the parameters achieves the best performance on plasticity benchmarks so far, highlighting the promise of the Bayesian approach. Furthermore, we found that our approach is particularly effective either on data distributions with a lot of similarity or on slowly changing distributions. Our findings open the door to a variety of exciting directions for future work, such as investigating the connection to continual learning and deepening our theoretical analysis of the proposed approach.

\bibliographystyle{plain}
\bibliography{main}

\begin{thebibliography}{10}

\bibitem{abbas2023loss}
Zaheer Abbas, Rosie Zhao, Joseph Modayil, Adam White, and Marlos~C. Machado.
\newblock Loss of plasticity in continual deep reinforcement learning, 2023.

\bibitem{ash2020warmstarting}
Jordan~T. Ash and Ryan~P. Adams.
\newblock On warm-starting neural network training, 2020.

\bibitem{bencomo2023implicit}
Gianluca~M. Bencomo, Jake~C. Snell, and Thomas~L. Griffiths.
\newblock Implicit maximum a posteriori filtering via adaptive optimization,
  2023.

\bibitem{berariu2021study}
Tudor Berariu, Wojciech Czarnecki, Soham De, Jorg Bornschein, Samuel Smith,
  Razvan Pascanu, and Claudia Clopath.
\newblock A study on the plasticity of neural networks.
\newblock {\em arXiv preprint arXiv:2106.00042}, 2021.

\bibitem{Blundell15}
Charles Blundell, Julien Cornebise, Koray Kavukcuoglu, and Daan Wierstra.
\newblock Weight uncertainty in neural network.
\newblock In Francis Bach and David Blei, editors, {\em Proceedings of the 32nd
  International Conference on Machine Learning}, volume~37 of {\em Proceedings
  of Machine Learning Research}, pages 1613--1622, Lille, France, 07--09 Jul
  2015. PMLR.

\bibitem{open_ai_gym}
Greg Brockman, Vicki Cheung, Ludwig Pettersson, Jonas Schneider, John Schulman,
  Jie Tang, and Wojciech Zaremba.
\newblock Openai gym, 2016.

\bibitem{Broderick2013}
Tamara Broderick, Nicholas Boyd, Andre Wibisono, Ashia~C Wilson, and Michael~I
  Jordan.
\newblock Streaming variational bayes.
\newblock In C.J. Burges, L.~Bottou, M.~Welling, Z.~Ghahramani, and K.Q.
  Weinberger, editors, {\em Advances in Neural Information Processing Systems},
  volume~26. Curran Associates, Inc., 2013.

\bibitem{CeLu06}
N.~Cesa-Bianchi and G.~Lugosi.
\newblock {\em Prediction, Learning, and Games}.
\newblock Cambridge University Press, Cambridge, 2006.

\bibitem{chang2023lowrank}
Peter~G. Chang, Gerardo Durán-Martín, Alexander~Y Shestopaloff, Matt Jones,
  and Kevin Murphy.
\newblock Low-rank extended kalman filtering for online learning of neural
  networks from streaming data, 2023.

\bibitem{Chen2018Lifelong}
Zhiyuan Chen and Bing Liu.
\newblock Lifelong machine learning.
\newblock {\em Synthesis Lectures on Artificial Intelligence and Machine
  Learning}, 12(3):1--207, 2018.

\bibitem{dance2022fastscalablespikeslab}
Hugh Dance and Brooks Paige.
\newblock Fast and scalable spike and slab variable selection in
  high-dimensional gaussian processes, 2022.

\bibitem{Dohare2024LossOP}
Shibhansh Dohare, J.~Fernando Hernandez-Garcia, Qingfeng Lan, Parash Rahman,
  Ashique~Rupam Mahmood, and Richard~S. Sutton.
\newblock Loss of plasticity in deep continual learning.
\newblock {\em Nature}, 632:768 -- 774, 2024.

\bibitem{dohare2022continual}
Shibhansh Dohare, Richard~S. Sutton, and A.~Rupam Mahmood.
\newblock Continual backprop: Stochastic gradient descent with persistent
  randomness, 2022.

\bibitem{elsayed2024addressing}
Mohamed Elsayed and A.~Rupam Mahmood.
\newblock Addressing loss of plasticity and catastrophic forgetting in
  continual learning, 2024.

\bibitem{brax2021github}
C.~Daniel Freeman, Erik Frey, Anton Raichuk, Sertan Girgin, Igor Mordatch, and
  Olivier Bachem.
\newblock Brax - a differentiable physics engine for large scale rigid body
  simulation, 2021.
\newblock \url{http://github.com/google/brax}.

\bibitem{pmlr-v9-glorot10a}
Xavier Glorot and Yoshua Bengio.
\newblock Understanding the difficulty of training deep feedforward neural
  networks.
\newblock In Yee~Whye Teh and Mike Titterington, editors, {\em Proceedings of
  the Thirteenth International Conference on Artificial Intelligence and
  Statistics}, volume~9 of {\em Proceedings of Machine Learning Research},
  pages 249--256, Chia Laguna Resort, Sardinia, Italy, 13--15 May 2010. PMLR.

\bibitem{GyLiLu12}
A.~Gy{\"o}rgy, T.~Linder, and G.~Lugosi.
\newblock Efficient tracking of large classes of experts.
\newblock {\em IEEE Transactions on Information Theory}, IT-58(11):6709--6725,
  Nov. 2012.

\bibitem{GySz16}
Andr\'as Gy\"orgy and Csaba Szepesv\'ari.
\newblock Shifting regret, mirror descent, and matrices.
\newblock In {\em Proceedings of The 33rd International Conference on Machine
  Learning}, pages 2943--2951, 2016.

\bibitem{haarnoja2018soft}
Tuomas Haarnoja, Aurick Zhou, Pieter Abbeel, and Sergey Levine.
\newblock Soft actor-critic: Off-policy maximum entropy deep reinforcement
  learning with a stochastic actor, 2018.

\bibitem{HADSELL20201028}
Raia Hadsell, Dushyant Rao, Andrei~A. Rusu, and Razvan Pascanu.
\newblock Embracing change: Continual learning in deep neural networks.
\newblock {\em Trends in Cognitive Sciences}, 24(12):1028--1040, 2020.

\bibitem{hall2013dynamical}
Eric Hall and Rebecca Willett.
\newblock Dynamical models and tracking regret in online convex programming.
\newblock In {\em International Conference on Machine Learning}, pages
  579--587. PMLR, 2013.

\bibitem{HaSe09}
E.~Hazan and C.~Seshadhri.
\newblock Efficient learning algorithms for changing environments.
\newblock In {\em Proc. 26th Annual International Conference on Machine
  Learning}, pages 393--400. ACM, 2009.

\bibitem{hazan2023introduction}
Elad Hazan.
\newblock Introduction to online convex optimization, 2023.

\bibitem{he2015delving}
Kaiming He, Xiangyu Zhang, Shaoqing Ren, and Jian Sun.
\newblock Delving deep into rectifiers: Surpassing human-level performance on
  imagenet classification, 2015.

\bibitem{HeWa98}
M.~Herbster and M.~K. Warmuth.
\newblock Tracking the best expert.
\newblock {\em Machine Learning}, 32(2):151--178, 1998.

\bibitem{Ishwaran_2005}
Hemant Ishwaran and J.~Sunil Rao.
\newblock Spike and slab variable selection: Frequentist and bayesian
  strategies.
\newblock {\em The Annals of Statistics}, 33(2), April 2005.

\bibitem{jones2024bayesianonlinenaturalgradient}
Matt Jones, Peter Chang, and Kevin Murphy.
\newblock Bayesian online natural gradient (bong), 2024.

\bibitem{Jones2024HumanlikeLI}
Matt Jones, Tyler~R. Scott, and Michael~Curtis Mozer.
\newblock Human-like learning in temporally structured environments.
\newblock In {\em AAAI Spring Symposia}, 2024.

\bibitem{khodak2019adaptive}
Mikhail Khodak, Maria-Florina Balcan, and Ameet Talwalkar.
\newblock Adaptive gradient-based meta-learning methods, 2019.

\bibitem{kingma2017adam}
Diederik~P. Kingma and Jimmy Ba.
\newblock Adam: A method for stochastic optimization, 2017.

\bibitem{Kirkpatrick_2017}
James Kirkpatrick, Razvan Pascanu, Neil Rabinowitz, Joel Veness, Guillaume
  Desjardins, Andrei~A. Rusu, Kieran Milan, John Quan, Tiago Ramalho, Agnieszka
  Grabska-Barwinska, Demis Hassabis, Claudia Clopath, Dharshan Kumaran, and
  Raia Hadsell.
\newblock Overcoming catastrophic forgetting in neural networks.
\newblock {\em Proceedings of the National Academy of Sciences},
  114(13):3521–3526, March 2017.

\bibitem{Krizhevsky2009LearningML}
Alex Krizhevsky.
\newblock Learning multiple layers of features from tiny images.
\newblock Technical report, University of Toronto, 2009.
\newblock \url{https://www.cs.toronto.edu/~kriz/learning-features-2009-TR.pdf}.

\bibitem{kumar2021implicit}
Aviral Kumar, Rishabh Agarwal, Dibya Ghosh, and Sergey Levine.
\newblock Implicit under-parameterization inhibits data-efficient deep
  reinforcement learning, 2021.

\bibitem{kumar2023maintaining}
Saurabh Kumar, Henrik Marklund, and Benjamin~Van Roy.
\newblock Maintaining plasticity in continual learning via regenerative
  regularization, 2023.

\bibitem{Kurle2020Continual}
Richard Kurle, Botond Cseke, Alexej Klushyn, Patrick van~der Smagt, and Stephan
  Günnemann.
\newblock Continual learning with bayesian neural networks for non-stationary
  data.
\newblock In {\em International Conference on Learning Representations}, 2020.

\bibitem{lin2021clear}
Zhiqiu Lin, Jia Shi, Deepak Pathak, and Deva Ramanan.
\newblock The clear benchmark: Continual learning on real-world imagery.
\newblock In {\em Thirty-fifth Conference on Neural Information Processing
  Systems Datasets and Benchmarks Track}, 2021.

\bibitem{lyle2022understanding}
Clare Lyle, Mark Rowland, and Will Dabney.
\newblock Understanding and preventing capacity loss in reinforcement learning,
  2022.

\bibitem{lyle2024disentanglingcausesplasticityloss}
Clare Lyle, Zeyu Zheng, Khimya Khetarpal, Hado van Hasselt, Razvan Pascanu,
  James Martens, and Will Dabney.
\newblock Disentangling the causes of plasticity loss in neural networks, 2024.

\bibitem{lyle2023understanding}
Clare Lyle, Zeyu Zheng, Evgenii Nikishin, Bernardo~Avila Pires, Razvan Pascanu,
  and Will Dabney.
\newblock Understanding plasticity in neural networks, 2023.

\bibitem{mnih2013playing}
Volodymyr Mnih, Koray Kavukcuoglu, David Silver, Alex Graves, Ioannis
  Antonoglou, Daan Wierstra, and Martin Riedmiller.
\newblock Playing atari with deep reinforcement learning, 2013.

\bibitem{variationalCL2018}
Cuong~V. Nguyen, Yingzhen Li, Thang~D. Bui, and Richard~E. Turner.
\newblock Variational continual learning.
\newblock In {\em International Conference on Learning Representations}, 2018.

\bibitem{nikishin2023deep}
Evgenii Nikishin, Junhyuk Oh, Georg Ostrovski, Clare Lyle, Razvan Pascanu, Will
  Dabney, and André Barreto.
\newblock Deep reinforcement learning with plasticity injection, 2023.

\bibitem{nikishin2022primacy}
Evgenii Nikishin, Max Schwarzer, Pierluca D'Oro, Pierre-Luc Bacon, and Aaron
  Courville.
\newblock The primacy bias in deep reinforcement learning, 2022.

\bibitem{proximal_boyd}
Neal Parikh and Stephen Boyd.
\newblock Proximal algorithms.
\newblock {\em Found. Trends Optim.}, 1(3):127–239, jan 2014.

\bibitem{PARISI201954}
German~I. Parisi, Ronald Kemker, Jose~L. Part, Christopher Kanan, and Stefan
  Wermter.
\newblock Continual lifelong learning with neural networks: A review.
\newblock {\em Neural Networks}, 113:54--71, 2019.

\bibitem{schulman2017proximal}
John Schulman, Filip Wolski, Prafulla Dhariwal, Alec Radford, and Oleg Klimov.
\newblock Proximal policy optimization algorithms, 2017.

\bibitem{sokar23a}
Ghada Sokar, Rishabh Agarwal, Pablo~Samuel Castro, and Utku Evci.
\newblock The dormant neuron phenomenon in deep reinforcement learning.
\newblock In Andreas Krause, Emma Brunskill, Kyunghyun Cho, Barbara Engelhardt,
  Sivan Sabato, and Jonathan Scarlett, editors, {\em Proceedings of the 40th
  International Conference on Machine Learning}, volume 202 of {\em Proceedings
  of Machine Learning Research}, pages 32145--32168. PMLR, 23--29 Jul 2023.

\bibitem{sokar2023dormantneuronphenomenondeep}
Ghada Sokar, Rishabh Agarwal, Pablo~Samuel Castro, and Utku Evci.
\newblock The dormant neuron phenomenon in deep reinforcement learning, 2023.

\bibitem{titsias2023kalman}
Michalis~K. Titsias, Alexandre Galashov, Amal Rannen-Triki, Razvan Pascanu,
  Yee~Whye Teh, and Jorg Bornschein.
\newblock Kalman filter for online classification of non-stationary data, 2023.

\bibitem{ou_process}
G.~E. Uhlenbeck and L.~S. Ornstein.
\newblock On the theory of the brownian motion.
\newblock {\em Phys. Rev.}, 36:823--841, Sep 1930.

\bibitem{van2019three}
Gido~M van~de Ven and Andreas~S Tolias.
\newblock Three scenarios for continual learning.
\newblock {\em arXiv preprint arXiv:1904.07734}, 2019.

\bibitem{metagrad}
Tim van Erven, Wouter~M. Koolen, and Dirk van~der Hoeven.
\newblock Metagrad: Adaptation using multiple learning rates in online
  learning.
\newblock {\em Journal of Machine Learning Research}, 22(161):1--61, 2021.

\bibitem{verwimp2024continual}
Eli Verwimp, Rahaf Aljundi, Shai Ben-David, Matthias Bethge, Andrea Cossu,
  Alexander Gepperth, Tyler~L. Hayes, Eyke Hüllermeier, Christopher Kanan,
  Dhireesha Kudithipudi, Christoph~H. Lampert, Martin Mundt, Razvan Pascanu,
  Adrian Popescu, Andreas~S. Tolias, Joost van~de Weijer, Bing Liu, Vincenzo
  Lomonaco, Tinne Tuytelaars, and Gido~M. van~de Ven.
\newblock Continual learning: Applications and the road forward, 2024.

\bibitem{wenzel2020good}
Florian Wenzel, Kevin Roth, Bastiaan~S. Veeling, Jakub Świątkowski, Linh
  Tran, Stephan Mandt, Jasper Snoek, Tim Salimans, Rodolphe Jenatton, and
  Sebastian Nowozin.
\newblock How good is the bayes posterior in deep neural networks really?,
  2020.

\bibitem{zhai2023online}
Runtian Zhai, Stefan Schroedl, Aram Galstyan, Anoop Kumar, Greg~Ver Steeg, and
  Pradeep Natarajan.
\newblock Online continual learning for progressive distribution shift
  ({OCL}-{PDS}): A practitioner's perspective, 2023.

\bibitem{zinkevich2003}
Martin Zinkevich.
\newblock Online convex programming and generalized infinitesimal gradient
  ascent.
\newblock In {\em Proceedings of the Twentieth International Conference on
  International Conference on Machine Learning}, ICML'03, page 928–935. AAAI
  Press, 2003.

\end{thebibliography}


\newpage
\section*{NeurIPS Paper Checklist}

\begin{enumerate}

\item {\bf Claims}
    \item[] Question: Do the main claims made in the abstract and introduction accurately reflect the paper's contributions and scope?
    \item[] Answer: \answerYes{} 
    \item[] Justification: We outline main contributions of the paper in the introduction and abstract.

\item {\bf Limitations}
    \item[] Question: Does the paper discuss the limitations of the work performed by the authors?
    \item[] Answer: \answerYes{} 
    \item[] Justification: We discuss limitations in the experimental section

\item {\bf Theory Assumptions and Proofs}
    \item[] Question: For each theoretical result, does the paper provide the full set of assumptions and a complete (and correct) proof?
    \item[] Answer: \answerNA{} 
    \item[] Justification:

    \item {\bf Experimental Result Reproducibility}
    \item[] Question: Does the paper fully disclose all the information needed to reproduce the main experimental results of the paper to the extent that it affects the main claims and/or conclusions of the paper (regardless of whether the code and data are provided or not)?
    \item[] Answer: \answerYes{} 
    \item[] Justification: We disclose the experimental information in Experimental and Appendix sections.

\item {\bf Open access to data and code}
    \item[] Question: Does the paper provide open access to the data and code, with sufficient instructions to faithfully reproduce the main experimental results, as described in supplemental material?
    \item[] Answer: \answerNo{} 
    \item[] Justification: Unfortunately, due to IP constrains, we cannot release the code for the paper.

\item {\bf Experimental Setting/Details}
    \item[] Question: Does the paper specify all the training and test details (e.g., data splits, hyperparameters, how they were chosen, type of optimizer, etc.) necessary to understand the results?
    \item[] Answer: \answerYes{} 
    \item[] Justification: We provide experimental details in the appendix.

\item {\bf Experiment Statistical Significance}
    \item[] Question: Does the paper report error bars suitably and correctly defined or other appropriate information about the statistical significance of the experiments?
    \item[] Answer: \answerYes{} 
    \item[] Justification: We specify that we report results with $3$ random seeds with mean and standard deviation.

\item {\bf Experiments Compute Resources}
    \item[] Question: For each experiment, does the paper provide sufficient information on the computer resources (type of compute workers, memory, time of execution) needed to reproduce the experiments?
    \item[] Answer: \answerYes{} 
    \item[] Justification: We provide information about compute resources required in the appendix.

\item {\bf Code Of Ethics}
    \item[] Question: Does the research conducted in the paper conform, in every respect, with the NeurIPS Code of Ethics \url{https://neurips.cc/public/EthicsGuidelines}?
    \item[] Answer: \answerYes{} 
    \item[] Justification: Based on our understanding, our work conforms to the every aspect of NeurIPS Code of Ethics.

\item {\bf Broader Impacts}
    \item[] Question: Does the paper discuss both potential positive societal impacts and negative societal impacts of the work performed?
    \item[] Answer: \answerNA{} 
    \item[] Justification:

\item {\bf Safeguards}
    \item[] Question: Does the paper describe safeguards that have been put in place for responsible release of data or models that have a high risk for misuse (e.g., pretrained language models, image generators, or scraped datasets)?
    \item[] Answer: \answerNA{} 
    \item[] Justification:

\item {\bf Licenses for existing assets}
    \item[] Question: Are the creators or original owners of assets (e.g., code, data, models), used in the paper, properly credited and are the license and terms of use explicitly mentioned and properly respected?
    \item[] Answer: \answerYes{} 
    \item[] Justification: We cite the works which introduced the publicly available datasets

\item {\bf New Assets}
    \item[] Question: Are new assets introduced in the paper well documented and is the documentation provided alongside the assets?
    \item[] Answer: \answerNA{} 
    \item[] Justification:

\item {\bf Crowdsourcing and Research with Human Subjects}
    \item[] Question: For crowdsourcing experiments and research with human subjects, does the paper include the full text of instructions given to participants and screenshots, if applicable, as well as details about compensation (if any)?
    \item[] Answer: \answerNA{} 
    \item[] Justification:

\item {\bf Institutional Review Board (IRB) Approvals or Equivalent for Research with Human Subjects}
    \item[] Question: Does the paper describe potential risks incurred by study participants, whether such risks were disclosed to the subjects, and whether Institutional Review Board (IRB) approvals (or an equivalent approval/review based on the requirements of your country or institution) were obtained?
    \item[] Answer: \answerNA{} 
    \item[] Justification:

\end{enumerate}

\newpage
\appendix

\section{Ornstein-Uhlenbeck process}
\label{app:ou_process}

We make use that the Ornstein-Uhlenbeck process~\citep{ou_process} defines a SDE that can be solved explicitly and written as a time-continuous Gaussian Markov process with transition density
\begin{equation}
    p(x_t | x_s) = \mathcal{N}(x_s e^{-(t-s)}, (1 - e^{- 2(t-s) }) \sigma_0^2 I ),
\end{equation}
for any pair of times  $t > s$. Based on this  as a drift model for the parameters $\theta_t$ (so $\theta_t$ is the state $x_t$) we use the  conditional density
\begin{equation}
    p(\theta_{t+1} | \theta_t) = \mathcal{N}(\theta_t \gamma_t, (1 - \gamma_t^2) \sigma_0^2 I ),
\end{equation}
where $\gamma_t = e^{- \delta_t}$ and $\delta_t \geq 0$ corresponds to the learnable discretization time step. In other words, by learning $\gamma_t$ online we equivalently learn the amount of a continuous “time shift” $\delta_t$ between two consecutive states in the OU process. This essentially models parameter drift since e.g. if $\gamma_t=1$, then $\delta_t=0$ and there is no “time shift” which means that the next state/parameter remains the same as the previous one, i.e. $\theta_{t+1}=\theta_t$.

\section{Other choices of drift model}
\label{sec:other_models}

In this section, we discuss alternative choices of a drift model instead of~\eqref{eq:ou_model}.

\paragraph{Independent mean and variance of the drift.} We consider the drift model where the mean and the variance are not connected, i.e.,
\begin{equation}
    p(\theta_{t+1} | \theta_t, \gamma_t, \beta_t) = \mathcal{N}(\theta_{t+1}; \gamma_t \theta_t + (1-\gamma_t) \mu_0; \beta_t^2),
    \label{eq_app:independent_variance_drift_model}
\end{equation}
where $\gamma_t \in [0,1]$ is the parameters controlling the mean of the distribution and $\beta_t$ is the learned variance. When $\beta$ is fixed, this would be similar to our experiment in Figure~\ref{fig:perfect_resets_ablation} where we assumed known task boundaries and we do not estimate the drift parameters but assume it as a hyperparameter. Figure~\ref{fig:perfect_resets_ablation}, left corresponds to the case when $\beta_t$ is a fixed parameter independent from $\gamma_t$ whereas Figure~\ref{fig:perfect_resets_ablation}, right corresponds to the case when $\beta_t = \sqrt{1-\gamma_t^2}\sigma_0$, i.e., when we use the drift model~\eqref{eq:ou_model}. We see from the results, using drift model~\eqref{eq:ou_model} leads to a better performance. In case when $\beta_t$ are learned, estimating the parameters of this model will likely overfit to the noise since there is a lot of degrees of freedom.

\paragraph{Shrink \& Perturb~\citep{ash2020warmstarting}.} When we do not use the mean of the initialization, we can use the following drift model
\begin{equation}
    p(\theta_{t+1} | \theta_t, \lambda_t, \beta_t) = \mathcal{N}(\theta_{t+1}; \lambda_t \theta_t; \beta_t^2)
    \label{eq_app:shrunk_and_perturb}
\end{equation}
Similarly to the case of~\eqref{eq_app:independent_variance_drift_model}, estimating both parameters $\lambda_t$ and $\beta_t$ from the data will likely overfit to the noise.

\paragraph{Arbitrary linear model.} We can use the arbitrary linear model of the form
\begin{equation}
    p(\theta_{t+1} | \theta_t, A_t, B_t) = \mathcal{N}(\theta_{t+1}; A_t \theta_t; B_t),
\end{equation}
but estimating the parameters $A_t$ and $B_t$ has too many degrees of freedom and will certainly overfit.

\paragraph{Gaussian Spike \& Slab} We consider a Gaussian~\citep{dance2022fastscalablespikeslab} approximation to Spike \& Slab~~\citep{Ishwaran_2005} prior
\begin{equation}
    p(\theta_{t+1} | \theta_{t}, \gamma_t) = \gamma_t p(\theta_{t+1} | \theta_t) + (1-\gamma_t) p_0(\theta_{t+1}),
\end{equation}
which is a mixture of two distributions - a Gaussian $p(\theta_{t+1} | \theta_t) = \mathcal{N}(\theta_{t+1}; \theta_t, \sigma^2)$ centered around the previous parameter $\theta_t$ and an initializing distribution $p_0(\theta_{t+1}) = \mathcal{N}(\theta_{t+1}; \mu_0, \sigma_0^2)$. This model, however, implements the mechanism of Hard reset as opposed to the soft ones. Moreover, estimating such a model and incorporating it into a learning update is more challenging since the mixture of Gaussian is not conjugate with respect to a Gaussian which will make the KL term~\eqref{eq_app:elbo} to be computed only approximately via Monte Carlo updates.

\section{Practical implementations of the drift model estimation}
\label{app:practical}

\paragraph{Stochastic approximation for drift parameters estimation} In practice, we use $M=1$, which leads to the stochastic approximation
\begin{equation}
    \textstyle
    \int p(y_{t+1} | x_{t+1}, \mu_t(\gamma^{k}_t) + \epsilon \sigma_{t}(\gamma^{k}_t)) \mathcal{N}(\epsilon; 0, I) d\epsilon \approx  p \left( y_{t+1} | x_{t+1}, \mu_t(\gamma^{k}_t) + \epsilon \sigma_{t}(\gamma^{k}_t) \right)
    \label{eq_app:stochastic_approx}
\end{equation}

\paragraph{Using NN initializing distribution.}
In the drift model~\eqref{eq:ou_model}, we assume that the initial distribution over parameters is given by $p_0(\theta) = \mathcal{N}(\theta; \mu_0; \sigma^2_0)$. In practice, we have access to the NN initializer $p_{init}(\theta) = \mathcal{N}(\theta; 0; \sigma^2_0)$ where $\mu_0 = 0$ (for most of the NNs). This means that we can replace $\epsilon$ from~\eqref{eq:drift_via_predictive_ll} by $\frac{1}{\sigma_0} \theta'_0$ where $\theta'_0 \sim p_{init}(\theta)$. This means that the term in~\eqref{eq_app:stochastic_approx} can be replaced by
\begin{equation}
    \textstyle
    p \left( y_{t+1} | x_{t+1}, \mu_t(\gamma^{k}_t) + \epsilon \sigma_{t}(\gamma^{k}_t) \right) = p \left( y_{t+1} | x_{t+1}, \mu_t(\gamma^{k}_t) + \theta_0' \sqrt{1-\gamma^2_t + \gamma^2_t \frac{\sigma^2_t}{\sigma^2_0}} \right),
    \label{eq_app:drift_from_init}
\end{equation}
where we used the fact that $\sigma^2_t(\gamma_t) = \gamma_t^2 \sigma^2_t + (1-\gamma_t^2) \sigma^2_0$. Note that in~\eqref{eq_app:drift_from_init}, we only need to know the ratio $\frac{\sigma^2}{\sigma^2_0}$ rather than both of these. We will see that in Section~\ref{sec:map_inference}, only this ratio is used for the underlying algorithm. Finally, in practice, we can tie $p_0(\theta)$ to the \emph{specific} initialization $\theta_0 \sim p_{init}(\theta)$. It was observed empricially~\citep{kumar2023maintaining} that using a specific initialization in gradient updates led to better performance than using samples from the initial distribution. This would imply that
\begin{equation}
    p_0(\theta) = \mathcal{N}(\theta; \theta_0, \tilde{\sigma}^2_{0}),
    \label{eq_app:modified_prior}
\end{equation}
with $\tilde{\sigma}^2_{0} = p^2 \sigma^2_0$. The parameter $p \leq 1$ accounts for the fact that the distribution $ p_0(\theta)$ should have lower than $p_{init}(\theta)$ variance since it uses the specific initializaiton from the prior. This modification would imply the following modification on the drift model term~\eqref{eq_app:drift_from_init}
\begin{equation}
    \textstyle
    p \left( y_{t+1} | x_{t+1}, \mu_t(\gamma^{k}_t) + \epsilon \sigma_{t}(\gamma^{k}_t) \right) = p \left( y_{t+1} | x_{t+1}, \mu_t(\gamma^{k}_t) + \theta'_0 p \sqrt{1-\gamma^2_t + \gamma^2_t \frac{\sigma^2_t}{\sigma^2_0}} \right)
    \label{eq_app:drift_from_specific_init}
\end{equation}

\section{Experimental details}
\label{app:experimental_details}

\subsection{Plasticity experiments}
\label{app:plasticity_experiment}

\paragraph{Tasks} In this section we provide experimental details. As plasticity tasks, we use a randomly selected subset of size $10000$ from CIFAR-10~\citep{Krizhevsky2009LearningML} and from MNIST. This subset is fixed for all the tasks. Within each task, we randomly permute labels for every image; we call such problems random-label classification problems. We study two regimes -- \emph{data efficient}, where we do $400$ epochs on a task with a batch size of $128$, and \emph{memorization}, a regime where we do only $70$ epochs with a batch size of $128$. As the main backbone architecture, we use MLP with $4$ hidden layers each having a hidden dimension of $256$ hidden units. We use ReLU activation function and do not use any batch or layer normalization. For the incoming data, we apply random crop, for MNIST to produce images of size $24 \times 24$ and for CIFAR-10 to produce images of size $28 \times 28$. We normalize images to be within $[0, 1]$ range by dividing by $255$. On top of that, we consider \emph{permuted MNIST} task with a similar training ragime as in~\citep{kumar2023maintaining} -- we consider a subset of $10000$ images, with batch size $16$ and each task is one epoch. As a backbone, we still use MLP with ReLU activation and $4$ hidden layers. Moreover, we considered \emph{permuted Patch MNIST}, where we permute patches, not individual pixels. In this case, we used a simple 4 layer convolutional neural network with 2 fully connected layers at the end.

\paragraph{Metrics} We use \emph{online accuracy} as first metric with results reported in Appendix~\ref{app:bayesian_is_better}. Moreover we use \emph{per-task Average Online Accuracy} which is
\begin{equation}
    \mathcal{A}_{t} = \frac{1}{N} \sum_{i=1}^{N} a_i^t,
    \label{eq:per_task_avg_accuracy}
\end{equation}
where $a_i^t$ are the online accuracies collected on the task $t$ via $N$ timesteps.

\paragraph{Baselines} First baseline is \emph{Online SGD} which sequentially learns over the sequence of task, with a fixed learning rate. \emph{Hard Reset} is the \emph{Online SGD} which resets all the parameters at task boundaries. \emph{L2 init}~\citep{kumar2023maintaining} adds a regularizer $\lambda ||\theta - \theta_0||^2$ term to each \emph{Online SGD} update where the regularization strength $\lambda$ is a hyperparameter. \emph{Shrink \& Perturb} applies the transformation $\lambda \theta_t + \sigma \epsilon, \epsilon \sim \mathcal{N}(\epsilon; 0, I)$ to each parameter before the gradient update. The hyperparameters are $\lambda$ and $\sigma$.

\paragraph{Soft Reset} corresponds to one update~\eqref{eq:drift_via_predictive_ll} starting from $1$ using $1$ Monte Carlo estimate. We always use $1$ Monte Carlo estimate for updating $\gamma_t$ as we found that it worked well in practice on these tasks. The hyperparameters of the method -- $\sigma^2_0$ initial variance of the prior, which we set to be equal to $p^2 \frac{1}{N}$ where $N$ is the width of the hidden layer and $p$ is a constant (hyperparameter).  It always equals to $p=0.1$. On top of that the second hyperparameter is $s$, such that $\sigma_t = s \sigma_0$, which controls the relative decrease of the constant posterior variance. This is the hyperparameter over which we sweep over. Another hyperparameter is the learning rate for learning $\gamma_t$. For \emph{Soft Reset Proximal}, we also have a proximal coefficient regularization constant $\lambda$. Besides that, we also sweep over the learning rate for the parameter. For the \emph{Bayesian Soft Reset}, we just add an additional learning rate for the variance $\alpha_{\sigma}$ and we do $1$ Monte Carlo sample for each ELBO update.

\paragraph{Hyper parameters selection and evaluation} For all the experiments, we run a sweep over the hyperparameters. We select the best hyperparameters based on the smallest cumulative error (sum of all $1 - a_i^t$ throughout the training). We then report the mean and the standard deviation across $3$ seeds in all the plots.

\paragraph{Hyperparameter ranges}. Learning rate $\alpha$ which is used to update parameters, for all the methods, is selected from $\{1e-4, 5e-4, 1e-3, 5e-3, 1e-2, 5e-2, 1e-1, 5e-1, 1.0\}$. The $\lambda_{init}$ parameter in \emph{L2 Init}, is selected from $\{10.0,1.0,0.0,1e-1,5e-1,1e-2,5e-2,1e-3,5e-3,1e-4,5e-4,1e-5,5e-5,1e-6,5e-6,1e-7,5e-7,1e-8,5e-8,1e-9,5e-9,1e-10,\}$. For S\&P, the shrink parameter $\lambda$ is selected from $\{1.0, 0.99999, 0.9999, 0.999, 0.99, 0.9, 0.8, 0.7, 0.5, 0.3, 0.2, 0.1\}$, and the perturbation parameter $\sigma$ is from $\{1e-1, 1e-2, 1e-3, 1e-4, 1e-5, 1e-6\}$. As noise distribution, we use the Neural Network initial distribution. For \emph{Soft Resets}, the learning rate for $\gamma_t$ is selected from $\{0.5, 0.1, 0.05, 0.01, 0.005, 0.001, 0.0005, 0.0001\}$, the constant $s$ is selected from $\{1.0, 0.95, 0.9, 0.8, 0.7, 0.6, 0.5, 0.3, 0.1\}$, the proximal cost $\tilde{\lambda}$ in ~\eqref{eq_app:proximal_objective} is selected from $\{1.0, 0.1, 0.01\}$, the same is true for the proximal cost in the Bayesian method~\eqref{eq_app:elbo_gaussian_expanded_efficient_temperature}. On top of that for the Bayesian method, we always use $p$ (see Algorithm~\ref{alg:alg_bayesian}) equal to $p=0.05$ and $s=0.9$, i.e. the posterior is always slightly smaller than the prior. Finally for the Bayesian method we had to learn the variance with learning rate from $\{0.01, 0.1, 1, 10 \}$ range.

In practice, we found that there is one learning rate of $0.1$, which was always the best in practice for most of the methods and only proximal \emph{Soft Resets} on \emph{memorization} CIFAR-10 required smaller learning rate $0.01$. This allowed us to significantly reduce the hyperparameter sweep.

\subsection{Impact of non-stationarity experiments}
\label{app:impact_of_non_stationarity}

\begin{figure}[tb]
    \centering
    \includegraphics[scale=0.2]{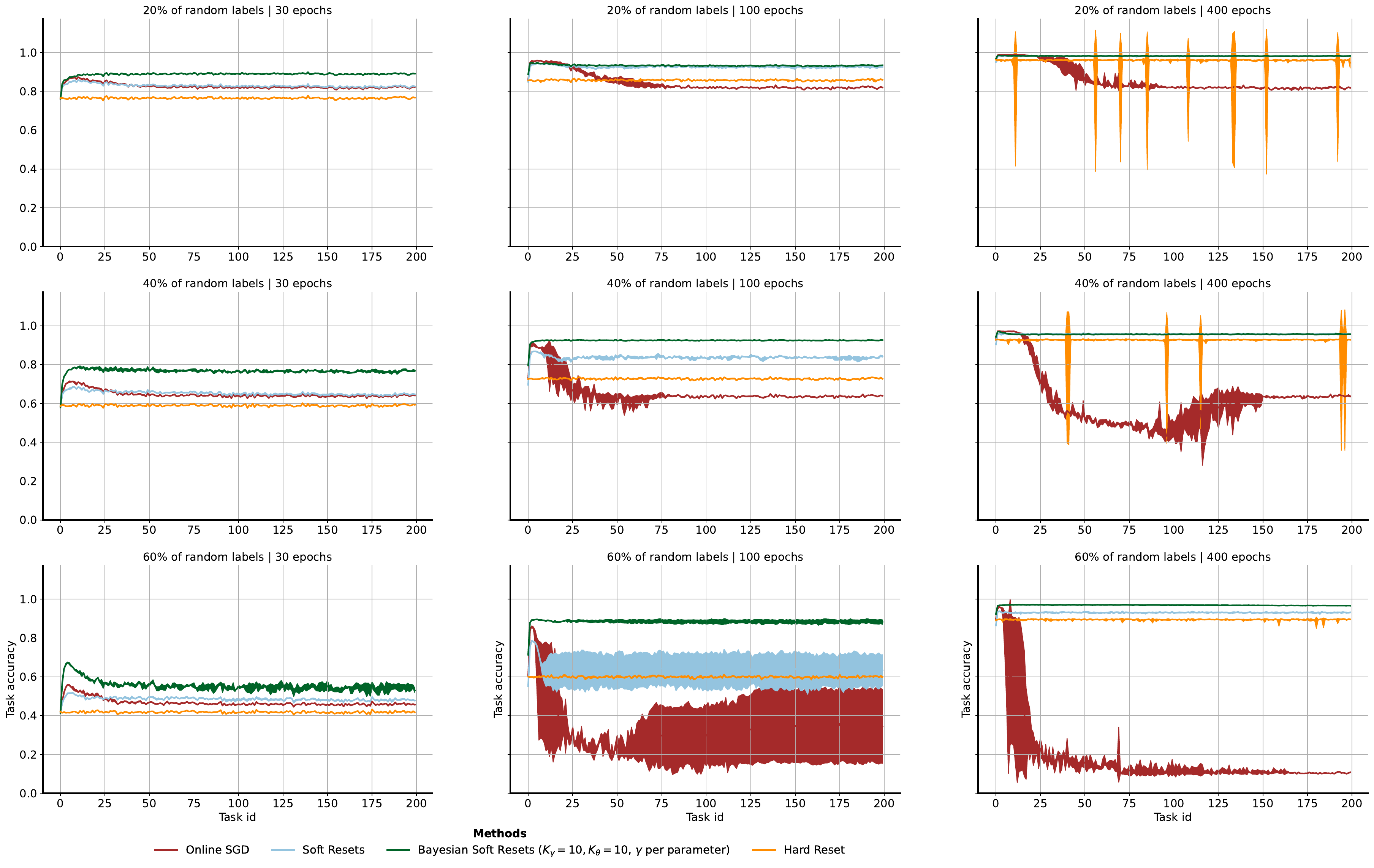}
    \caption{\textbf{Non-stationarity impact}. The x-axis denotes task id, each column denotes the duration, whereas a row denotes the amount of label noise. Each color denotes the method studied. The y-axis denotes average over 3 seeds online accuracy. }
    \label{fig:impact_of_randomness_detailed}
\end{figure}

In this experiment, we consider a subset of $10000$ images from MNIST (fixed throughtout all experiment) and a sequence of tasks. Each task is constructed by assigning either a true or a random label to each image from MNIST, where the probability of assignment is controlled by the experiment. The duration of each is controlled by the number of epochs with batch size of $128$. As backbone we use MLP with $4$ hidden layers and $256$ hidden units and ReLU activation. For all the methods, the learning rate is $0.1$. For \emph{Soft Resets}, we use $s = 0.9$ and $p=1$ and $\eta_{\gamma}=0.01$. Bayesian method uses proximal cost $\lambda=0.01$. Detailed results are given in Figure~\ref{fig:impact_of_randomness_detailed}.

\subsection{Reinforcement learning experiments}
\label{app:rl_experiment}

We conduct experiments in the RL environments. We take the canonical implementation of Soft-Actor Critic(SAC) from Brax~\citep{brax2021github} repo in github, which uses $2$ layer MLPs for both policy and Q-function. It employs ReLU activation functions for both. On top of that, it uses $2$ MLP networks to parameterize $Q$-function (see Brax~\citep{brax2021github}) for more details. To employ \emph{Soft Reset}, we do the following. After we have collected a chunk of data ($128$) time-steps, we do one update~\eqref{eq:drift_via_predictive_ll} on $\gamma_t$ starting from $1$ at every update of $\gamma_t$, where $\gamma_t$ is shared for all the parameters within each layer of a Neural Network, separately for weights and biases. On top of that, since we have policy and value function networks, we have separate $\gamma_t$ for each of these. After the update on $\gamma_t$, we compute $\theta_{t}(\gamma_t)$ and $\alpha_t(\gamma_t)$, see Section~\ref{sec:map_inference}. After that, we employ the proximal objective~\eqref{eq_app:proximal_objective} with a fixed regularization target $\theta_t(\gamma_t)$. Concretely, we use the update rule~\eqref{eq_app:many_grad_updates} where for each update the gradient is estimate on the batch of data from the replay buffer. This is not exactly the same as what we did with \emph{plasticity benchmarks} since there the update was applied to the same batch of data, multiple times. Nevertheless, we found this strategy effective and easy to implement on top of a SAC algorithm. In practice, we swept over the parameter $s$ (similar for both, policy and the value function) which controls the relative learning rate increase in~\eqref{eq:adapted_lr}. Moreover, we swept over the proximal regularization constant $\tilde{\lambda}$ from eqn.~\eqref{eq_app:proximal_objective}, which was different for the policy and for the value function. In practice, we found that using proximal constant of $0$ for the policy led to the best empirical results. The range for the proximal constants $\tilde{\lambda}$ was $\{0.1, 0.01, 0.001\}$ and for $s$ was $\{0.8, 0.9, 0.95, 0.97, 1.0\}$. We used $p=1$ for all the experiments. For each experiment, we used a $3$ hours of the $A100$ GPU with $40$ Gb of memory.

\begin{figure}[tb]
    \centering
    \includegraphics[scale=0.3]{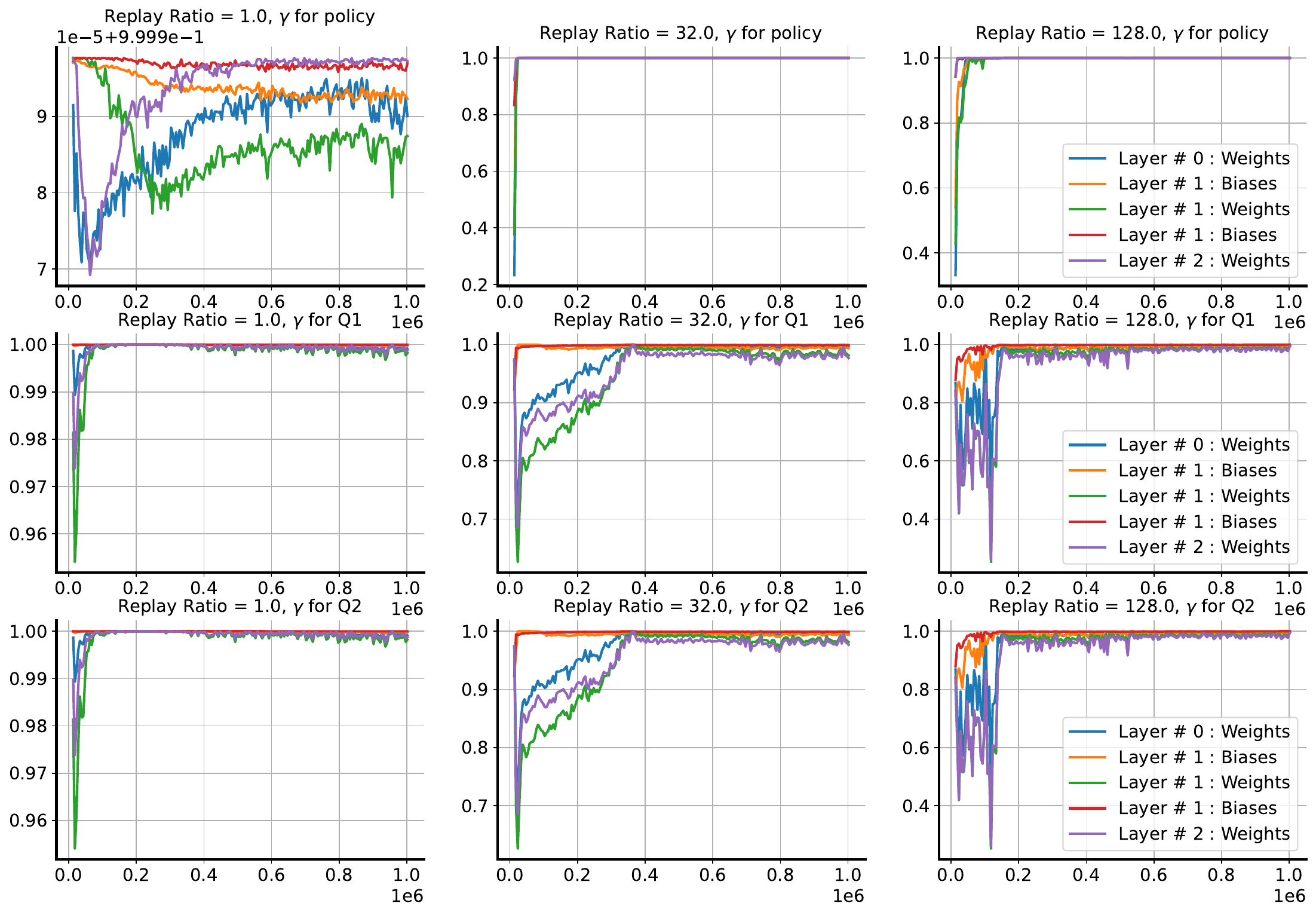}
    \caption{Visualization of the $\gamma_t$ dynamics for the run on Humanoid environment. Each column corresponds to the replay ratio studied. First row denotes the $\gamma_t$ for the policy $\pi$. The second and the third rows denote the $\gamma_t$ for the two $Q$-functions.}
    \label{fig:rl_qualitative}
\end{figure}

\section{Computational complexity}
\label{app:computational_complexity}

We provide the study of computational cost for all the proposed methods. Notations:

\begin{itemize}
    \item $P$ be the number of parameters in the Neural Network
    \item $L$ is the number of layers
    \item $O(S)$ is the cost of SGD backward pass.
    \item $M_{\gamma}$ - number of Monte Carlo samples for the drift model
    \item $M_{\theta}$ - number of Monte Carlo samples for the parameter updates (Bayesian Method).
    \item $K_{\gamma}$ - number of updates for the drift parameter
    \item $K_{\theta}$ - number of NN parameter updates.
\end{itemize}

\begin{table}[ht]
\centering
\begin{tabular}{|c|c|c|}
\hline
\textbf{Method} & \textbf{Comp. cost} & \textbf{Memory} \\
\hline
SGD & $O(S)$ & $O(P)$ \\
\hline
Soft resets $\gamma$ per layer & $O(K_{\gamma} M_{\gamma} S + S)$ & $O(L + (M_{\gamma}+1)P)$ \\
\hline
Soft resets $\gamma$ per param. & $O(K_{\gamma}M_{\gamma} S + S)$ & $O(P + (M_{\gamma}+1)P)$ \\
\hline
Soft resets $\gamma$ per layer + proximal ($K_{\theta}$ iters) & $O(K_{\gamma} M_{\gamma}S + K_{\theta} S)$ & $O(L + (M_{\gamma}+1)P)$ \\
\hline
Soft resets $\gamma$ per param. + proximal ($K_{\theta}$ iters) & $O(K_{\gamma} M_{\gamma}S + K_{\theta} S)$ & $O(P + (M_{\gamma}+1)P)$ \\
\hline
Bayesian Soft Reset Proximal ($K_{\theta}$ iters) $\gamma$ per layer & $O(K_{\gamma} M_{\gamma}S + 2 M_{\theta} K_{\theta} S)$ & $P(L + (M_{\gamma}+2)P)$ \\
\hline
Bayesian Soft Reset Proximal ($K_{\theta}$ iters) $\gamma$ per param. & $O(K_{\gamma} M_{\gamma}S + 2 M_{\theta} K_{\theta} S)$ & $P(P + (M_{\gamma}+2)P)$ \\
\hline
\end{tabular}
\caption{Comparison of methods, computational cost, and memory requirements}
\label{table:theoretical_complexity}
\end{table}

The general theoretical cost of all the proposed approaches is given in Table~\ref{table:theoretical_complexity}. In practice, for all the experiments, we assume that $M_{\gamma} = 1$ and $M_{\theta} = 1$. Moreover, we used $K_\gamma = 1$ and $K_\theta=1$ for \emph{Soft Reset}, $K_\gamma = 10$ and $K_\theta=1$ for \emph{Soft Reset} with more computation. On top of that, for \emph{Soft Reset} proximal and all Bayesian methods, we used $K_\gamma = 10$ and $K_\theta=10$. Table~\ref{table:real_complexity}, quantifying the complexity of all the methods from Figure~\ref{fig:plasticity_benchmarks}.

\begin{table}[ht]
\centering
\begin{tabular}{|c|c|c|}
\hline
\textbf{Method} & \textbf{Comp. cost} & \textbf{Memory} \\
\hline
SGD & $O(S)$ & $O(P)$ \\
\hline
Soft resets $\gamma$ per layer & $O(2S)$ & $O(L + 2P)$ \\
\hline
Soft resets $\gamma$ per param. & $O(2S)$ & $O(3P)$ \\
\hline
Soft resets $\gamma$ per layer + proximal ($K_{\theta}=10$ iters) & $O(20S)$ & $O(L + 2P)$ \\
\hline
Soft resets $\gamma$ per param. + proximal ($K_{\theta}$ iters) & $O(20S)$ & $O(3P)$ \\
\hline
Bayesian Soft Reset Proximal ($K_{\theta}$ iters) $\gamma$ per layer & $O(30S)$ & $P(L + 3P)$ \\
\hline
Bayesian Soft Reset Proximal ($K_{\theta}$ iters) $\gamma$ per param. & $O(30S)$ & $P(4)$ \\
\hline
\end{tabular}
\caption{Comparison of methods, computational cost, and memory requirements for methods in Figure~\ref{fig:plasticity_benchmarks}.}
\label{table:real_complexity}
\end{table}

The complexity $O(2S)$ of Soft Resets comes from one update on drift parameter and one updat eon NN parameters. The memory complexity requires storing $O(L)$ parameters gamma (one for each layer), parameters $\theta_t$ with $O(P)$ and sampled parameters for drift model update which requires $O(P)$.

\begin{figure}[tb]
    \centering
    \includegraphics[scale=0.3]{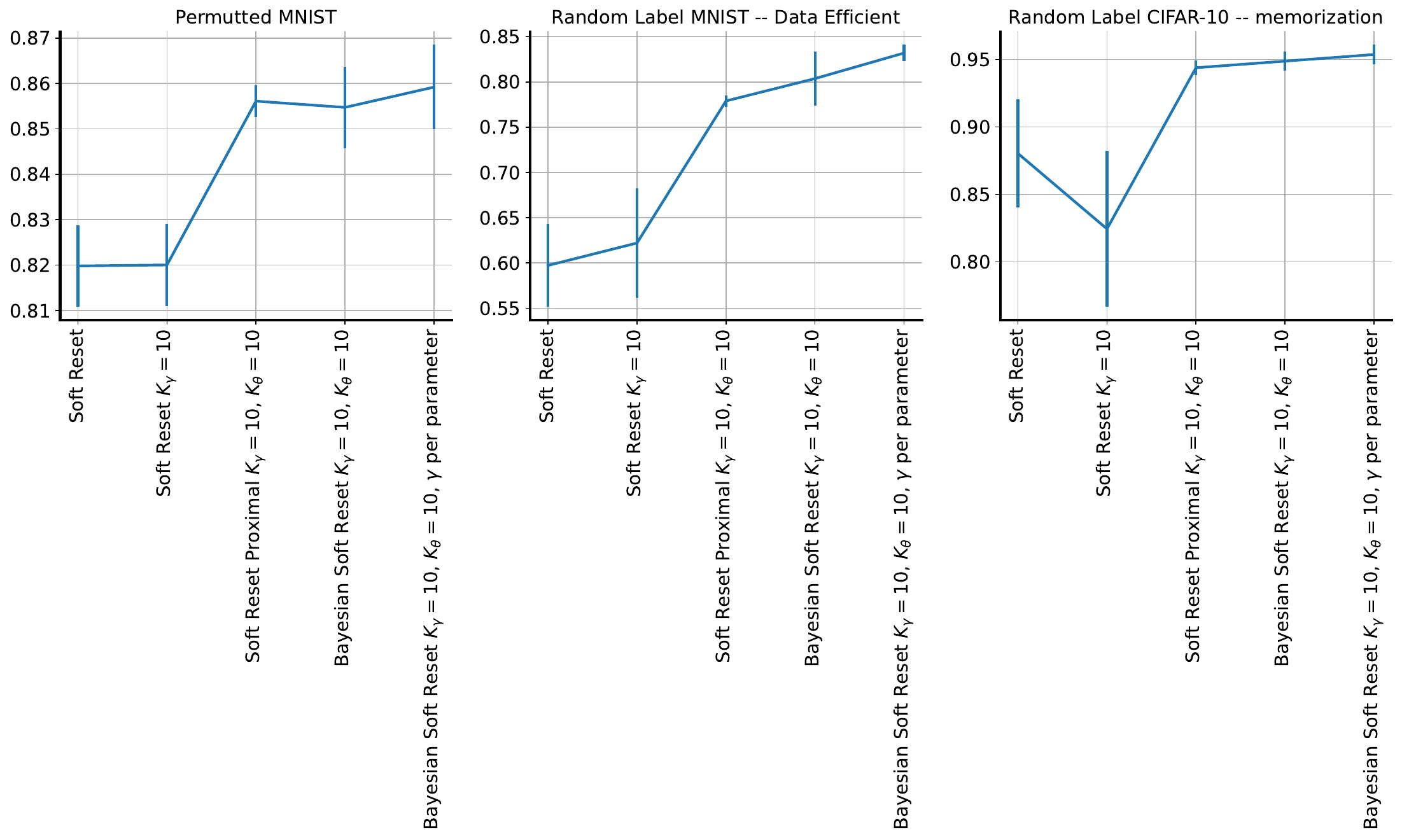}
    \caption{Compute-performance tradeoff. The x-axis indicates the method going from the cheapest (left) to the most expensive (right). See Table~\ref{table:real_complexity} for complexity analysis. The y-axis is the average performance on all the tasks across the stream.}
    \label{fig:compute_plot}
\end{figure}

Note that as Figure~\ref{fig:compute_plot} suggests, it is beneficial to spend more computational cost on optimizing gamma and on doing multiple updates on parameters. However, even the cheapest version of our method \emph{Soft Resets} still leads to a good performance as indicated in Figure~\ref{fig:plasticity_benchmarks}.

The complexity of soft resets in reinforcement learning setting requires only one gradient update on $\gamma$ after each new chunk of fresh data from the environment. In SAC, we do $G$ gradient updates on parameters for every new chunk of data. Assuming that complexity of one gradient update in SAC is $O(S)$, soft reset only requires doing one additional gradient update to fit $\gamma$ parameter.

\begin{table}[ht]
\centering
\begin{tabular}{|c|c|c|}
\hline
\textbf{Method} & \textbf{Comp. cost} & \textbf{Memory} \\
\hline
SAC & $O(G S)$ & $O(P)$ \\
\hline
Soft resets $\gamma$ per layer & $O(S + GS)$ & $O(L + 2P)$ \\
\hline
\end{tabular}
\smallskip
\caption{Comparison of methods, computational cost, and memory requirements for methods in for RL.}
\label{table:rl_complexity}
\end{table}

The computation complexity of Soft Reset in Reinforcement Learning is marginally higher than SAC but leads to better empirical performance in a highly off-policy regime, see Appendix~\ref{app:rl_experiment}.

\section{Sensitivity analysis}
\label{app:sensitivity}

We study the sensitivity of Soft Resets where $\gamma$ is defined per layer when trained on random-label MNIST (data efficient). We fix the learning rate to $\alpha=0.1$. We study the sensitivity of learning rate for the drift parameter, $\eta_{\gamma}$, as well as $p$ -- initial prior standard deviation rescaling, and $s$ -- posterior standard deviation rescaling parameter.

\begin{figure}[!htb]
    \centering
    \includegraphics[scale=0.23]{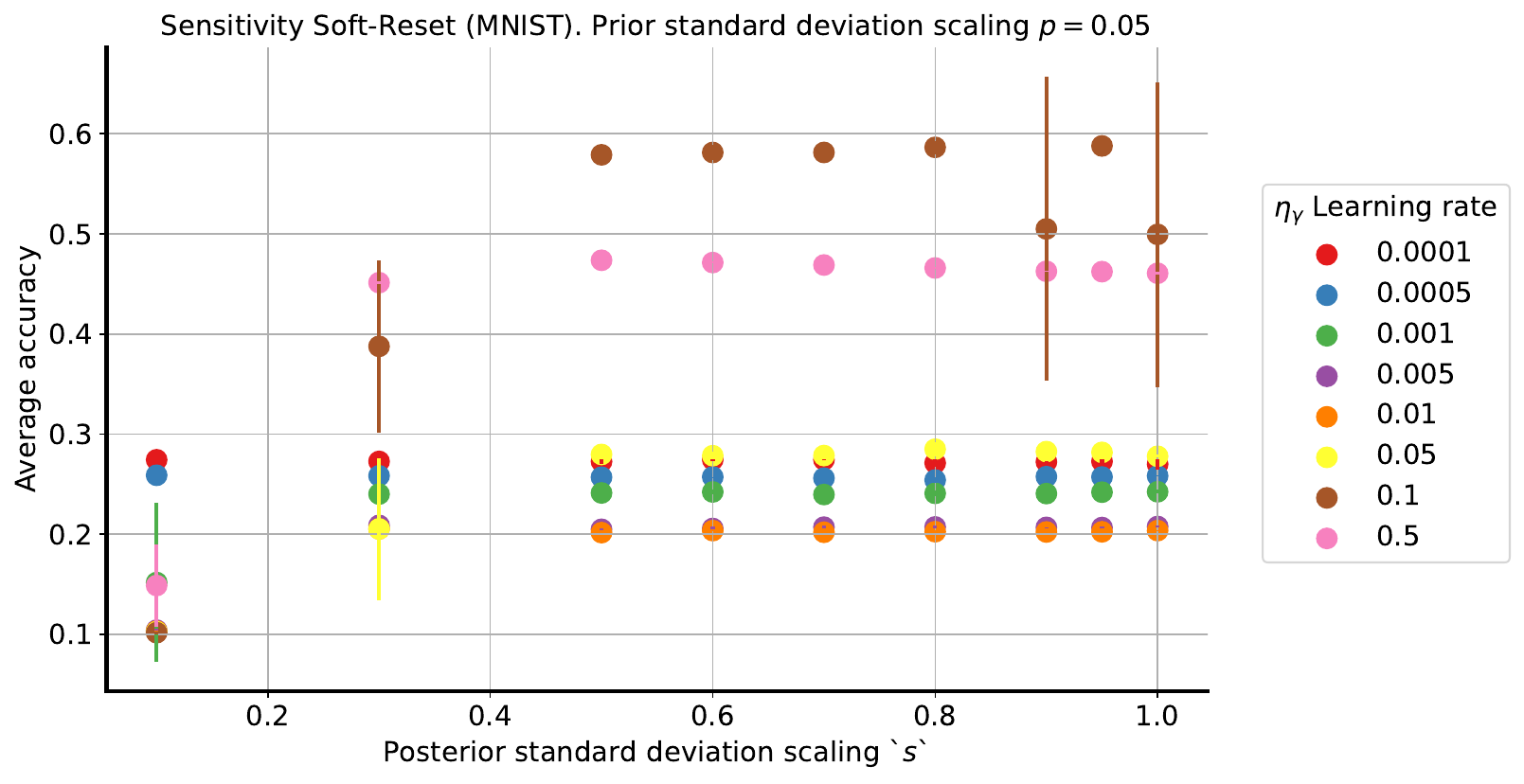}
    \includegraphics[scale=0.23]{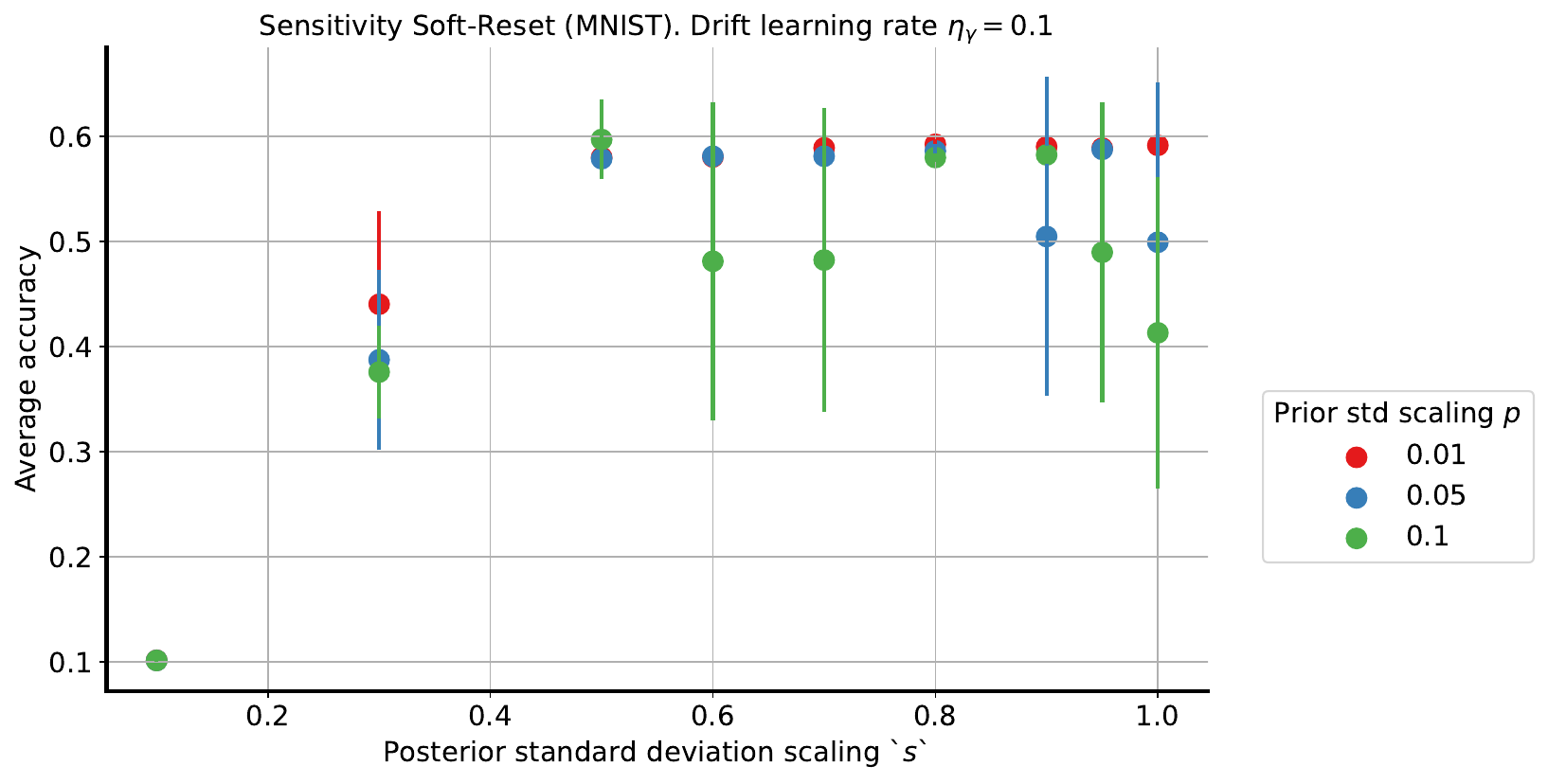}
    \caption{\textbf{\emph{Soft Reset}, sensitivity analysis} of performance with respect to the hyperparameters on data-efficient random-label MNIST. The x-axis denotes the studied hyperparameter, whereas the y-axis denotes the average performance across the tasks. The standard deviation is computed over 3 random seeds. The color  indicates additional studied hyperparameter. \textbf{(Left)} shows sensitivity analysis where the x-axis is the posterior standard deviation scaling $s$ and the color indicates the drift model learning rate $\eta_{\gamma}$. \textbf{(Right)} shows sensitivity of \emph{Soft Reset} where the x-axis is the posterior standard deviation scaling $s$ and the color indicates initial prior standard deviation scaling $p$.}
    \label{fig:sensitivity_soft_reset}
\end{figure}

\begin{figure}[!htb]
    \centering
    \includegraphics[scale=0.25]{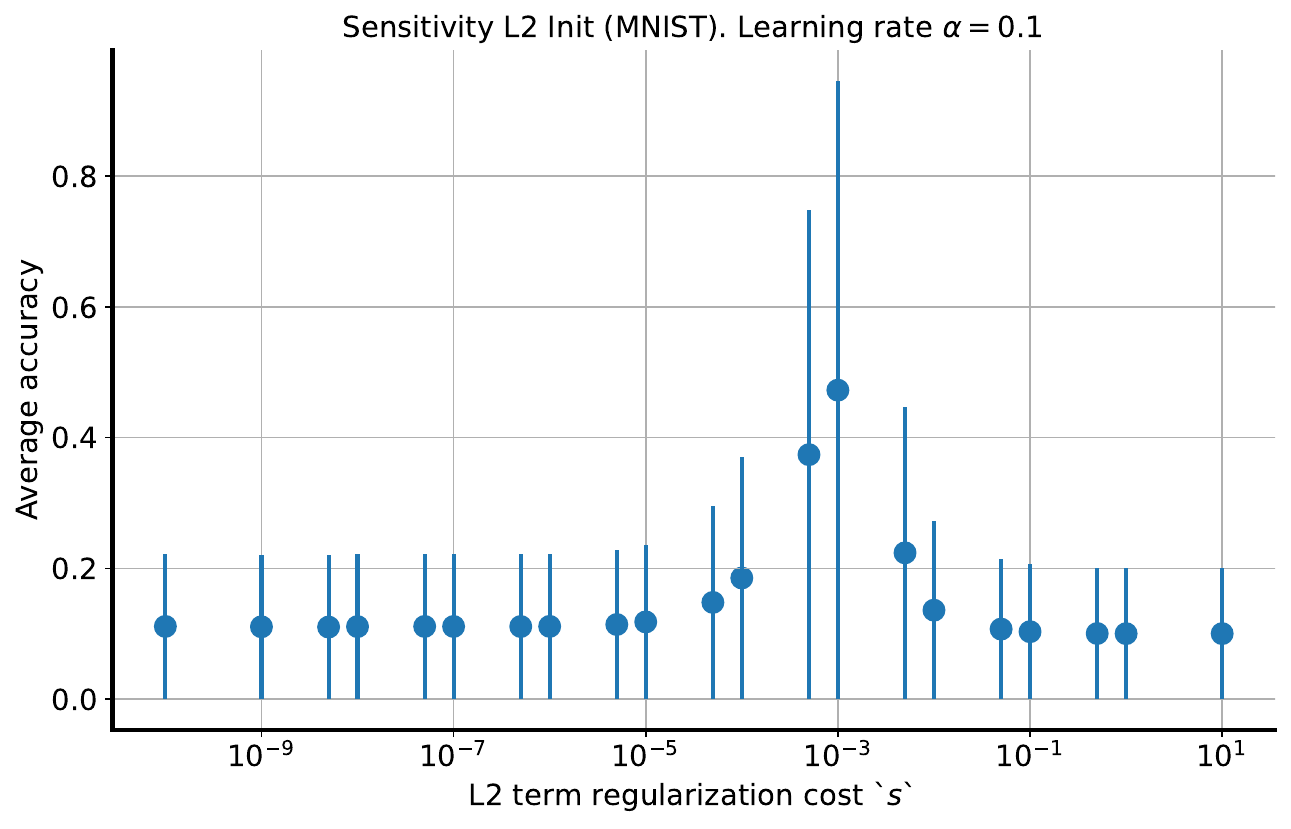}
    \includegraphics[scale=0.25]{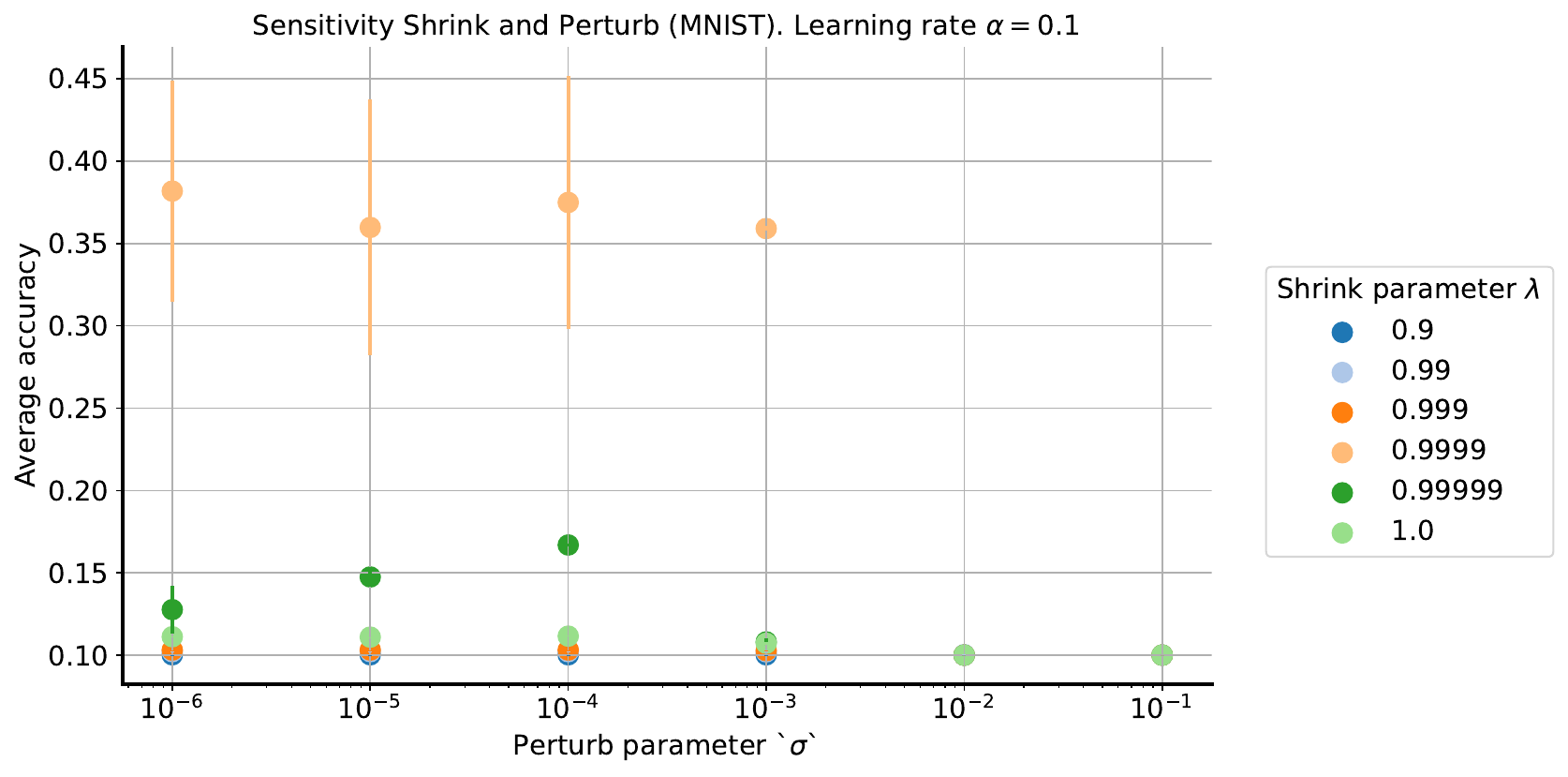}
    \caption{\textbf{L2 Init and Shrink\&Perturb sensitivity analysis} of performance with respect to the hyperparameters on data-efficient random-label MNIST. The x-axis denotes the studied hyperparameter, whereas the y-axis denotes the average performance across the tasks. The standard deviation is computed over 3 random seeds. The color optionally indicates additional studied hyperparameter. \textbf{(Left)} shows sensitivity of \emph{L2 Init} with respect to the $L2$ penalty regularization cost $\lambda$ applied to $||\theta - \theta_0||^2$ term. We do not use an additional hyperparameter, therefore there is only one color. \textbf{(Right)} shows sensitivity of \emph{Shrink\&Perturb} method where the x-axis is the perturb parameter $\sigma$ while the color indicates the shrink parameter $\lambda$.}
    \label{fig:sensitivity_baselines}
\end{figure}

On top of that, we conduct the sensitivity analysis of L2 Init~\cite{kumar2023maintaining} and Shrink\&Perturb~\citep{ash2020warmstarting} methods. The x-axis of each plot denotes one of the studied hyperparameters, whereas y-axis is the average performance across all the tasks (see Experiments section for tasks definition). The standard deviation is reported over 3 random seeds. A color indicates a second hyperparameter which is studied, if available. In the title of each plot, we write hyperparameters which are fixed. The analysis is provided in Figure~\ref{fig:sensitivity_soft_reset} for \emph{Soft Resets} and in Figure~\ref{fig:sensitivity_baselines} for the baselines.

The most important parameter is the learning rate of the drift model $\eta_{\gamma}$. For each method, there exists a good value of this parameter and performance is sensitive to it. This makes sense since this parameter directly impacts how we learn the drift model.

The performance of Soft Resets is robust with respect to the posterior standard deviation scaling $s$ parameter as long as it is $s \geq 0.5$. For $s < 0.5$, the performance degrades. This parameter is defined from $\sigma_t = s \sigma_0$ and affects relative increase in learning rate given by $\frac{1}{\gamma^2 + (1-\gamma^2)/s^2)}$ which could be ill-behaved for small $s$.

We also study the sensitivity of the baseline methods. We find that L2 Init~\citep{kumar2023maintaining} is very sensitive to the parameter $\lambda$, which is a penalty term for $\lambda ||\theta - \theta_0||^2$. In fact, Figure~\ref{fig:sensitivity_baselines}, left shows that there is only one good value of this parameter which works. Shrink\&Perturb~\citep{ash2020warmstarting} is very sensitive to the shrink parameter $\lambda$. Similar to L2 Init, there is only one value which works, 0.9999 while values 0.999 and values 0.99999 lead to bad performance. This method however, is not very sensitive to the perturb parameter $\sigma$ provided that $\sigma \leq 0.001$.

\begin{figure}[!htb]
    \centering
    \includegraphics[scale=0.23]{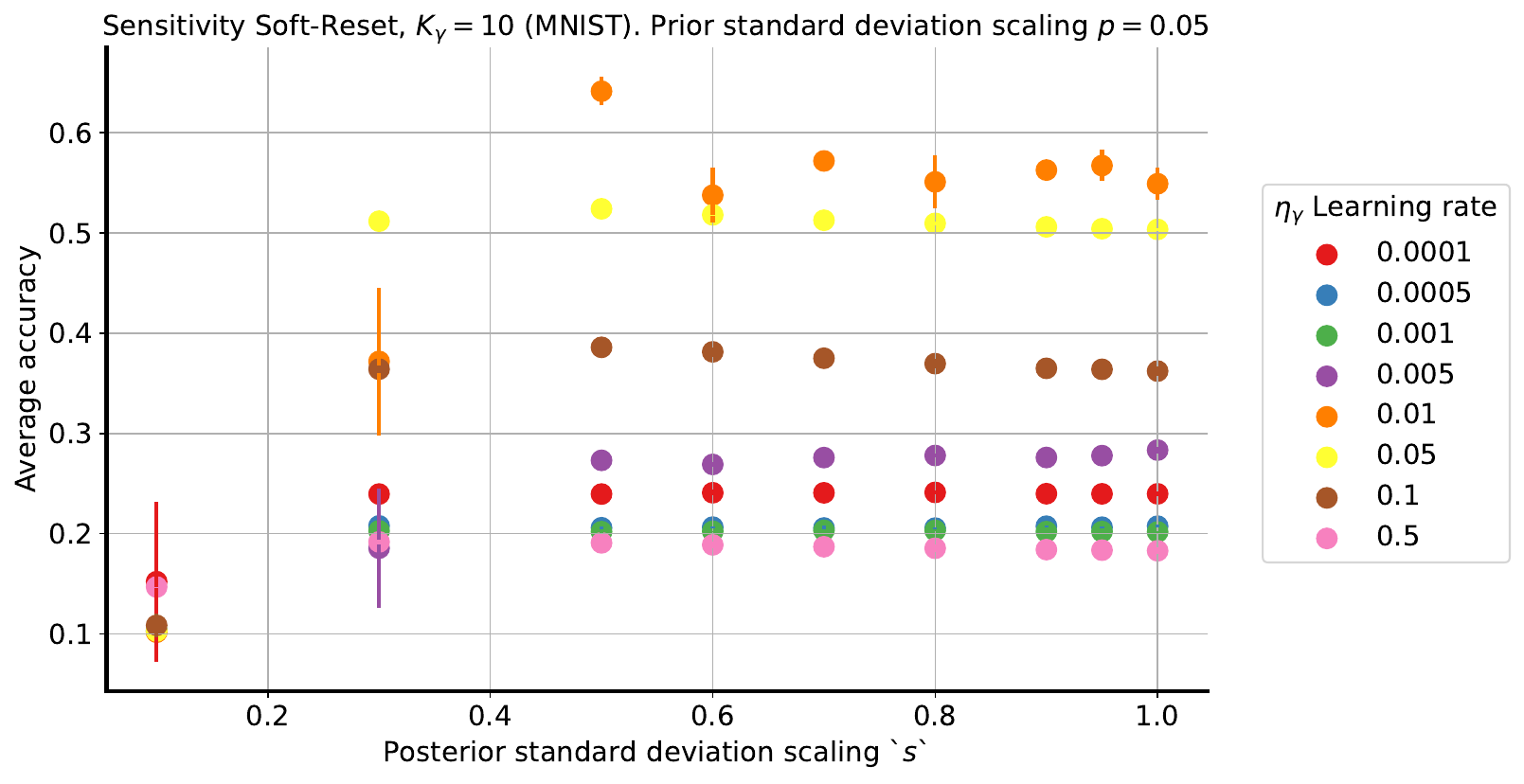}
    \includegraphics[scale=0.23]{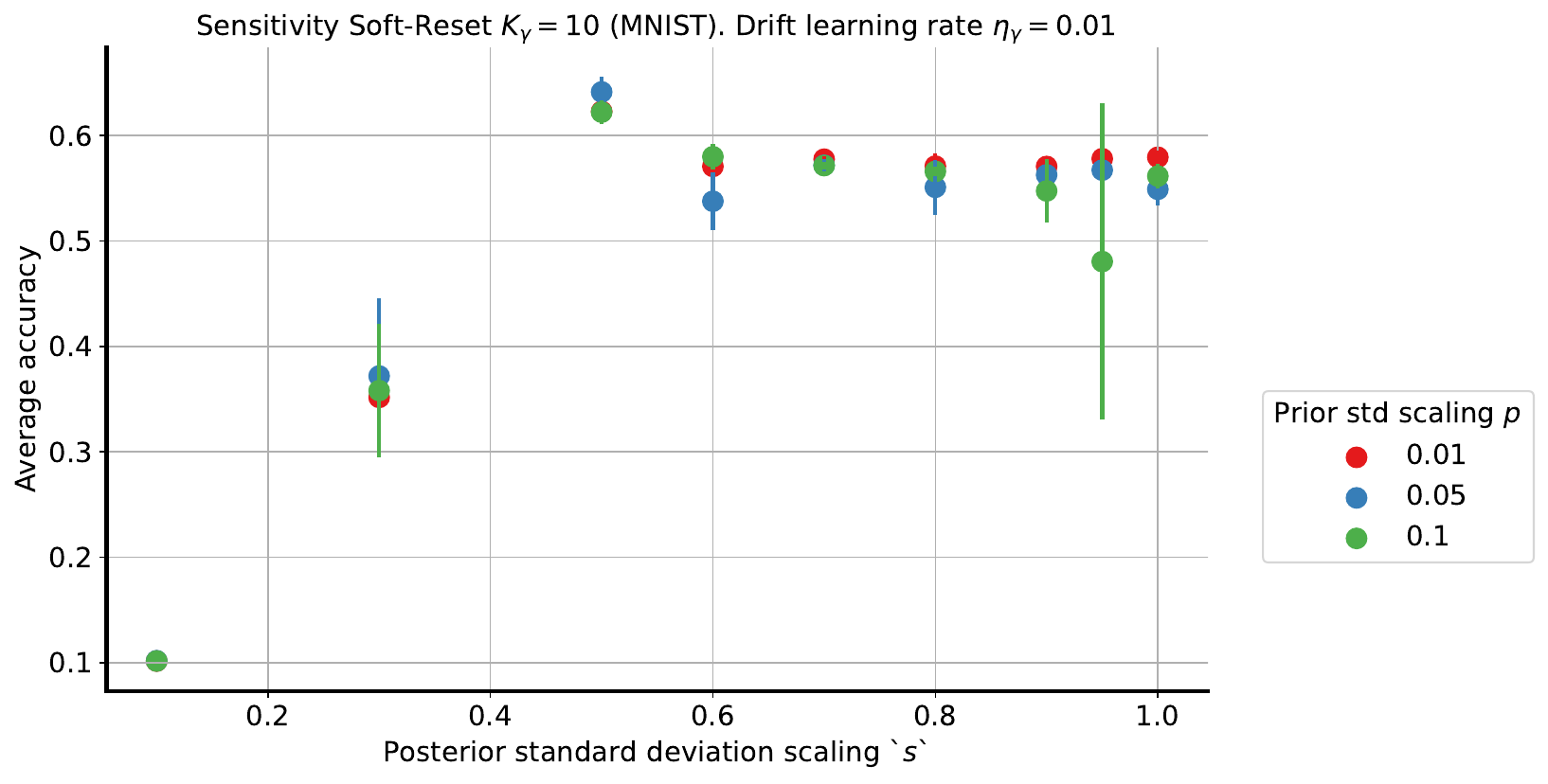}
    \caption{\textbf{\emph{Soft Reset}, $K_\gamma=10$, sensitivity analysis} of performance with respect to the hyperparameters on data-efficient random-label MNIST. The x-axis denotes the studied hyperparameter, whereas the y-axis denotes the average performance across the tasks. The standard deviation is computed over 3 random seeds. The color  indicates additional studied hyperparameter. \textbf{(Left)} shows sensitivity analysis where the x-axis is the posterior standard deviation scaling $s$ and the color indicates the drift model learning rate $\eta_{\gamma}$. \textbf{(Right)} shows sensitivity analysis where the x-axis is the posterior standard deviation scaling $s$ and the color indicates initial prior standard deviation scaling $p$.}
    \label{fig:sensitivity_soft_reset_more_compute}
\end{figure}

\begin{figure}[!htb]
    \centering
    \includegraphics[scale=0.23]{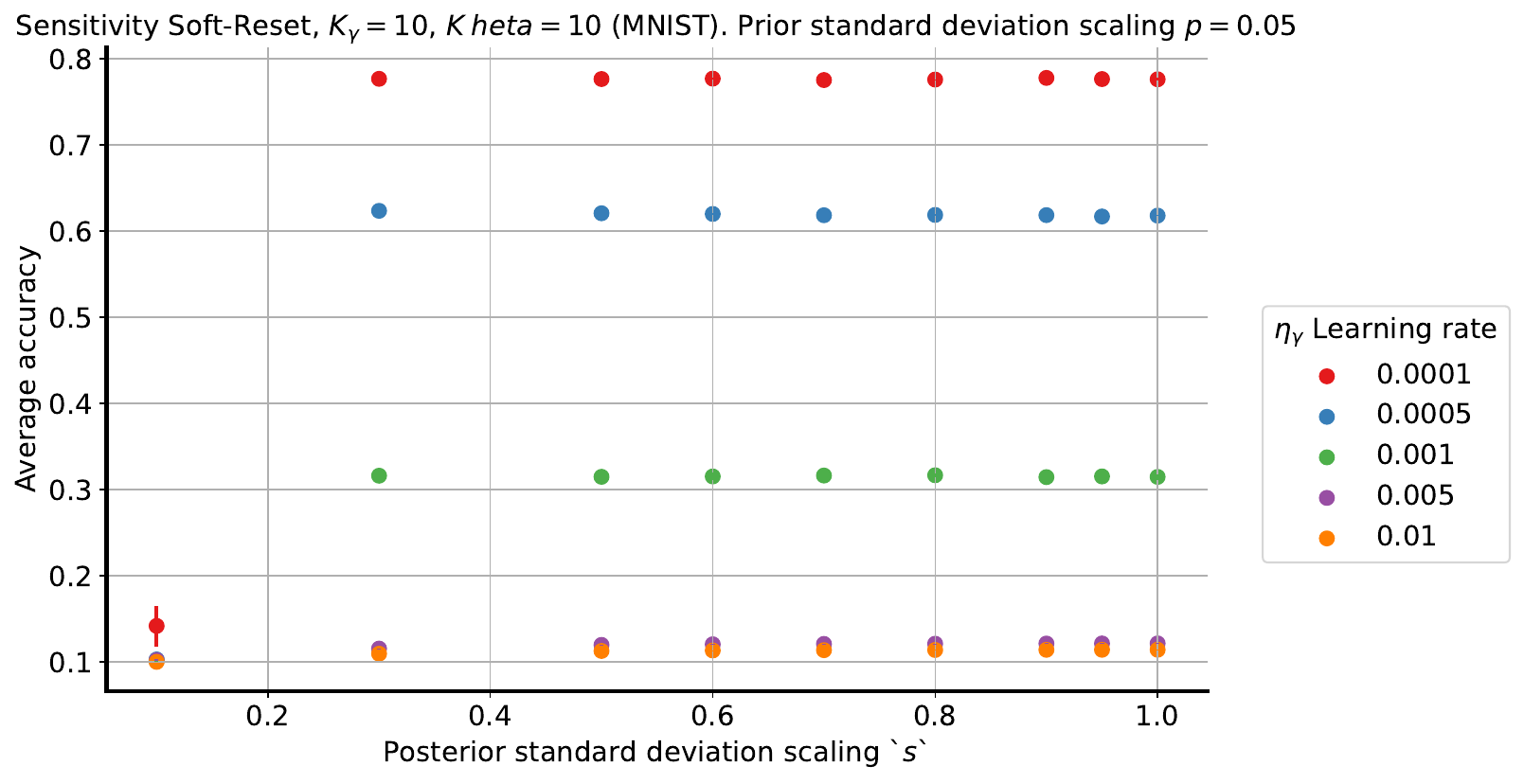}
    \includegraphics[scale=0.23]{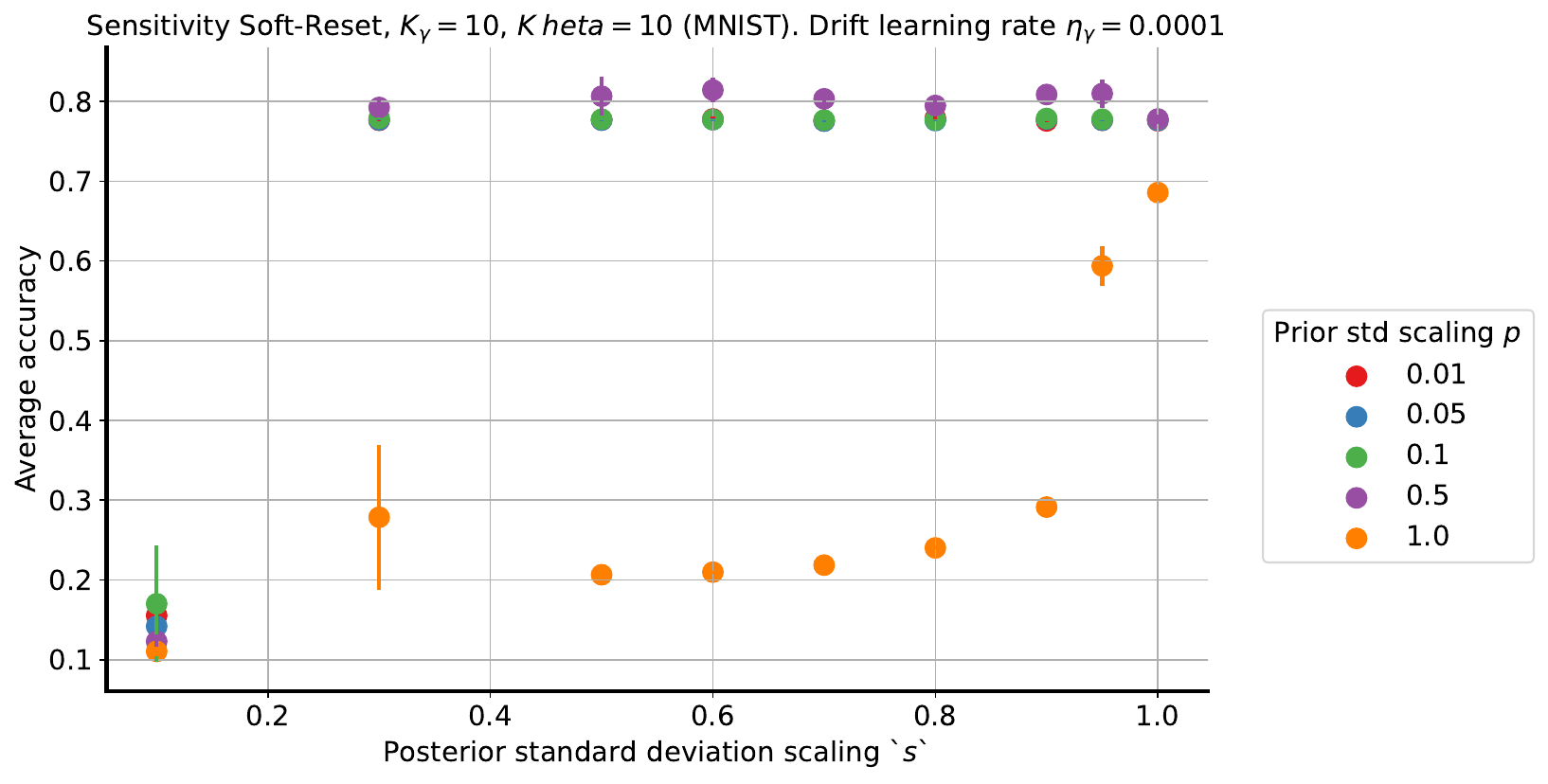}
    \caption{\textbf{\emph{Soft Reset}, $K_\gamma=10, K_\theta=10$, sensitivity analysis} of performance with respect to the hyperparameters on data-efficient random-label MNIST. The x-axis denotes the studied hyperparameter, whereas the y-axis denotes the average performance across the tasks. The standard deviation is computed over 3 random seeds. The color  indicates additional studied hyperparameter. \textbf{(Left)} shows sensitivity analysis where the x-axis is the posterior standard deviation scaling $s$ and the color indicates the drift model learning rate $\eta_{\gamma}$. \textbf{(Right)} shows sensitivity analysis where the x-axis is the posterior standard deviation scaling $s$ and the color indicates initial prior standard deviation scaling $p$.}
    \label{fig:sensitivity_soft_reset_proiximal}
\end{figure}

\begin{figure}[!htb]
    \centering
    \includegraphics[scale=0.25]{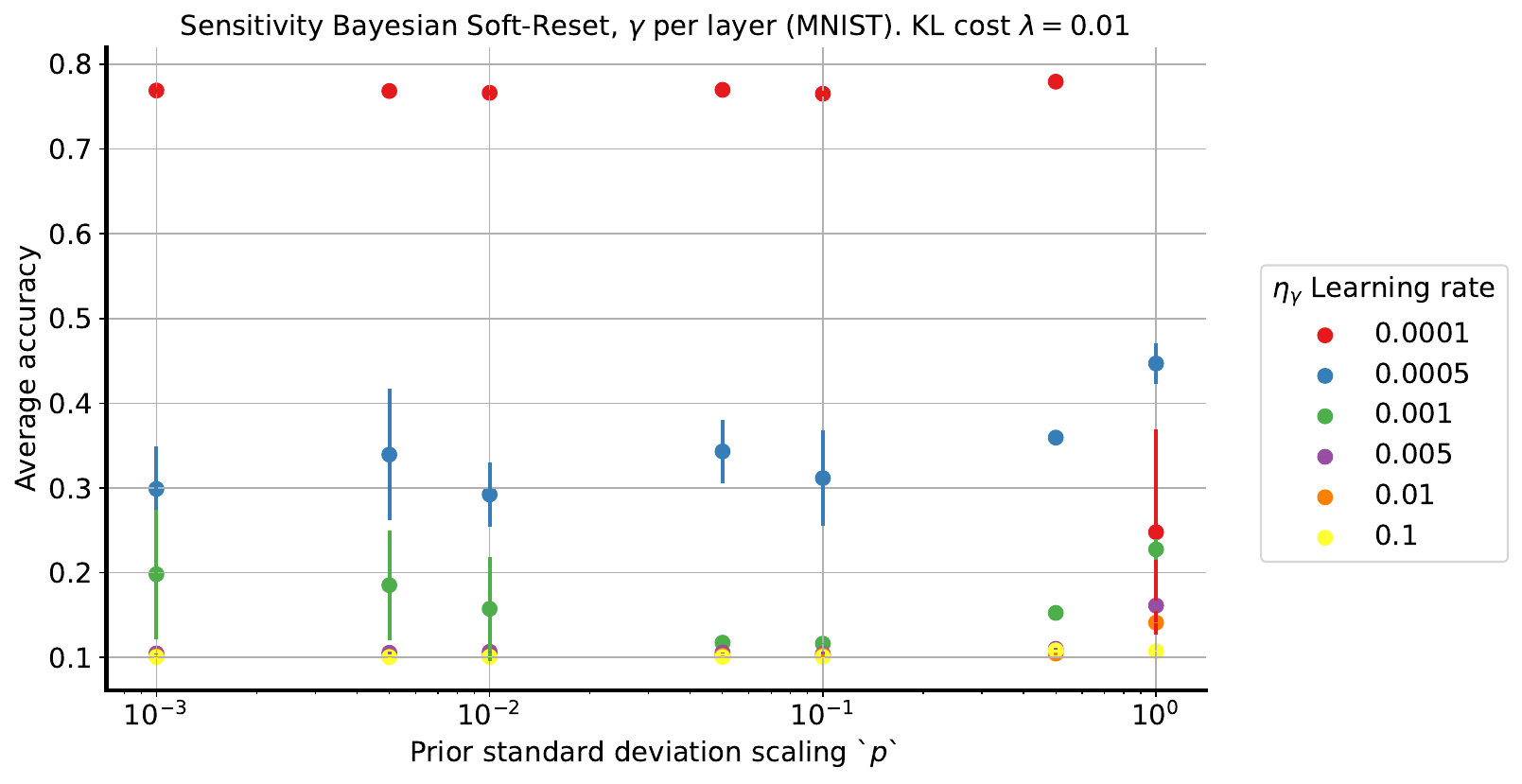}
    \includegraphics[scale=0.25]{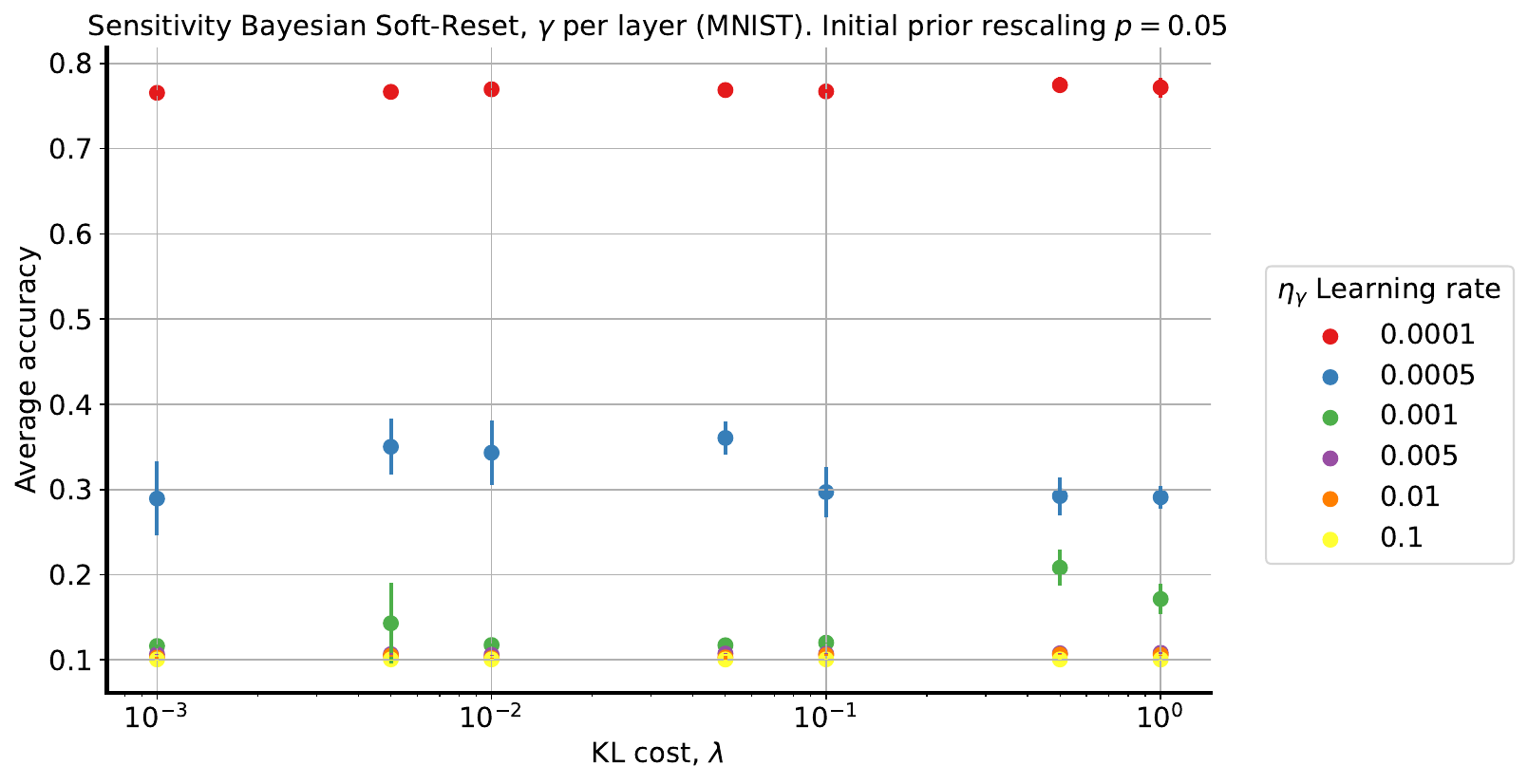}
    \caption{\textbf{\emph{Bayesian Soft Reset}, $K_\gamma=10, K_\theta=10$ with $\gamma_t$ per layer, sensitivity analysis} of performance with respect to the hyperparameters on data-efficient random-label MNIST. The x-axis denotes the studied hyperparameter, whereas the y-axis denotes the average performance across the tasks. The standard deviation is computed over 3 random seeds. The color  indicates additional studied hyperparameter. \textbf{(Left)} shows sensitivity analysis where the x-axis is the prior standard deviation initial scaling $p$ and the color indicates the drift model learning rate $\eta_{\gamma}$. \textbf{(Right)} shows sensitivity analysis where the x-axis is the KL divergence coefficient $\lambda$ while the color indicates the learning rate $\eta_{\gamma}$.}
    \label{fig:sensitivity_bayesian_soft_reset_per_layer}
\end{figure}

\begin{figure}[!htb]
    \centering
    \includegraphics[scale=0.25]{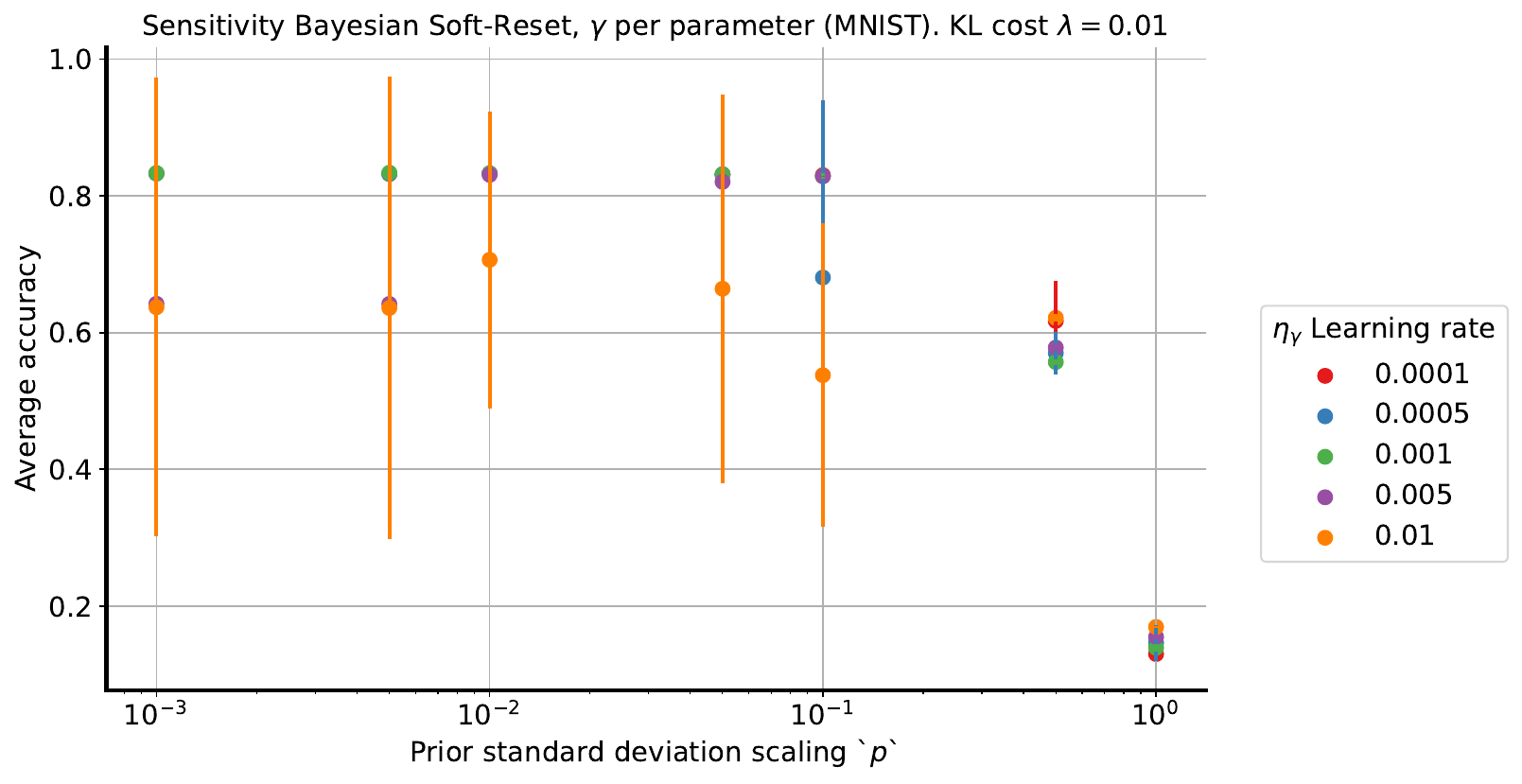}
    \includegraphics[scale=0.25]{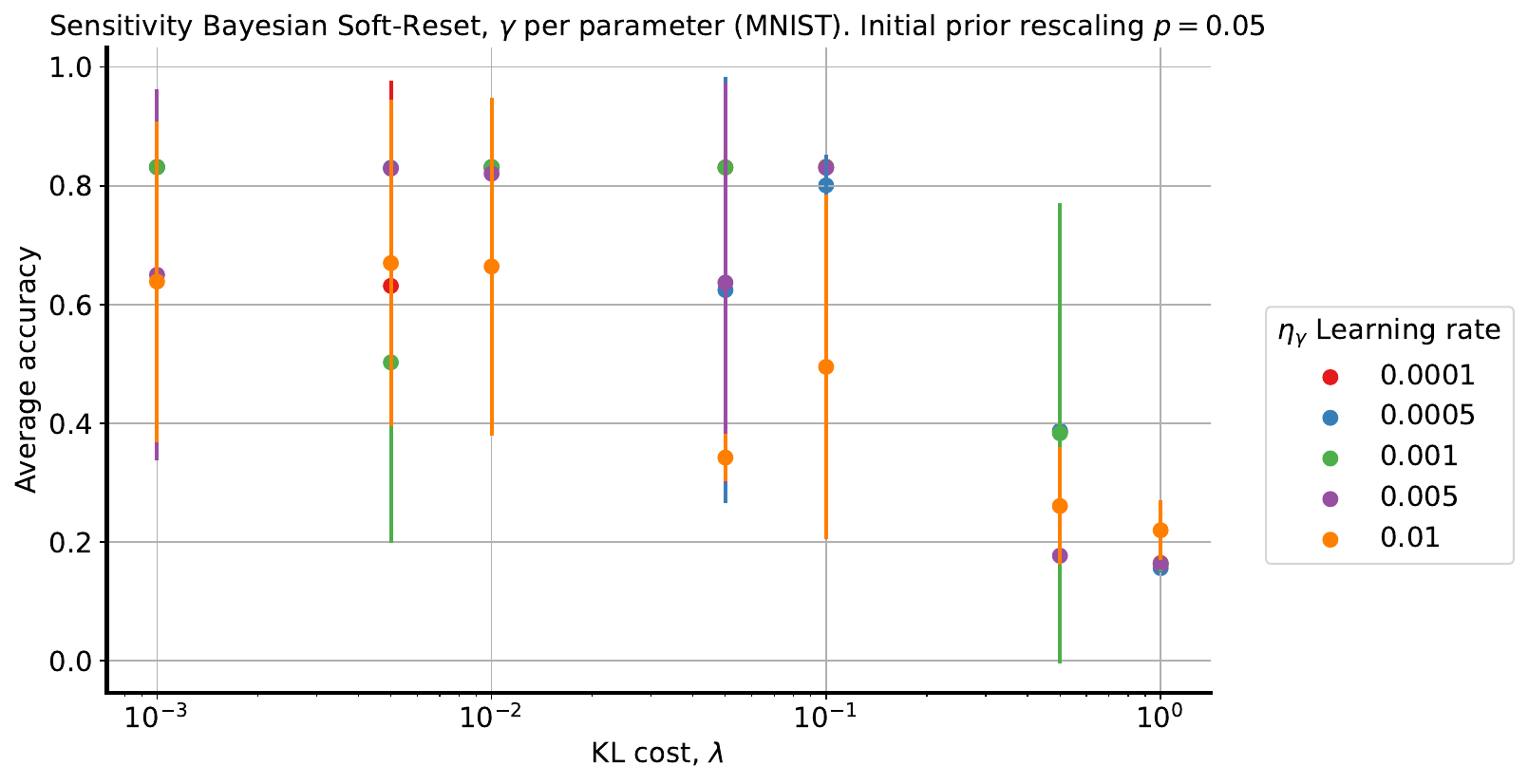}
    \caption{\textbf{\emph{Bayesian Soft Reset}, $K_\gamma=10, K_\theta=10$ with $\gamma_t$ per parameter, sensitivity analysis} of performance with respect to the hyperparameters on data-efficient random-label MNIST. The x-axis denotes the studied hyperparameter, whereas the y-axis denotes the average performance across the tasks. The standard deviation is computed over 3 random seeds. The color  indicates additional studied hyperparameter. \textbf{(Left)} shows sensitivity analysis where the x-axis is the prior standard deviation initial scaling $p$ and the color indicates the drift model learning rate $\eta_{\gamma}$. \textbf{(Right)} shows sensitivity analysis where the x-axis is the KL divergence coefficient $\lambda$ while the color indicates the learning rate $\eta_{\gamma}$.}
    \label{fig:sensitivity_bayesian_soft_reset_per_parameter}
\end{figure}

Compared to the baselines, our method is more robust to the hyperparameters choice. Below, we also add sensitivity analysis for other method variants. Figure~\ref{fig:sensitivity_soft_reset_more_compute} shows sensitivity of \emph{Soft Resets}, $K_\gamma=10$, Figure~\ref{fig:sensitivity_soft_reset_proiximal} shows sensitivity of \emph{Soft Resets}, $K_\gamma=10$, $K_\theta=10$, Figure~\ref{fig:sensitivity_bayesian_soft_reset_per_layer} shows sensitivity of \emph{Bayesian Soft Resets}, $K_\gamma=10$, $K_\theta=10$ with $\gamma_t$ per layer, Figure~\ref{fig:sensitivity_bayesian_soft_reset_per_parameter} shows sensitivity of \emph{Bayesian Soft Resets}, $K_\gamma=10$, $K_\theta=10$ with $\gamma_t$ per parameter.

\section{Proximal SGD}
\label{app:proximal_sgd}

Each step of online SGD can be seen in terms of a regularized minimization problem referred to as the proximal form~\citep{proximal_boyd}:
\begin{equation}
    \textstyle
    \label{eq:sgd_proximal}
    \hat{\theta}_{t+1} = \arg\min_{\theta} \mathcal{L}_{t+1}(\theta) + \frac{1}{2 \alpha_t} ||\theta - \theta_{t}||^2.
\end{equation}
In general, we cannot solve~\eqref{eq:sgd_proximal} directly, so  we  consider a
Taylor expansion of $\mathcal{L}_{t+1}$ around $\theta_t$, giving
\begin{equation}
    \textstyle
    \label{eq:sgd_as_mirror_descent}
    \theta_{t+1} =  \arg\min_{\theta} \nabla_{\theta} \mathcal{L}_{t+1}(\theta_t)^\top (\theta - \theta_t) + \frac{1}{2 \alpha_t} ||\theta_t - \theta||^2.
\end{equation}
Here we see the role of $\alpha_t>0$ as both enforcing that the Taylor expansion around $\theta_t$ is accurate, and regularising $\theta_{t+1}$ towards the old parameters $\theta_t$ (hence ensuring that the learning from past data is not forgotten). Solving \eqref{eq:sgd_as_mirror_descent}
naturally leads to the well known SGD update:
\begin{equation}
    \textstyle
    \label{eq:sgd}
    \theta_{t+1} = \theta_{t} - \alpha_{t} \nabla_{\theta} \mathcal{L}_{t+1}(\theta_t),
\end{equation}
where $\alpha_{t}$ can now also be interpreted as the learning rate.

\section{Qualitative behavior of soft resets and additional results on Plasticity benchmarks}
\label{app:bayesian_is_better}

\subsection{Perfect Soft Resets}
\label{app:perfect_soft_resets}

To understand the impact of drift model~\eqref{eq:ou_model}, we study the \emph{data efficient} random-label MNIST setting where task boundaries are known. We run \emph{Online SGD}, \emph{Hard Reset} which resets all parameters at task boundaries, and \emph{Hard Reset (only last)} which resets only the last layer. We use  \emph{Soft Reset} method~\eqref{eq:linearised_map} where $\gamma_t = 1$ all the time and becomes $\gamma_t = \hat{\gamma_t}$ (with manually chosen $ \hat{\gamma_t}$) at task boundaries. We consider constant learning rate $\alpha_t(\gamma_t)$ and increasing learning rate~\eqref{eq:adapted_lr} at task boundary for \emph{Soft Reset}. On top of that, we run \emph{Soft Reset} method unaware of task boundaries which learns $\gamma_t$. We report \emph{Average training task accuracy} metric in Figure~\ref{fig:perfect_resets_ablation}. See Appendix~\ref{app:plasticity_experiment} for details. The results suggest that with the appropriate choice of $\hat{\gamma_t}$, \emph{Soft Reset} is much more efficient than \emph{Hard Reset} and the effect becomes stronger if the learning rate $\alpha_t(\gamma_t)$ increases. We also see that \emph{Soft Reset} could learn an appropriate $\gamma_t$ without the knowledge of task boundary.

\begin{figure}[tb]
    \centering
    \includegraphics[scale=0.3]{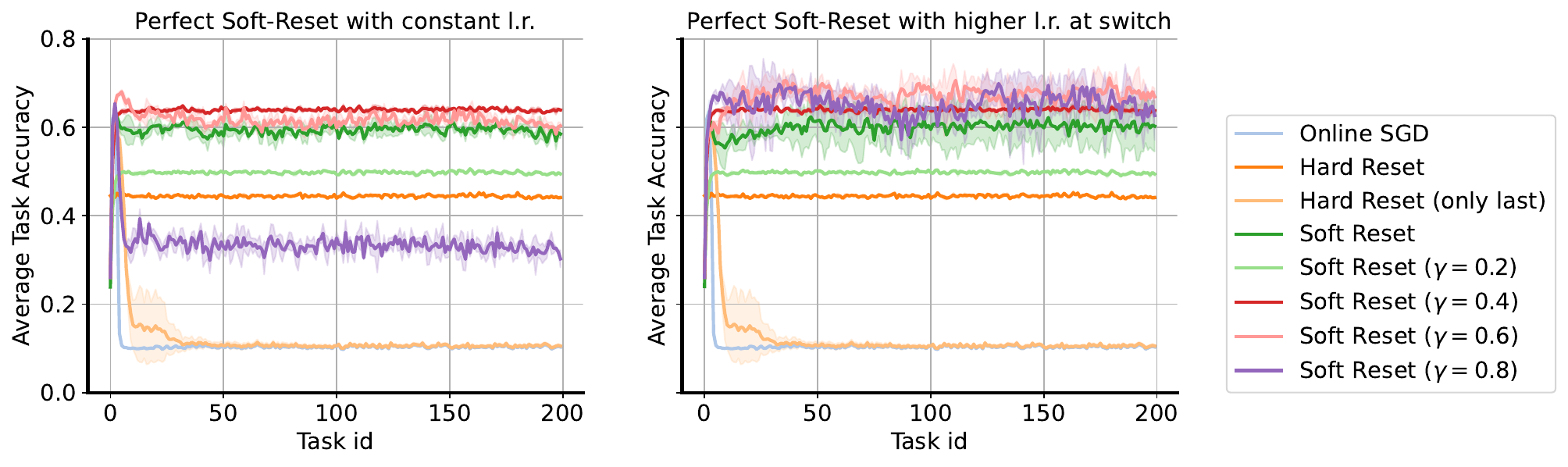}
    \caption{\textbf{Perfect soft-resets} on \emph{data-efficient} random-label MNIST. \emph{Left}, \emph{Soft Reset} method does not use higher learning rate when $\gamma < 1$. \emph{Right}, \emph{Soft Reset} increases the learning rate when $\gamma<1$, see~\eqref{eq:adapted_lr}. The x-axis represents task id, whereas the y-axis is the average training accuracy on the task.}
    \label{fig:perfect_resets_ablation}
\end{figure}

\subsection{Qualitative Behaviour on \emph{Soft Resets} on random-label tasks.}
\label{app:qualitative_behavior_of_soft_resets_on_random_label_tasks}

\begin{figure}[tb]
    \centering
    \includegraphics[scale=0.2]{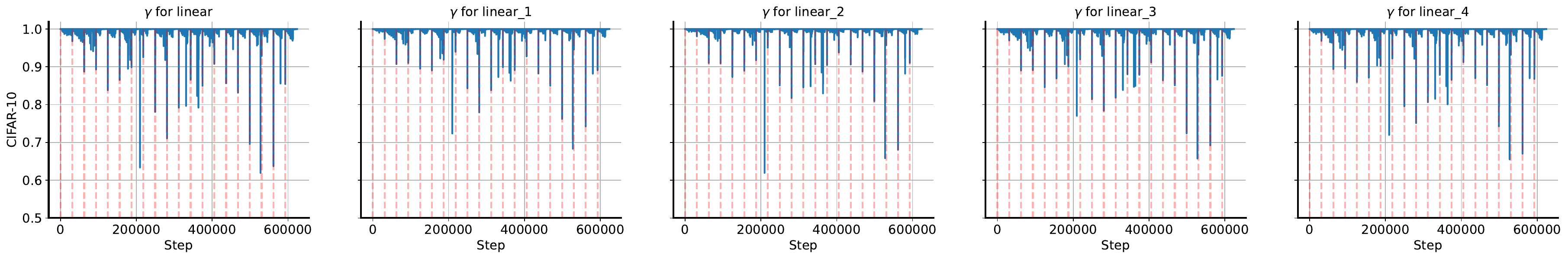}
    \caption{Behaviour of $\gamma_t$ for different layers on random-label MNIST (data efficient) for the first 20 tasks.}
    \label{fig:detailed_results_mlp_random_label_mnist}
\end{figure}

\begin{figure}[tb]
    \centering
    \includegraphics[scale=0.2]{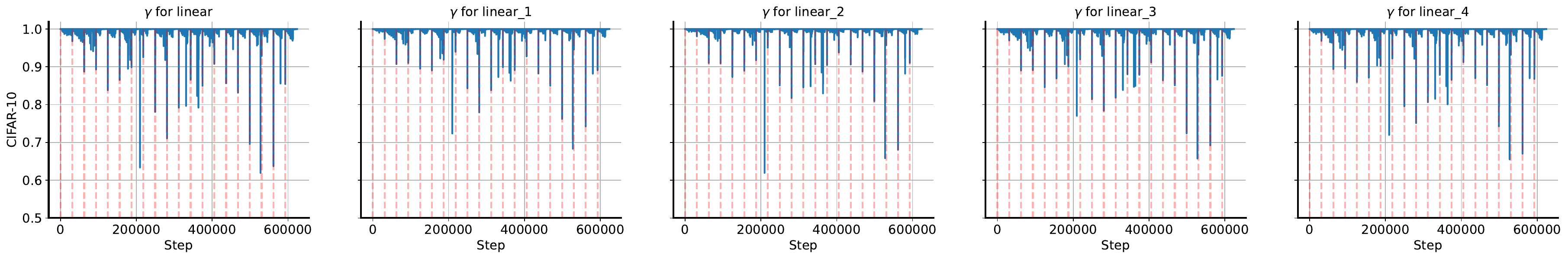}
    \caption{Behaviour of $\gamma_t$ for different layers on random-label CIFAR-10 (memorization) for the first 20 tasks.}
    \label{fig:detailed_results_mlp_random_label_cifar}
\end{figure}

 We observe what values of $\gamma_t$ we get as we train \emph{Soft Reset} method on random-label MNIST (data-efficient) and CIFAR-10 (memorization). The results are given in Figure~\ref{fig:detailed_results_mlp_random_label_mnist} for MNIST and in Figure~\ref{fig:detailed_results_mlp_random_label_cifar} for CIFAR-10. We report these for the first $20$ tasks.

\subsection{Qualitative Behaviour on \emph{Soft Resets} on permuted patches of MNIST.}
\label{app:qualitative_behavior_of_soft_resets_on_permutted_patches_mnist}

\begin{figure}[tb]
    \centering
    \includegraphics[scale=0.18]{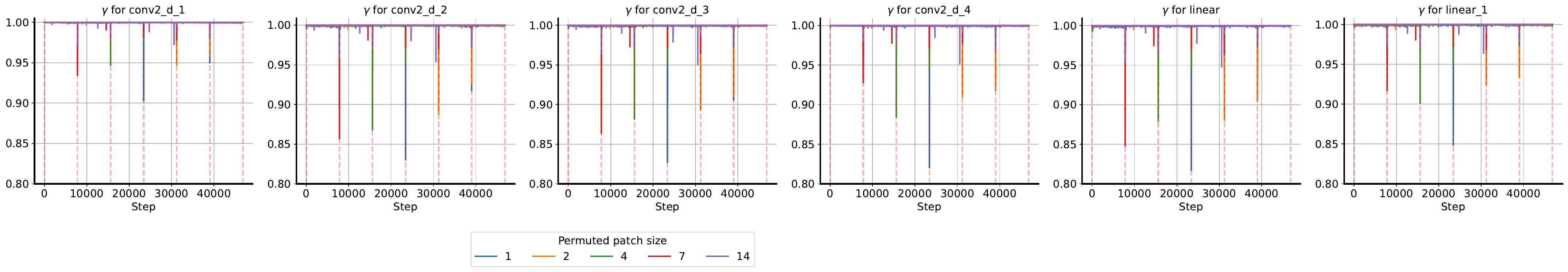}
    \caption{Behaviour of $\gamma_t$ for different layers on permuted MNIST}
    \label{fig:permuted_mnist_patch_size_qualitative}
\end{figure}

We consider a version of permuted MNIST where instead of permuting all the pixels, we permute patches of pixels with a patch size varying from $1$ to $14$. The patch size of $1$ corresponds to permututed MNIST and therefore the most non-stationary case, while patch size of $14$ corresponds to least non-stationary case. We use a convolutional Neural Network in this case. In Figure~\ref{fig:permuted_mnist_patch_size_qualitative}, we report the behavior of $\gamma$ for different convolutional and fully connected layers on first few tasks.

\subsection{Bayesian method is better than non-Bayesian}
\label{app:bayesian_method_is_better_than_non_bayesian}

\begin{figure}[!htb]
    \centering
    \includegraphics[scale=0.4]{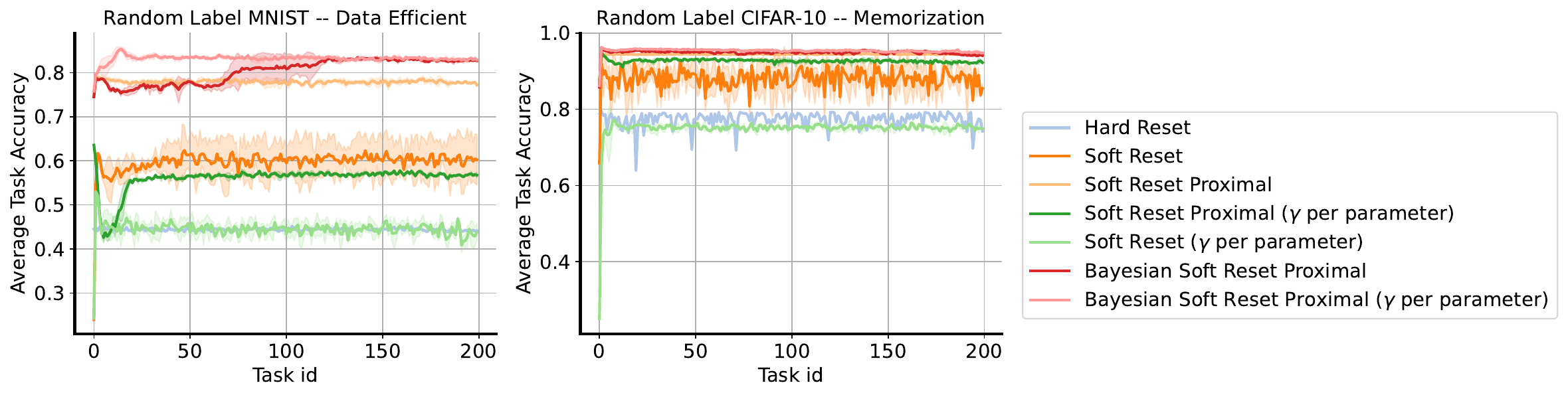}
    \caption{Performance of $\gamma$ per-parameter methods}
    \label{fig:app_per_parameter}
\end{figure}

As discussed in Section~\ref{sec:experiments}, we found that in practice \emph{Soft Reset} and \emph{Soft Reset Proximal} where $\gamma$ is learned per-parameter, did not perform well on the plasticity benchmarks. However, the Bayesian variant described in Section~\ref{sec:bnns}, actually benefited from specifying $\gamma$ for every parameter in Neural Network. We report these additional results in Figure~\ref{fig:app_per_parameter}. We see that the non Bayesian variants where $\gamma_t$ is specified per parameter, do not perform well. The fact that the Bayesian method performs better here suggests that it is important to have a good uncertainty estimate $\sigma^2_t$ for the update~\eqref{eq:drift_via_predictive_ll} on $\gamma_t$. When, however, we regularize $\gamma_t$ to be shared across all parameters within each layer, this introduces useful inductive bias which mitigates the lack of uncertainty estimation in the parameters. This is because for non-Bayesian methods, we assume that the uncertainty is fixed, given by a hyperparameter -- assumption which would not always hold in practice.

\subsection{Qualitative behavior of soft resets}
\label{app:qualitative_behavior_of_soft_resets}
\begin{figure}[!htb]
    \centering
    \includegraphics[scale=0.3]{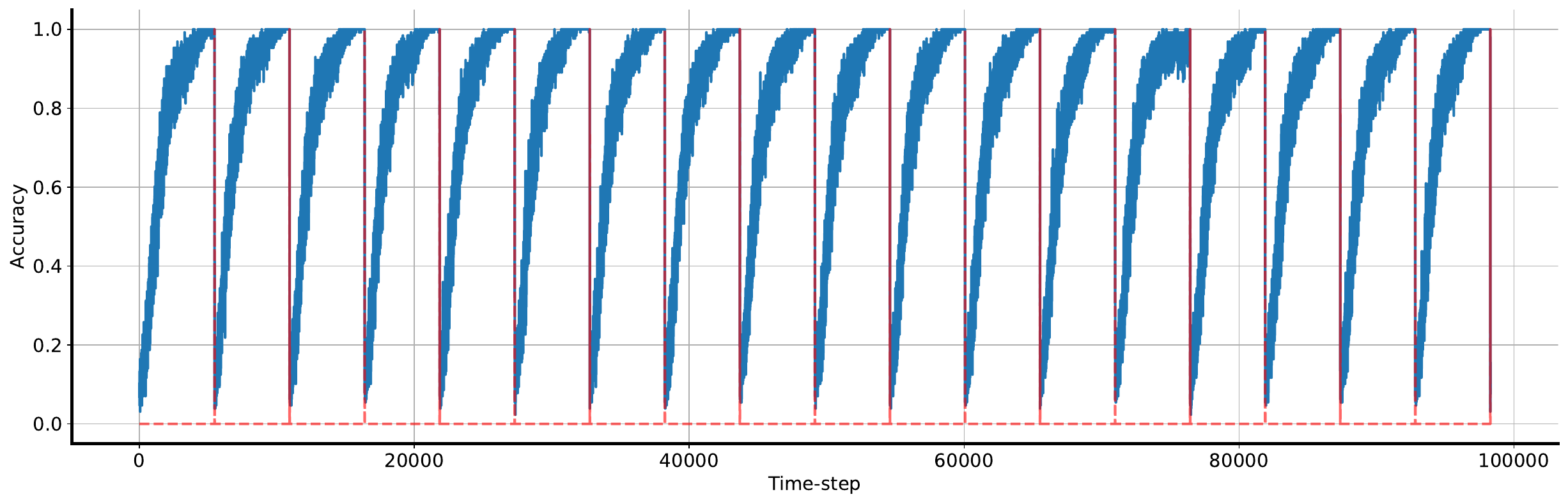}
    \caption{Visualization of accuracy when trained on \emph{data efficient} random-label MNIST task. The dashed red lines correspond to a task boundary.}
    \label{fig:app_accuracy}
\end{figure}

\begin{figure}[!htb]
    \centering
    \includegraphics[scale=0.14]{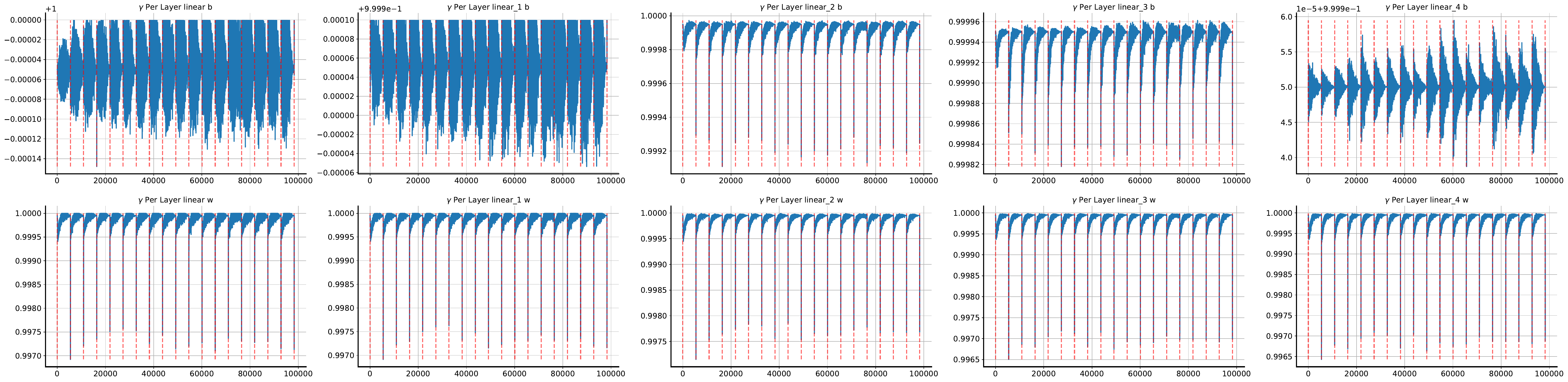}
    \caption{Visualization of $\gamma$ and task boundaries on \emph{data-efficient} Random-label MNIST.}
    \label{fig:app_boundaries}
\end{figure}

In this section, we zoom-in in the \emph{data-efficient} experiment on random-label MNIST. We use \emph{Soft Reset Proximal ($\gamma$ per layer)} method with separate $\gamma$ for layer (different for each weight and for each bias) and run it for $20$ tasks on random-label MNIST. In Figure~\ref{fig:app_accuracy} we show the online accuracy as we learn over this sequence of tasks. In Figure~\ref{fig:app_boundaries}, we visualize the dynamics of parameters $\gamma$ for each layer. First of all, we see that $\gamma_t$ seems to accurately capture the task boundaries. Second, we see that the amount by which each $\gamma_t$ changes depends on the parameter type -- weights versus biases, and it depends on the layer. The architecture in this setting starts form $linear$ and goes up to $linear 4$, which represent the $4$ MLP hidden layers with a last layer $linear 4$.

\subsection{Impact of specific initialization}
\label{app:specific_initialization}

\begin{figure}[!htb]
    \centering
    \includegraphics[scale=0.4]{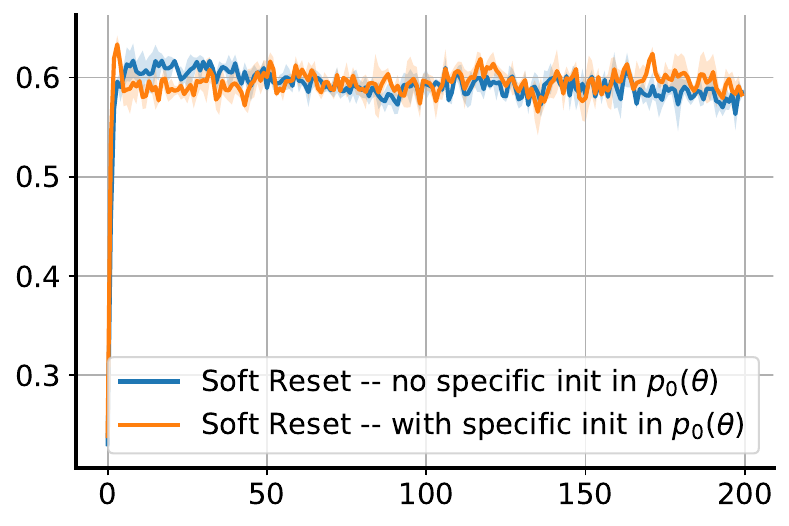}
    \caption{Impact of specific initialization $\theta_0$ as a mean of $p_0(\theta)$ in \emph{Soft Resets}. The x-axis represents task id. The y-axis represents the average task accuracy with standard deviation computed over $3$ random seeds. The task is random label MNIST -- data efficient.}
    \label{fig_app:specific_initialization}
\end{figure}

In this section, we study the impact of using specific initialization $\theta_0 \sim p_{init}(\theta)$ in $p_0(\theta)$ as discussed in Appendix~\ref{app:practical}. Using the specific initialization in \emph{Soft Resets} leads to fixing the mean of the $p_0(\theta)$ to be $\theta_0$, see~\eqref{eq_app:modified_prior}. This, in turn, leads to the predictive distribution~\eqref{eq_app:drift_from_specific_init}. In case when we are not using specific initialization $\theta_0$, the mean of $p_0(\theta)$ is $0$ and the predictive distribution is given by~\eqref{eq_app:drift_from_init}. To understand the impact of this design decision, we conduct an experiment on random label MNIST with \emph{Soft Reset}, where we either use the specific initialization or not. For each of the variants, we do a hyperparameters sweep. The results are given in Figure~\ref{fig_app:specific_initialization}. We see that both variants perform similarly.

\section{Learning parameters with estimated drift models}
\label{app:learning_with_drift}

In this section, we provide a Bayesian Neural Network algorithm to learn the distributions of NN parameters when there is a drift in the data distribution. Moreover, we provide a MAP-like inference algorithm which does not require to learn the distributions over parameters, but simply propagates the MAP estimate over these.

\subsection{Bayesian Neural Networks algorithm}
\label{sec:bnns}

In this section, we describe an algorithm for parameters update based on Bayesian Neural Networks (BNN). It is based on the online variational Bayes setting described below.

Let the family of distributions over parameters be
\begin{equation}
    \mathcal{Q} = \{ q(\theta) : q(\theta) \sim \prod_{i=1}^{D} \mathcal{N}(\theta^i; \mu_{i}, \sigma_{i}^2); \theta = (\theta^1,\ldots,\theta^D)\},
    \label{eq_app:approx_family}
\end{equation}
which is the family of Gaussian mean-field distributions over parameters $\theta \in \mathbb{R}^{D}$ (separate Gaussian per parameter). For simplicity of notation, we omit the index $i$. Let $\Gamma_t=(\gamma_1, \ldots, \gamma_t)$ be the history of observed parameters of the drift model and $\mathcal{S}_{t}=\{(x_{1},y_{1}),\ldots,(x_t,y_t)\}$ be the history of observed data. We denote by $q_t(\theta) \triangleq q_t(\theta | \mathcal{S}_{t}, \Gamma_{t-1}) \in \mathcal{Q}$ the Gaussian \emph{approximate} posterior at time $t$ with mean $\mu_t$ and variance $\sigma^2_t$ for every parameter. The approximate predictive look-ahead prior is given by
\begin{equation}
    \textstyle
    q_t(\theta |\gamma_t) = \int q_t(\theta_{t}) p(\theta | \theta_{t}, \gamma_{t}) d \theta_{t} = \mathcal{N}(\theta; \mu_t(\gamma_t), \sigma^2_t(\gamma_t)),
    \label{eq_app:approx_predictive_prior}
\end{equation}
that has parameters $\mu_t(\gamma_t) = \gamma_t \mu_t + (1-\gamma_t) \mu_0, \sigma^2_t(\gamma_t) = \gamma_t^2 \sigma^2_t + (1-\gamma_t^2) \sigma^2_0$. To see this, we will use the law of total expectation and the law total variance. For two random variables $X$ and $Y$ defined on the same space, law of total expectation says
\begin{equation}
    \mathbb{E}[Y] = \mathbb{E}[\mathbb{E}[Y|X]]
\end{equation}
and the law of total variance says
\begin{equation}
    \mathbb{V}[Y] = \mathbb{E}[\mathbb{V}[Y|X]] + \mathbb{V}[\mathbb{E}[Y|X]]
\end{equation}
In our case, from the drift model~\eqref{eq:ou_model}, we have the conditional distribution
\begin{equation}
    \theta | \theta_t = \gamma_t \theta_t + (1-\gamma_t) \mu_0 + \sqrt{(1-\gamma^2_t)\sigma^2_0} \epsilon, \epsilon \sim \mathcal{N}(0; I)
    \label{eq:cond_random}
\end{equation}
From~\eqref{eq:cond_random}, we have
\begin{align}
    \mathbb{E}[\theta | \theta_t] = \gamma_t \theta_t + (1-\gamma_t) \mu_0 \\
    \mathbb{V}[\theta | \theta_t] = (1-\gamma^2_t)\sigma^2_0
\end{align}
From here, we have that the mean is given by
\begin{align}
    \mathbb{E}[\theta] = \mathbb{E}[\mathbb{E}[\theta | \theta_t]] = \gamma_t \mu_t + (1-\gamma_t)\mu_0
    \label{eq:gaussian_mean_inferred}
\end{align}
and the variance is given by
\begin{align}
    \mathbb{V}[\theta] = \mathbb{E}[\mathbb{V}[\theta | \theta_t]] + \mathbb{V}[\mathbb{E}[\theta | \theta_t]] \\
    \mathbb{V}[\theta] =  (1-\gamma^2_t)\sigma^2_0 + \gamma^2_t \theta^2_t
    \label{eq:gaussian_variance_inferred}
\end{align}
Now, we note that $q_t(\theta |\gamma_t)$ is a Gaussian and its parameters are given by $\mathbb{E}[\theta]= \gamma_t \mu_t + (1-\gamma_t)\mu_0$ from~\eqref{eq:gaussian_mean_inferred} and by $\mathbb{V}[\theta]=(1-\gamma^2_t)\sigma^2_0 + \gamma^2_t \theta^2_t$ from~\eqref{eq:gaussian_variance_inferred}. Then, for new data $(x_{t+1}, y_{t+1})$ at time $t+1$, the \emph{approximate predictive log-likelihood} equals to
\begin{equation}
    \textstyle
    \log q_t(y_{t+1} | x_{t+1}, \gamma_t) = \log \int p(y_{t+1} | x_{t+1}, \theta) q_t(\theta | \gamma_{t}) d \theta.
    \label{eq_app:approximate_predictive_likelihood}
\end{equation}

We are looking for a new approximate posterior $q_{t+1}(\theta)$ such that
\begin{equation}
    q_{t+1}(\theta) = \arg\min_{q} \mathbb{KL} \left[q(\theta) || p(y_{t+1} | x_{t+1}, \theta) q_t(\theta |\gamma_t) \right]
    \label{eq_app:variation_bayes_update}
\end{equation}
The optimization problem~\eqref{eq_app:variation_bayes_update} can be written as minimization of the following loss
\begin{equation}
    \mathcal{F}_{t}(\theta, \gamma_t) = \mathbb{E}_{q}\left[\mathcal{L}_{t+1}(\theta) \right] + \mathbb{KL}\left[q(\theta) || q_t(\theta |\gamma_t) \right],
    \label{eq_app:elbo}
\end{equation}
since $\mathcal{L}_{t+1}(\theta)=-\log p(y_{t+1} | x_{t+1}, \theta)$. Using the fact that we are looking for a member $q \in \mathcal{Q}$ from~\eqref{eq_app:approx_family}, we can write the objective~\eqref{eq_app:elbo} as
\begin{equation}
    \mathcal{F}_{t}(\mu, \sigma, \gamma_t) = \mathbb{E}_{\epsilon \sim \mathcal{N}(0; I)}\left[\mathcal{L}_{t+1}(\mu + \epsilon \sigma) \right] + \mathbb{KL}\left[q(\theta) || q_t(\theta |\gamma_t) \right],
    \label{eq_app:elbo_gaussian}
\end{equation}
where we used the reparameterisation trick for the loss term. We now expand the regularization term to get
\begin{equation}
    \mathcal{F}_{t}(\mu, \sigma, \gamma_t) = \mathbb{E}_{\epsilon \sim \mathcal{N}(0; I)}\left[\mathcal{L}_{t+1}(\mu + \epsilon \sigma) \right] + \sum_i \left[ \frac{(\mu_i - \mu_{t,i}(\gamma_t))^2 + \sigma_i^2}{2 \sigma^2_{t,i}(\gamma_t)} - \frac{1}{2}\log \sigma_i^2 \right]
    \label{eq_app:elbo_gaussian_expanded}
\end{equation}
Since the posterior variance of NN parameters may become small, the optimization of~\eqref{eq_app:elbo_gaussian_expanded} may become numerically unstable due to division by $\sigma^2_{t,i}(\gamma_t)$. It was shown~\citep{wenzel2020good} that using small temperature on the prior led to better empirical results when using Bayesian Neural Networks, a phenomenon known as cold posterior. Here, we define a temperature per-parameter, i.e., $\lambda_{t,i} > 0$ for every time-step $t$, such that the objective above becomes
\begin{equation}
    \tilde{\mathcal{F}}_{t}(\mu, \sigma, \gamma_t; \{\lambda_{t,i}\}_i) = \mathbb{E}_{\epsilon \sim \mathcal{N}(0; I)}\left[\mathcal{L}_{t+1}(\mu + \epsilon \sigma) \right] + \sum_i \textcolor{red}{\lambda_{t,i}} \left[ \frac{(\mu_i - \mu_{t,i}(\gamma_t))^2 + \sigma_i^2}{2 \sigma^2_{t,i}(\gamma_t)} - \frac{1}{2}\log \sigma_i^2 \right]
    \label{eq_app:elbo_gaussian_expanded_temp}
\end{equation}
As said above, it is common to use the same temperature $\lambda_{t,i} = \lambda$ for all the parameters. In this work, we propose the specific choice of the temperature to be
\begin{equation}
    \lambda_{t,i} = \lambda \sigma^2_{t,i},
    \label{eq_app:temperature}
\end{equation}
where $\lambda > 0$ is some globally chosen temperature parameter. This leads to the following objective
\begin{equation}
    \hat{\mathcal{F}}_{t}(\mu, \sigma, \gamma_t;\{r{t,i}\}_i) = \mathbb{E}_{\epsilon \sim \mathcal{N}(0; I)}\left[\mathcal{L}_{t+1}(\mu + \epsilon \sigma) \right] + \frac{1}{2} \sum_i \textcolor{red}{r_{t,i}} \left[ (\mu_i - \mu_{t,i}(\gamma_t))^2 + \sigma_i^2 - \sigma^2_{t,i}(\gamma_t) \log \sigma_i^2 \right],
    \label{eq_app:elbo_gaussian_expanded_efficient_temperature}
\end{equation}
where the quantity $r_{t,i}$ is defined as
\begin{equation}
    \textstyle
    \label{eq_app:relative_variance_increase}
    r_{t,i} = \frac{\sigma^2_{t,i}}{\sigma^2_t(\gamma_t)}
    = \frac{\sigma^2_{t,i}}{\gamma_t^2 \sigma^2_{t,i} + (1-\gamma_t^2) \sigma^2_0},
\end{equation}
which represents the relative change in the posterior variance due to the drift. In the exact stationary case, when $\gamma_{t} = 1$, this ratio is $r_{t,i}=1$ while for $\gamma_{t} < 1$ , since typically $\sigma^2_{t} < \sigma^2_{0}$, we have $r_{t,i} < 1$. This means that in the non-stationary case, the strength of the regularization in~\eqref{eq_app:elbo_gaussian_expanded_efficient_temperature} in favor of the data term $\mathbb{E}_{\epsilon \sim \mathcal{N}(0; I)}\left[\mathcal{L}_{t+1}(\mu + \epsilon \sigma) \right]$, allowing the optimization to respond faster to the change in the data distribution. In practice, this data term is approximated via Monte-Carlo, i.e.
\begin{equation}
    \mathbb{E}_{\epsilon \sim \mathcal{N}(0; I)}\left[\mathcal{L}_{t+1}(\mu + \epsilon \sigma) \right] \sim \frac{1}{M} \sum_{i=1}^{M} \mathcal{L}_{t+1}(\mu + \epsilon_{i} \sigma)
    \label{eq_app:mc_loss_estimation}
\end{equation}

To find new parameters, $\mu_{t+1}$ and $\sigma_{t+1}$, we let $\mu^{0}_{t+1}=\mu_t(\gamma_t)$ and $\sigma^{0}_{t+1} = \sigma_t(\gamma_t)$ and perform multiple $K$ updates on~\eqref{eq_app:elbo_gaussian_expanded_efficient_temperature}
\begin{equation}
    \textstyle
    \label{eq_app:bayesian_updates}
    \mu^{k+1}_{t+1} = \mu^{k}_{t+1} - \alpha_{\mu} \hat{\mathcal{F}}_t(\mu^{k}_{t+1}, \sigma^{k}_{t+1}, \gamma_t, \{r{t,i}\}_i), \ \  \sigma^{k+1}_{t+1} = \sigma^{k}_{t+1} - \alpha_{\sigma} \hat{\mathcal{F}}_t(\mu^{k}_{t+1}, \sigma^k_{t+1}, \gamma_t, \{r{t,i}\}_i),
\end{equation}
where $\alpha_{\mu}$ and $\alpha_{\sigma}$ are corresponding learning rates. The full algorithm of learning the drift parameters $\gamma_t$ as well as learning the Bayesian Neural Network parameters using the procedure above is given in Algorithm~\ref{alg:alg_bayesian}.

\begin{algorithm}[tb]
   \caption{Bayesian \emph{Soft-Reset} algorithm}
   \label{alg:alg_bayesian}
\begin{algorithmic}
   \STATE {\bfseries Input:} Data-stream $\mathcal{S}_{T}=\{(x_{t}, y_{t})\})_{t=1}^{T}$
   \STATE Neural Network initial variance for every parameter $\sigma^2_0$ coming from standard NN library
   \STATE NN initializer $p_{init}(\theta)$
   \STATE Proximal cost $\lambda \geq 0$.
   \STATE Initial prior variance rescaling $p \in [0, 1]$.
   \STATE Initial posterior variance rescaling $f \in [0, 1]$.
   \STATE Learning rate for the mean $\alpha_{\mu}$ and for the standard deviation $\alpha_{\sigma}$
   \STATE Number of gradient updates $K_{\theta}$ to be applied on $\mu$ and $\sigma$
   \STATE Number of  Monte-Carlo samples $M_{\theta}$ for estimating $\mu$ and $\sigma$ in~\eqref{eq_app:mc_loss_estimation}
   \STATE Number of gradient updates $K_{\gamma}$ on drift parameter $\gamma_t$ in~\eqref{eq:drift_via_predictive_ll}
   \STATE Number of Monte-Carlo samples $M_\gamma$ to estimate $\gamma_t$ in~\eqref{eq:monte_carlo}
   \STATE Learning rate $\eta_{\gamma}$ for drift parameter
   \STATE Initial drift parameters $\gamma_0 = 1$ for every iteration.
   \STATE \textbf{Initialization}:
   \STATE Initialize NN parameters $\theta_0 \sim p_{init}(\theta)$
   \STATE Initialize prior distribution $p_0(\theta) = \mathcal{N}(\theta; 0; p^2\sigma^2_0)$ to be used for drift model~\eqref{eq:ou_model}.
   \STATE Initialize posterior $q_{0}(\theta)$ to be $\mathcal{N}(\theta; \theta_0; \sigma^2_{init})$, where $\sigma^2_{init} = f^2 p^2 \sigma^2_0$.
   \FOR{step $t=0,1,2,\ldots,T$}
    \STATE Current posterior $q_{t} = \mathcal{N}(\theta; \mu_t, \sigma^2_t)$
   \STATE For $(x_{t+1}, y_{t+1})$, predict $\hat{y}_{t+1} = f(x_{t+1}| \mu_t)$ with current posterior mean parameters $\mu_t$
   \STATE Compute performance metric based on $(y_{t+1}, \hat{y}_{t+1})$
   \STATE \textbf{Estimating the drift}
   \STATE Initialize drift parameter $\gamma^{0}_t = \gamma_0$.
   \STATE Compute $\mu_t(\gamma_t) = \gamma_t \mu_t$ and $\sigma^2(\gamma_t) = \gamma^2_t \sigma^2_t + (1-\gamma^2_t)\sigma^2_0$
    \FOR{$k=0,\ldots,K_{\gamma}-1$}
       \STATE $\gamma^{k+1}_t = \gamma^{k}_t \eta_{\gamma} \nabla_{\gamma} \log \frac{1}{M_{\gamma}} \sum_{i=1}^{M_{\gamma}} p(y_{t+1} | x_{t+1}, \mu_t(\gamma^{k}_t) + \epsilon_{i} \sigma_{t}(\gamma^{k}_t))$
   \ENDFOR
   \STATE \textbf{Updating variational posterior}
   \STATE Let $\mu^{0}_{t+1} = \gamma_t \mu_t$, $\sigma^0_{t+1} = \sqrt{\gamma^2_t \sigma^2_t + (1-\gamma^2_t)\sigma^2_0}$
   \STATE Let $r_{t,i} = \frac{\sigma^2_{t,i}}{\sigma^2_t(\gamma_t)}
    = \frac{\sigma^2_{t,i}}{\gamma_t^2 \sigma^2_{t,i} + (1-\gamma_t^2) \sigma^2_0}$ to be used in
    \FOR{$k=0,\ldots,K_{\theta}-1$}
        \STATE $\mu^{k+1}_{t+1} = \mu^{k}_{t+1} - \alpha_{\mu} \hat{\mathcal{F}}_t(\mu^{k}_{t+1}, \sigma^{k}_{t+1}, \gamma_t, \{r_{t,i}\}_i, \lambda)$
        \STATE $\sigma^{k+1}_{t+1} = \sigma^{k}_{t+1} - \alpha_{\sigma} \hat{\mathcal{F}}_t(\mu^{k}_{t+1}, \sigma^k_{t+1}, \gamma_t, \{r_{t,i}\}_i, \lambda)$
    \ENDFOR
   \ENDFOR
\end{algorithmic}
\end{algorithm}

\subsection{Modified SGD with drift model}
\label{app:map_inference}

Instead of propagating the posterior~\eqref{eq:variation_bayes_update}, we do MAP updates on~\eqref{eq:online_bayesian_update_drifted} with the prior $p_0(\theta) = \mathcal{N}(\theta; \mu_0; \sigma^2_0)$ and the posterior $q_t(\theta) = \mathcal{N}(\theta; \theta_t; s^2 \sigma^2_0)$, where $s \leq 1$ is hyperparameter controlling the variance $\sigma^2_t$ of the posterior $q_t(\theta)$. Since fixed $s$ may not capture the true parameters variance, using Bayesian method (see Appendix~\ref{sec:bnns}) is preferred but comes at a high computational cost. Instead of Bayesian update~\eqref{eq_app:variation_bayes_update}, we consider maximum a-posteriori (MAP) update
\begin{equation}
    \max_{\theta} \log p(y_{t+1} | x_{t+1}, \theta) + \log q_t(\theta | \gamma_t),
\end{equation}
with $q_t(\theta | \gamma_t)$ given by~\eqref{eq_app:approx_predictive_prior}. Denoting $\mathcal{L}_{t+1}(\theta) = \log p(y_{t+1} | x_{t+1}, \theta)$ and using the definition of $q_t(\theta | \gamma_t)$, we get the following problem
\begin{equation}
    \max_{\theta} -\mathcal{L}_{t+1}(\theta) - \sum_i \textcolor{red}{\lambda_{t,i}} \left[ \frac{(\mu_i - \mu_{t,i}(\gamma_t))^2}{2 \sigma^2_{t,i}(\gamma_t)} \right],
    \label{eq_app:map_objective}
\end{equation}
where similarly to~\eqref{eq_app:elbo_gaussian_expanded_temp}, we use a per-parameter temperature $\lambda_{t,i} \geq 0$. We choose temperature to be equal to
\begin{equation}
    \lambda_{t, i} = s^2 \sigma^2_{0, i} \lambda,
\end{equation}
where $\lambda$ is some constant. Such choice of temperature is motivated by the same logic as in~\eqref{eq_app:temperature} -- it is a constant multiplied by the posterior variance $\sigma^2_{t,i} = s^2 \sigma^2_{0, i}$. With such choice of temperature, maximizing~\eqref{eq_app:map_objective} is equivalent to minimizing
\begin{equation}
    \textstyle
     G(\theta; \lambda) = \mathcal{L}_{t+1}(\theta) + \frac{\lambda}{2} \sum_{i=1}^{D} \frac{|\theta^i - \theta^i_{t}(\gamma_t)|^2}{r_{t,i}(\gamma)}
    \label{eq_app:proximal_objective}
\end{equation}
where the regularization target for the dimension $i$ is
\begin{equation}
    \textstyle
    \theta^i_{t}(\gamma^i_t) = \gamma^i_t \theta^i_t + (1-\gamma^i_t)\mu^i_0
    \label{eq_app:adapted_reg}
\end{equation}
and the constant $r_{t,i}(\gamma)$ is given by
\begin{equation}
    \textstyle
    r_{t,i}(\gamma_t) = \left( (\gamma^i_t)^2 + \frac{1 - (\gamma^i_t)^2}{s^2} \right)
    \label{eq_app:adapted_lr}
\end{equation}
We can perform $K$ gradient updates~\eqref{eq_app:proximal_objective} with a learning rate $\alpha_t$ starting from $\theta^{0}_{t+1} = \theta_{t}(\gamma_t)$,
\begin{equation}
    \label{eq_app:many_grad_updates}
    \theta^{k+1}_{t+1} = \theta^{k}_{t+1} - \alpha_t(\gamma_t) \circ  \nabla_{\theta} G_{t+1}(\theta^{k}_{t+1}; \lambda),
\end{equation}
where the vector-valued learning rate $\alpha_t(\gamma_t)$ is given by
\begin{equation}
    \alpha_t(\gamma_t) = \alpha_t r_{t,i}(\gamma_t) = \alpha_t \left( (\gamma^i_t)^2 + \frac{1 - (\gamma^i_t)^2}{s^2} \right),
    \label{eq_app:increased_lr}
\end{equation}
with $\alpha_t$ the base learning rate. Note that doing one update is equivalent to modified SGD method~\eqref{eq:linearised_map}. Doing multiple updates on~\eqref{eq_app:many_grad_updates} allows us to perform multiple computations on the \emph{same} data. The corresponding algorithm is given in Algorithm~\ref{alg:alg_proximal}.

\begin{algorithm}[tb]
   \caption{Proximal \emph{Soft-Reset} algoritm}
   \label{alg:alg_proximal}
\begin{algorithmic}
   \STATE {\bfseries Input:} Data-stream $\mathcal{S}_{T}=\{(x_{t}, y_{t})\})_{t=1}^{T}$
   \STATE Neural Network (NN) initializing distribution $p_{init}(\theta)$ and specific initialization $\theta_0 \sim p_{init}(\theta)$
   \STATE Learning rate $\alpha_t$ for parameters and $\eta_{\gamma}$ for drift parameters
   \STATE Number of gradient updates $K_{\gamma}$ on drift parameter $\gamma_t$
   \STATE Number of gradient updates $K_{\theta}$ on NN parameters
   \STATE Proximal term cost $\lambda \geq 0$
   \STATE NN initial standard deviation (STD) scaling $p \leq 1$ (see~\eqref{eq_app:modified_prior}) and ratio $s = \frac{\sigma_t}{p\sigma_0}$.
   \FOR{step $t=0,1,2,\ldots,T$}
   \STATE For $(x_{t+1}, y_{t+1})$, predict $\hat{y}_{t+1} = f(x_{t+1}| \theta_{t})$
   \STATE Compute performance metric based on $(y_{t+1}, \hat{y}_{t+1})$
   \STATE Initialize drift parameter $\gamma^{0}_t = 1$
   \FOR{step $k=0,1,2,\ldots,K_{\gamma}$}
    \STATE Sample $\theta'_0 \sim p_{init}(\theta)$
    \STATE Stochastic update~\eqref{eq_app:stochastic_approx} on drift parameter using specific initialization~\eqref{eq_app:drift_from_specific_init}
    \STATE $\gamma^{k+1}_t = \gamma^{k}_t + \eta_{\gamma} \nabla_{\gamma} \left[\log p(y_{t+1} | x_{t+1}, \gamma_t\theta_t + (1-\gamma_t)\theta_0 + \theta'_0 p \sqrt{1-\gamma_t^2 + \gamma_t^2 s^2}) \right]_{\gamma_t = \gamma^k_t}$
   \ENDFOR
   \STATE Initialize $\theta^0_{t+1} = \theta_t(\gamma^K_t)$ with \eqref{eq_app:adapted_reg} and use $\alpha_t(\gamma^K_t) =  \alpha_t \left( (\gamma^i_t)^2 + \frac{1 - (\gamma^i_t)^2}{s^2} \right)$ with \eqref{eq_app:increased_lr}
   \FOR{step $k=0,1,2,\ldots,K_{\theta}$}
     \STATE $\theta^{k+1}_{t+1} = \theta^{k}_{t+1} - \alpha_t(\gamma_t) \circ  \nabla_{\theta} G_{t+1}(\theta^{k}_{t+1}; \lambda)$
   \ENDFOR
   \ENDFOR
\end{algorithmic}
\end{algorithm}

\section{Proof of linearisation}
\label{app:linearisation}

\paragraph{Interpretation of $\gamma_t$.} By linearising $\log p(y_{t+1}| x_{t+1}, \theta)$ around $\mu_{t}$, we can simplify~\eqref{eq:approximate_predictive_likelihood} to get
\begin{equation}
    \textstyle
    \label{eq_app:closed_form_gradient_steps}
    \mathcal{F}(\gamma_t) = (\gamma_t \odot \mu_t + (1-\gamma_t) \odot \mu_0 )^T g_{t+1} - 0.5 (\sigma^2_{t}(\gamma_t) \odot g_{t+1})^T g_{t+1} - \lambda \sum_{i=1}^{K} (\gamma_{t,i} - \gamma^{0}_{t,i})^2,
\end{equation}
where $\odot$ denotes elementwise product, $g_{t} = -\nabla \mathcal{L}_{t+1}(\mu_{t})$ is the negative gradient of the loss~\eqref{eq:loss_def} evaluated at $\mu_{t}$ and we added the $\ell_2$-penalty $ \frac{1}{2}\lambda (\gamma_{t,i} - \gamma^{0}_{t,i})^2$ to take into account the initialization.

\textbf{Proof}. We assume that the following linearisation is correct
\begin{equation}
    \log p(y_{t+1}|x_{t+1},\theta)\sim\log p(y_{t+1}|x_{t+1},\mu_t)+g_{t+1}^T(\theta-\mu_t),
\end{equation}
where
\begin{equation}
    g_{t+1}=-\nabla_{\theta} \log p(y_{t+1}|x_{t+1},\theta=\mu_t) = \nabla_{\theta} \mathcal{L}_{t+1}(\mu_t)
\end{equation}
Then, we have
\begin{equation}
    p(y_{t+1}|x_{t+1},\theta) \sim p(y_{t+1}|x_{t+1},\mu_t) \exp^{g_{t+1}^T(\theta-\mu_t)}
\end{equation}
Let's write the integral from~\eqref{eq:approximate_predictive_likelihood}
\begin{align}
    \log \int p(y_{t+1} | x_{t+1}, \theta) \exp^{-\frac{1}{2}(\theta - \mu_t(\gamma_t))^T \Sigma^{-1}_{t}(\gamma_t)(\theta - \mu_t(\gamma_t))} d\theta \frac{1}{\sqrt{(2\pi)^D}|\Sigma_{t}(\gamma_t)|} = \\
    \log \int p(y_{t+1}|x_{t+1},\mu_t) \exp^{g_{t+1}^T(\theta-\mu_t)} \exp^{-\frac{1}{2}(\theta - \mu_t(\gamma_t))^T \Sigma^{-1}_{t}(\gamma_t)(\theta - \mu_t(\gamma_t))} d\theta \frac{1}{\sqrt{(2\pi)^D}|\Sigma_{t}(\gamma_t)|} = \\
    \log p(y_{t+1}|x_{t+1},\mu_t) + \log \int \exp^{g_{t+1}^T(\theta-\mu_t)} \exp^{-\frac{1}{2}(\theta - \mu_t(\gamma_t))^T \Sigma^{-1}_{t}(\gamma_t)(\theta - \mu_t(\gamma_t))} d\theta \frac{1}{\sqrt{(2\pi)^D}|\Sigma_{t}(\gamma_t)|}
\end{align}

Consider only the exp term inside the integral:
\begin{align}
    g_{t+1}^T(\theta-\mu_t) - \frac{1}{2}(\theta - \mu_t(\gamma_t))^T \Sigma^{-1}_{t}(\gamma_t)(\theta - \mu_t(\gamma_t)) = \\
    g_{t+1}^T\theta - g_{t+1}^T\mu_t - \frac{1}{2} \theta^T \Sigma^{-1}_{t}(\gamma_t) \theta + \theta^T \Sigma^{-1}_{t}(\gamma_t) \mu_t(\gamma_t) -\frac{1}{2} \mu_t(\gamma_t)^T \Sigma^{-1}_{t}(\gamma_t) \mu_t(\gamma_t) = \\
    -\frac{1}{2} \theta^T \Sigma^{-1}_{t} \theta + \theta^T ( \Sigma^{-1}_{t}(\gamma_t) \mu_t(\gamma_t) + g_{t+1}) - g_{t+1}^T\mu_t -\frac{1}{2} \mu_t(\gamma_t)^T \Sigma^{-1}_{t}(\gamma_t) \mu_t(\gamma_t) = \\
    -\frac{1}{2} \left( \theta^T \Sigma^{-1}_{t} \theta - 2 \theta^T ( \Sigma^{-1}_{t}(\gamma_t) \mu_t(\gamma_t) + g_{t+1}) \right) - g_{t+1}^T\mu_t -\frac{1}{2} \mu_t(\gamma_t)^T \Sigma^{-1}_{t}(\gamma_t) \mu_t(\gamma_t)
\end{align}

Let's focus on this term
\begin{align}
    -\frac{1}{2} \left( \theta^T \Sigma^{-1}_{t} \theta - 2 \theta^T ( \Sigma^{-1}_{t}(\gamma_t) \mu_t(\gamma_t) + g_{t+1}) \right) = \\
    -\frac{1}{2} \left( \theta^T \Sigma^{-1}_{t} \theta - 2 \theta^T \Sigma^{-1}_{t} b(\gamma_t) \right) = \\
    -\frac{1}{2} \left( \theta^T \Sigma^{-1}_{t} \theta - 2 \theta^T \Sigma^{-1}_{t} b(\gamma_t) + b(\gamma_t)^T \Sigma^{-1}_{t} b(\gamma_t) - b(\gamma_t)^T \Sigma^{-1}_{t} b(\gamma_t)\right) = \\
    -\frac{1}{2} \left( \theta^T \Sigma^{-1}_{t} \theta - 2 \theta^T \Sigma^{-1}_{t} b(\gamma_t) + b(\gamma_t)^T \Sigma^{-1}_{t} b(\gamma_t) \right) + \frac{1}{2} b(\gamma_t)^T \Sigma^{-1}_{t} b(\gamma_t) = \\
    -\frac{1}{2} (\theta - b(\gamma_t))^T \Sigma^{-1}_{t} (\theta - b(\gamma_t)) + \frac{1}{2} b(\gamma_t)^T \Sigma^{-1}_{t} b(\gamma_t)
\end{align}
where
\begin{equation}
    b(\gamma_t) = \Sigma_t(\gamma_t) \left[ \Sigma^{-1}_t(\gamma_t) \mu_t(\gamma_t) + g_{t+1} \right]
\end{equation}

Therefore, the integral could be written as
\begin{align}
    \log \int \exp^{g_{t+1}^T(\theta-\mu_t)} \exp^{-\frac{1}{2}(\theta - \mu_t(\gamma_t))^T \Sigma^{-1}_{t}(\gamma_t)(\theta - \mu_t(\gamma_t))} d\theta \frac{1}{\sqrt{(2\pi)^D}|\Sigma_{t}(\gamma_t)|} = \\
    \frac{1}{2} b(\gamma_t)^T \Sigma^{-1}_{t} b(\gamma_t) + \log \int \exp^{-\frac{1}{2} (\theta - b(\gamma_t))^T \Sigma^{-1}_{t} (\theta - b(\gamma_t))} d\theta \frac{1}{\sqrt{(2\pi)^D}|\Sigma_{t}(\gamma_t)|} - g_{t+1}^T \mu_t - \frac{1}{2}\mu_t(\gamma_t)^T\Sigma_t^{-1}(\gamma_t) \mu_t(\gamma_t) = \\
    \frac{1}{2} b(\gamma_t)^T \Sigma^{-1}_{t}(\gamma_t) b(\gamma_t) - g_{t+1}^T \mu_t - \frac{1}{2}\mu_t(\gamma_t)^T\Sigma_t^{-1}(\gamma_t) \mu_t(\gamma_t)
\end{align}
Now, we only keep the terms depending on $\gamma_t$
\begin{align}
    \frac{1}{2} b(\gamma_t)^T \Sigma^{-1}_{t}(\gamma_t) b(\gamma_t) -  \frac{1}{2}\mu_t(\gamma_t)^T\Sigma_t^{-1}(\gamma_t) \mu_t(\gamma_t) = \\
    \frac{1}{2} \left[ \Sigma^{-1}_t(\gamma_t) \mu_t(\gamma_t) + g_{t+1} \right]^T \Sigma_t(\gamma_t) \left[ \Sigma^{-1}_t(\gamma_t) \mu_t(\gamma_t) + g_{t+1} \right] - \frac{1}{2}\mu_t(\gamma_t)^T\Sigma_t^{-1}(\gamma_t) \mu_t(\gamma_t) = \\
    \frac{1}{2} \mu_t(\gamma_t)^T \Sigma^{-1}_t(\gamma_t) \mu_t(\gamma_t) + g_{t+1}^T \mu_t(\gamma_t) + g_{t+1}^T \frac{1}{2} \Sigma_t(\gamma_t) g_{t+1} - \frac{1}{2}\mu_t(\gamma_t)^T\Sigma_t^{-1}(\gamma_t) \mu_t(\gamma_t) = \\
    g_{t+1}^T \mu_t(\gamma_t) + \frac{1}{2}g_{t+1}^T \Sigma_t(\gamma_t) g_{t+1}
\end{align}
Since $\Sigma_t(\gamma_t) = diag(\sigma^2_t \gamma_t^2 + (1-\gamma_t^2)\sigma^2_0)$, we recover
\begin{equation}
    g_{t+1}^T (\gamma_t \odot \mu_t + (1-\gamma_t) \odot \mu_0) + \frac{1}{2} g_{t+1}^T \left((\sigma^2_t \gamma_t^2 + (1-\gamma_t^2)\sigma^2_0) \odot g_{t+1}\right)
\end{equation}

Now, we add an l2-penalty $\frac{\lambda}{2} ||\gamma_t - \gamma_{t,i}^0||^2$ and we get

\begin{equation}
    F(\gamma_t) = g_{t+1}^T (\gamma_t \odot \mu_t + (1-\gamma_t) \odot \mu_0) + \frac{1}{2} g_{t+1}^T \left((\sigma^2_t \gamma_t^2 + (1-\gamma_t^2)\sigma^2_0) \odot g_{t+1}\right) - \frac{\lambda}{2} ||\gamma_t - \gamma_{t,i}^0||^2
\end{equation}

Let's take the gradient wrt $\gamma_t$, we get
\begin{align}
    \nabla  F(\gamma_t) = g_{t+1}^T(\mu_t - \mu_0) + g_{t+1}^T \left((\sigma^2_t \gamma_t - \sigma^2_0 \gamma_t) \odot g_{t+1} \right) - \lambda (\gamma_t - \gamma_{t,i}^0) = \\
    g_{t+1}^T (\mu_t - \mu_0) + \lambda \gamma_{t,i}^0 - \gamma_t (\lambda + g_{t+1}^T \left((\sigma^2_0 - \sigma^2_t) \odot g_{t+1} \right) = 0
\end{align}
Then
\begin{equation}
    \gamma_t = \frac{g_{t+1}^T (\mu_t - \mu_0) + \lambda \gamma_{t,i}^0}{\lambda + g_{t+1}^T \left((\sigma^2_0 - \sigma^2_t) \odot g_{t+1} \right)}
\end{equation}

If $\gamma_t$ is defined per parameter, this becomes
\begin{equation}
    \gamma_t = \frac{g_{t+1} (\mu_t - \mu_0) + \lambda \gamma_{t,i}^0}{\lambda + g_{t+1}^2 (\sigma^2_0 - \sigma^2_t)}
\end{equation}

\section{Toy illustrative example for SGD underperformance in the non-stationary regime}
\label{app:toy_sgd_failure}

\paragraph{Illustrative example of SGD on a non-stationary stream.} We consider a toy problem of tracking a changing mean value. Let the observations in the stream $\mathcal{S}_{t}$ follow $y_{t} = \mu_t + \sigma \epsilon$, where $\epsilon \sim \mathcal{N}(0, 1)$, $\sigma=0.01$. Every $50$ timesteps the mean $\mu_{t}$ switches from $-2$ to $2$. We fit a 3-layer MLP with layer sizes $(10, 5, 1)$ and ReLU activations, using SGD with two different  choices for the learning rate: $\alpha=0.05$ and $\alpha=0.15$. Moreover,
given that we know
when a switch of the mean happens,
we reset (or not reset) all the parameters at every switch as we run SGD. Only during the reset, we use different learning rate $\beta=0.05$ or $\beta=0.15$. Using higher learning rate during reset allows SGD to learn faster from new data. We also ran SGD with $\alpha=0.05$ and $\beta=0.15$, where the higher learning rate is used during task switch but we do not reset the parameters. We found that it performed the same as SGD with $\alpha=0.05$, which highlights the benefit of reset.

\begin{figure}[!htb]
    \centering
    \includegraphics[scale=0.4]{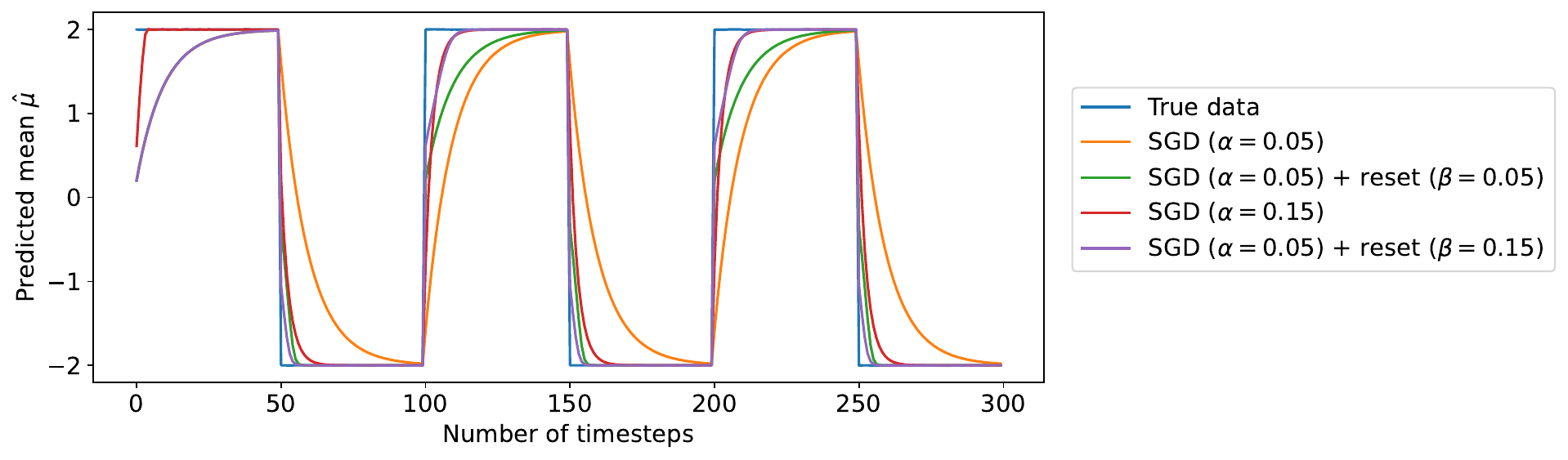}
    \caption{Non-stationary mean tracking with SGD.}
    \label{fig:illustrative_mean_tracking}
\end{figure}

We report the predicted mean $\hat{\mu}_{t}$ for all SGD variants in Figure~\ref{fig:illustrative_mean_tracking}. We see that after the first switch of the mean, the SGD without reset takes more time to learn the new mean compared to the version with parameters reset. Increasing the learning rate 
speeds up the adaptation to new data, but it still remains slower during the 
mean change from $2$ to $-2$ compared to the version that resets parameters. This example highlights that resets could be highly beneficial for improving the performance of SGD which could be slowed down by the implicit regularization towards the previous parameters $\theta_{t}$ and the impact of the regularization strength induced by the learning rate.

\section{Using arbitrary drift models}

\subsection{Using arbitrary drift models}
\label{sec:arbitrary_models}

The approach described in section~\ref{sec:map_inference} provides a general strategy of incorporating arbitrary Gaussian drift models $p(\theta | \theta_t; \psi_t) = \mathcal{N}(\theta; f(\theta_t; \psi_t); g^2(\theta_t; \psi_t))$ which induces proximal optimization problem
\begin{equation}
    \label{eq:general_optimization_problem}
    \theta_{t+1} = \arg\min_{\theta} \mathcal{L}_{t+1}(\theta) + \frac{1}{2 g(\theta_t; \psi_t)} || \theta - f(\theta_t; \psi_t)||^2
\end{equation}
The choice of $f(\theta_t; \psi_t)$ and $g(\theta_t; \psi_t)$ affects the behavior of the estimate $\theta_{t+1}$ from \eqref{eq:general_optimization_problem} and ultimately depends on the problem in hand. The objective function of the form ~\eqref{eq:general_optimization_problem} was studied in context of online convex optimization in \citep{hall2013dynamical},\citep{khodak2019adaptive}, where the underlying algorithms estimated the \emph{deterministic} drift model online. These worked demonstrated improved regret bounds depending on model estimation errors. This approach could also be used together with a Bayesian Neural Network (BNN).

\end{document}